\documentclass[12pt]{article}

\usepackage{color}

\usepackage{amsfonts,amstext,amsmath,amssymb,amsthm}
\usepackage{multirow}
\usepackage{subfigure}
\usepackage[utf8]{inputenc} 
\usepackage[T1]{fontenc}    
\usepackage{hyperref}       
\hypersetup{
	colorlinks=true,       
	linkcolor=blue,        
	citecolor=blue,        
	filecolor=magenta,     
	urlcolor=blue
}
\usepackage{url}            
\usepackage{booktabs}       
\usepackage{amsfonts}       
\usepackage{nicefrac}       
\usepackage{microtype}      
\usepackage{lipsum}
\usepackage{graphicx}
\usepackage{wrapfig}
\graphicspath{ {./images/} }
\usepackage{cite}
\usepackage{algorithm,algorithmic}
\usepackage[symbol]{footmisc}
\usepackage{setspace}   
\usepackage{geometry}
\usepackage{fontawesome}

\usepackage[title,toc]{appendix}
\usepackage{minitoc}
\noptcrule
\renewcommand \thepart{}
\renewcommand \partname{}

\usepackage[dvipsnames]{xcolor}

\def\T{{ \mathrm{\scriptscriptstyle T} }}
\newcommand{\norm}[1]{\|#1\|}

\newcommand{\cA}{\mathcal{A}}

\newcommand{\cE}{\mathcal{E}}

\newcommand{\cG}{\mathcal{G}}

\newcommand{\cV}{\mathcal{V}}

\newcommand{\PP}{\mathbb{P}}

\newcommand{\RR}{\mathbb{R}}

\newtheoremstyle{mytheoremstyle} 
    {\topsep}                    
    {\topsep}                    
    {\normalfont}                   
    {}                           
    {\bfseries}                   
    {.}                          
    {.5em}                       
    {}  

\theoremstyle{mytheoremstyle}

\ifx\BlackBox\undefined
\newcommand{\BlackBox}{\rule{1.5ex}{1.5ex}}  
\fi

\ifx\QED\undefined
\def\QED{~\rule[-1pt]{5pt}{5pt}\par\medskip}
\fi

\ifx\proof\undefined
\newenvironment{proof}{\par\noindent{\bf Proof\ }}{\hfill\BlackBox\\[2mm]}
\fi

\ifx\theorem\undefined
\newtheorem{theorem}{Theorem}
\fi
\ifx\example\undefined
\newtheorem{example}{Example}
\fi
\ifx\property\undefined
\newtheorem{property}{Property}
\fi
\ifx\lemma\undefined
\newtheorem{lemma}{Lemma}
\fi
\ifx\proposition\undefined
\newtheorem{proposition}{Proposition}
\fi
\ifx\remark\undefined
\newtheorem{remark}{Remark}
\fi
\ifx\corollary\undefined
\newtheorem{corollary}{Corollary}
\fi
\ifx\definition\undefined
\newtheorem{definition}{Definition}
\fi
\ifx\conjecture\undefined
\newtheorem{conjecture}{Conjecture}
\fi
\ifx\fact\undefined
\newtheorem{fact}{Fact}
\fi
\ifx\claim\undefined
\newtheorem{claim}{Claim}
\fi
\ifx\assumption\undefined
\newtheorem{assumption}{Assumption}
\fi
\ifx\condition\undefined
\newtheorem{condition}{Condition}
\fi

\begin{document}
%
\title{Causal Graph Discovery \\ from Self and Mutually Exciting Time Series}
%
%
%

\author{Song Wei$^\mathrm{a}$ \quad Yao Xie$^{\mathrm{a} \star}$ \quad Christopher S. Josef$^\mathrm{b}$ \quad Rishikesan Kamaleswaran$^\mathrm{c,d}$\\
\\
  \small $^\mathrm{a}$School of Industrial and Systems Engineering, Georgia Institute of Technology. \\
  \small $^\mathrm{b}$Department of Surgery, Emory University School of Medicine.\\
  \small $^\mathrm{c}$Department of Biomedical Informatics, Emory University School of Medicine.\\
  \small $^\mathrm{d}$Department of Biomedical Engineering, Georgia Institute of Technology.
}

\date{\vspace{-20pt}}
\maketitle

\footnotetext{$^{\star}$Corresponding author: Yao Xie (e-mail: {yao.xie@isye.gatech.edu}).}


\begin{abstract}
We present a generalized linear structural causal model, coupled with a novel data-adaptive linear regularization, to recover causal directed acyclic graphs (DAGs) from time series.
By leveraging a recently developed stochastic monotone Variational Inequality (VI) formulation, we cast the causal discovery problem as a general convex optimization. Furthermore, we develop a non-asymptotic recovery guarantee and quantifiable uncertainty by solving a linear program to establish confidence intervals for a wide range of non-linear monotone link functions. 
We validate our theoretical results and show the competitive performance of our method via extensive numerical experiments. 
Most importantly, we demonstrate the effectiveness of our approach in recovering highly interpretable causal DAGs over Sepsis Associated Derangements (SADs) while achieving comparable prediction performance to powerful ``black-box'' models such as XGBoost. Thus, the future adoption of our proposed method to conduct continuous surveillance of high-risk patients by clinicians is much more likely.   
\end{abstract}



{\small \noindent\textbf{Keywords:} Causal structural learning; Directed acyclic graph; Data-adaptive approach; Generalized linear model.}

\newpage
\doparttoc 
\faketableofcontents 

\part{} 


\renewcommand*{\thefootnote}{\arabic{footnote}}

\vspace{-0.45in}
\section{Introduction}

Continuous, automated surveillance systems incorporating machine learning models are becoming increasingly common in healthcare environments. These models can capture temporally dependent changes across multiple patient variables and enhance a clinician's situational awareness by providing an early alarm of an impending adverse event. Among those adverse events, we are particularly interested in sepsis, which is a life-threatening medical condition contributing to one in five deaths globally \cite{world2020global} and stands as one of the most important cases for automated in-hospital surveillance. 
Recently, many machine learning methods have been developed to predict the onset of sepsis, utilizing electronic medical record (EMR) data \cite{fleuren2020machine}. A recent sepsis prediction competition \cite{reyna2019early} demonstrated the robust performance of XGBoost models \cite{du2019automated,zabihi2019sepsis,yang2020explainable}; meanwhile, Deep Neural Networks \cite{shashikumarDeepAISEInterpretableRecurrent2021} are also commonly used. However, most approaches offer an alert adjudicator very little information pertaining to the reasons for the prediction, leading many to refer to them as ``black box'' models. Thus, model predictions related to disease identification, particularly for complex diseases, still need to be adjudicated (i.e., interpreted) by a clinician before further action (i.e., treatment) can be initiated. Among the aforementioned works, \cite{yang2020explainable} provided one of the best attempts at identifying causality for their models' predictions by reporting feature importance at a global level for all patients; still, this did not convey which features were most important in arriving at a given prediction for an individual patient. The common lack of interpretability of many clinical models, particularly those related to sepsis, suggests a strong need for principled methods to study the interactions among time series in medical settings.

A natural approach is to model relationships between time series and their effects on sepsis through {\it Granger causal graphs}.
Granger causality assesses whether the history/past of one time series is predictive of another and is a popular notion of causality for time series data. Traditional approaches typically rely on a linear vector autoregressive (VAR) model \cite{lutkepohl2005new} and consider tests on the VAR coefficients in the bivariate setting. However, it has been recognized that such traditional VAR models have many limitations, including linearity assumption \cite{shojaie2021granger} and the absence of directed acyclic graph (DAG) structure, which is essential in causal structural learning \cite{pearl2009causality}. 
On the one hand, recent advancements in non-linear Granger causality consider Neural Network based approaches coupled with sparsity-inducing penalties \cite{tank2018neural,khanna2019economy}, but render the optimization problem non-convex. 
On the other hand, structural vector autoregressive (SVAR) models, which combines the structural causal model (SCM) with the VAR model, leverage DAG-inducing penalties to uncover causal DAGs. Notable contributions include \cite{zhang2009causality,hyvarinen2010estimation}, who leveraged adaptive Lasso \cite{zou2006adaptive} to recover a Causal DAG, and \cite{pamfil2020dynotears}, who applied a recently proposed continuous DAG characterization \cite{zheng2018dags} to encourage such DAG structure.
Despite recent advancements, leveraging the well-developed convex optimization techniques to learn a causal DAG remains an open problem. Moreover, the commonly considered DAG structure is less than satisfactory since it cannot capture the lagged self-exciting components, which are important for clinicians to understand how long a node (i.e., a certain type of disease or organ dysfunction) will last once it is triggered.

\begin{figure*}[!htp]
\centerline{
\includegraphics[width = \textwidth]{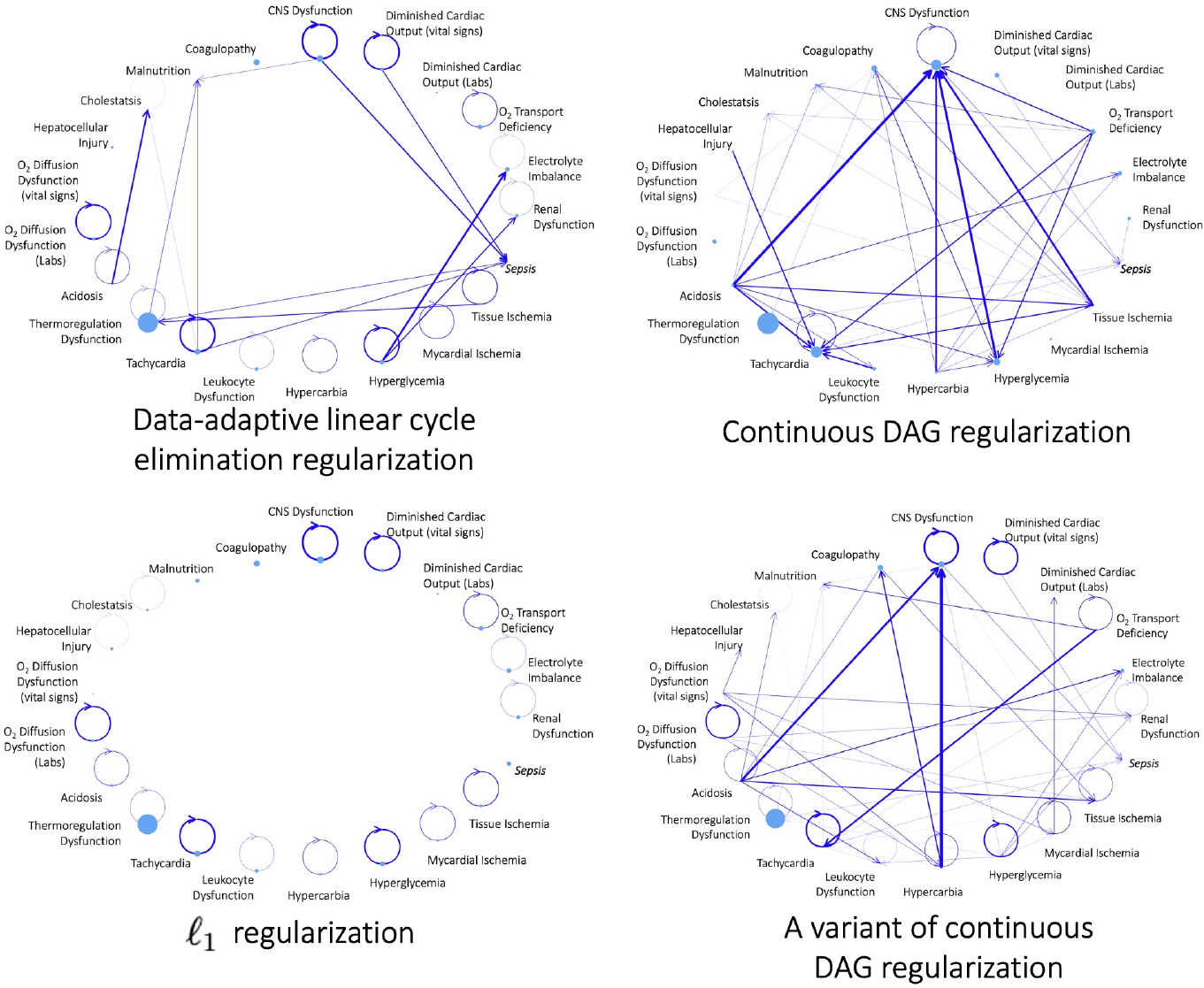}
}
\caption{Causal DAGs for SADs obtained via discrete-time Hawkes network coupled with various types of regularization. The node's size is proportional to the background intensity, and the width of the directed edge is proportional to the exciting effect magnitude. The out-of-sample total CE loss is 2.89 (proposed regularization),
2.75 ($\ell_1$ regularization),
3.37 (DAG regularization \cite{zheng2018dags,ng2020role}) and
4.13 (a variant of DAG regularization).
Our proposed regularization can help output a DAG with self-exciting edges while achieving good prediction accuracy; although $\ell_1$ regularization achieves the best CE loss, it fails to capture the interactions among SADs, leading to a very uninformative graph.
}
\label{fig:GC_graph_bp_reg} 
\end{figure*}

In this work, we present a generalized linear structural causal model to recover the causal graph from mutually exciting time series, {called} {\it discrete-time Hawkes network}.
To encourage the desired DAG structure, we propose a novel data-adaptive linear regularizer, enabling us to cast the causal structural learning problem as a convex optimization via a monotone operator Variational Inequality (VI) formulation. 
Furthermore, performance guarantees {are established} via recent advances in optimization \cite{juditsky2019signal,juditsky2020convex} by developing a non-asymptotic estimation error bound verified by numerical examples. 
We show the good performance of our proposed method and validate our theoretical findings using extensive numerical experiments. In particular, {our real data experiments demonstrate} that our proposed method can achieve comparable prediction performance to powerful black-box methods such as XGBoost while outputting highly interpretable causal DAGs for Sepsis Associated Derangements (SADs), as shown in Figure~\ref{fig:GC_graph_bp_reg}.
Although this work only shows the effectiveness of our approach in causal DAG recovery for SADs in medical {settings}, it can be broadly applicable to other applications. 

\subsection{Motivating application {and dataset}}\label{sec:motivating_eg}
This work is motivated by a real study on Sepsis, {which is formally defined as life-threatening organ dysfunction caused by a dysregulated host response to infection \cite{singer2016third}. In a recent study of adult sepsis patients \cite{seymourTimeTreatmentMortality2017}, each hour of delayed treatment was associated with higher risk-adjusted in-hospital mortality (odds ratio, 1.04 per hour), and thus early recognition of the physiologic aberrations preceding sepsis would afford clinicians more time to intervene and may contribute to improving outcomes.} 
{Specifically, we handle a large-scale dataset containing} in-hospital EMR derived from the Grady hospital system (an academic, level 1 trauma center located in Atlanta, GA) spanning 2018-2019. The data was collected and analyzed in accordance with Emory Institutional Review Board approved protocol \#STUDY00000302. We unitize a retrospective cohort of patients created in our prior work \cite{wei2022granger}, where the patients were included in the Sepsis-3 cohort if they met Sepsis-3 criteria while in the hospital and were admitted for more than 24 hours. The resulting descriptive statistics are provided in Table~\ref{table:demo}. 
The raw features of each patient include
\begin{itemize}
    \item {\it Vital Signs.}
In Intensive Care Unit (ICU) environments, vital signs are normally recorded at hourly intervals. However, patients on the floor may only have vital signs measured once every 8 hours. 
\item {\it Lab Results.}
 The laboratory tests are most commonly collected once every 24 hours. However, this collection frequency may change based on the severity of a patient's illness.
\end{itemize}

\begin{table}[htp]
\caption{Median (Med.) and interquartile range (IQR) of patients demographics. The unit of time measurement for patient trajectory length is an hour. {We truncate the data based on expert advice to ensure both patient cohorts have comparable Sequential Organ Failure Assessment (SOFA) scores; refer to \cite{wei2022granger} for patient cohort construction}.}\label{table:demo}
\begin{center}
\begin{small}
\resizebox{\textwidth}{!}{%
\begin{tabular}{lcccc}
\toprule[1pt]\midrule[0.3pt]
& \multicolumn{2}{c}{{Sepsis-3 patients}} & \multicolumn{2}{c}{{Non-sepatic patients}} \\ 
Year & 2018 ($n=409$) & 2019 ($n = 454$) & 2018 ($n = 960$) & 2019 ($n = 1169$) $^\star$ \\
\cmidrule(l){2-3} \cmidrule(l){4-5}
{Age (Med. and IQR)} & 58 (38 - 68) & 59 (46 - 68) & 56 (38 - 67) & 55 (37 - 66) \\
{Gender (female percentage)} & 30.1$\%$ & 36.6$\%$ & 37.1$\%$ & 35.8$\%$ \\
{Sofa score (mean)} & 3.32& 3.14 & 2.18 & 2.28  \\
{Trajectory length (Med. and IQR)} & 25 (25 - 25) & 25 (25 - 25) & 17 (13 - 22) & 17 (13 - 22) \\
\midrule[0.3pt]\bottomrule[1pt]
\multicolumn{5}{l}{$\star$ $n$ represents the total number of patients in the corresponding cohort.}
\end{tabular}
}
\end{small}
\end{center}
\end{table}

The goal is to construct a predictive model for sepsis and an interpretable causal DAG that captures the interactions among those vital signs and Lab results. One difficulty comes from synchronous and continuous-valued observations assumption --- in many applications, especially in the medical setting, the observations can be both continuous and categorical-valued. They can also be asynchronous and sampled with different frequencies. For example, vital signs are recorded regularly, whereas Lab results are only ordered when clinically necessary. Since the absence of a lab carries meaning itself, this cannot be formulated into a missing data problem. One method to obtain interpretable predictive models is to consider their syndromic nature --- there is often a constellation of different physiologic derangements that can combine to create the condition. For example, our prior works \cite{wei2021inferringb,wei2022granger} leveraged expert opinion to identify the distinct types of measurable, physiologic change that accompanies sepsis-related illness, which is called SADs; see their definition in {Table~\ref{table:SADcutoff} in Appendix~\ref{appendix:lab_vital_SAD}}. However, the clinician did not determine the relationship among the SADs; rather, these relationships were an output of model fitting. Although there is a recently proposed principled method to handle such mixed-frequency time series \cite{tank2019identifiability}, we adopt a similar approach with \cite{wei2022granger} in this study based on expert advice. {In Section~\ref{sec:real_data}, we will report the analysis results of this dataset using our proposed method.}

\subsection{Literature}\label{subsec:literature}
We briefly review several closely related areas and defer an extended literature survey to Appendix~\ref{appendix:extended_literature}.

\subsubsection{Causal structural learning}
Structural causal model-based causal discovery methods often boil down to maximizing a score function within the DAG family \cite{glymour2019review}, making efficient DAG learning fundamental to causal discovery.
However, learning DAGs from observational data, i.e., the structural learning problem, is NP-hard due to the combinatorial acyclicity constraint \cite{chickering2004large}. This motivated many research efforts to find efficient approaches for learning DAGs.
Recently, \cite{yuan2019constrained} proposed an indicator function-based approach to enumerate and eliminate all possible directed cycles; they used truncate $\ell_1$-function as a continuous surrogate of indicator function and proposed to use alternating direction method of multipliers to solve the problem numerically.
Later on, \cite{manzour2021integer} followed this approach, transferred indicators into binary variables, and leveraged mixed integer programming to solve the problem.
In addition, there are also dynamic programming-based approaches, e.g., \cite{loh2014high}, but they are not scalable in high dimensions unless coupled with a sparsity-inducing penalty, e.g., $A*$ Lasso \cite{xiang2013Lasso}. 

One notable recent contribution in structural learning is \cite{zheng2018dags}, who formulated the DAG recovery problem as a constrained continuous optimization via a smooth DAG characterization; they applied an augmented Lagrangian method to transfer constraint as penalty and achieved efficient DAG recovery.
Later on, \cite{ng2020role} proposed to use the non-convex DAG characterization as a penalty directly and showed an asymptotic recovery guarantee for linear Gaussian models. 
Other notable extensions along this direction include a discrete backpropagation method, exploration of low-rank structure \cite{fang2020low} and neural DAG learning \cite{yu2019dag,ke2019learning,lachapelle2019gradient}. 
We refer readers to \cite{scanagatta2019survey,scholkopf2021toward,kitson2021survey,vowels2022d} for systematic surveys on structural learning and causal discovery.

We would like to highlight that the DAG structure with lagged self-exciting components considered in this work is new in the literature. Existing works typically allow directed cycles in the adjacency matrices representing the lagged effects \cite{zhang2009causality,hyvarinen2010estimation,pamfil2020dynotears,tong2022bayesian,wei2022granger}. As a question of science, we believe those lagged cycles are less explainable. For example, our prior work \cite{wei2022granger} discovered a ``Renal Dysfunction $\rightarrow$ O$_2$ Diffusion Dysfunction $\rightarrow$ Renal Dysfunction'' cyclic chain pattern, but we believe the ``Renal Dysfunction $\rightarrow$ O$_2$ Diffusion Dysfunction'' coupled with the self-exciting pattern of Renal Dysfunction uncovered by our proposed method here is more convincing; see the bottom right panel in Figure~\ref{fig:GC_graph_bp_reg}.

\subsubsection{Causality for time series}
One line of research \cite{lutkepohl2005new} combines SCM and VAR models and develops the so-called structural vector autoregressive models to help uncover the causal graphs with certain desired structures, such as DAG structure. Notable contributions include \cite{zhang2009causality,hyvarinen2010estimation}, who applied adaptive Lasso \cite{zou2006adaptive} to encourage the DAG structure. Moreover, following \cite{shimizu2006linear}, \cite{hyvarinen2010estimation} extended the finding that the non-Gaussian measurement noise helps the model identifiability to time series setting; later on, \cite{tank2019identifiability} further proved identifiability of SVAR models of order one under arbitrary subsampling and mixed frequency scheme. 
In addition to adaptive Lasso, there are also other approaches to encouraging DAG structure in the SVAR model, such as the aforementioned continuous DAG characterization \cite{pamfil2020dynotears}.
As a comparison, our proposed generalized linear model (GLM) can be reformulated into a {\it stochastic} SCM by using Gumbel-Max trick/technique \cite{maddison2017concrete,jang2017categorical,oberst2019counterfactual,lorberbom2021learning,noorbakhsh2021counterfactual}, which is slightly different from the {\it deterministic} SCMs with measurement noise in SVAR models \cite{zhang2009causality,hyvarinen2010estimation,pamfil2020dynotears}.
Moreover, compared with the commonly adopted DAG-inducing penalties in SVAR models, e.g., the continuous, differentiable but non-convex DAG characterization \cite{zheng2018dags}, our proposed data-adaptive linear method for structural learning approach is not only convex but also flexible in the sense that it can encourage a DAG structure while keeping lagged self-exciting components.

Another line of research focuses on non-linear Granger causality. Common non-linear approaches consider additive non-linear effects from history that decouple across multiple time series, such as \cite{sindhwani2012scalable}, which leveraged a separable matrix-valued kernel to infer the non-linear Granger causality. To further capture the potential non-linear interactions between predictors, Neural Networks coupled with sparsity-inducing penalties are adopted \cite{tank2018neural,khanna2019economy}. 
Even though our GLM can be viewed as a Neural Network without a hidden layer, our model is convex, theoretically grounded, and easy to train, which are the major advantages over Neural Network-based methods.
In addition, there are also efforts to tackle the high-dimensionality via regularization, such as group Lasso \cite{bolstad2011causal,basu2015network} and nuclear norm regularization \cite{basu2019low}. For a comprehensive survey on Granger causality, we refer readers to \cite{shojaie2021granger}.

\subsection{Notations}
We use $\RR_+$ to denote the collection of non-negative real numbers, i.e., $\RR_+ =  [0,\infty)$. 
For integers $0 < m \leq n$, we denote $[m:n] = \{m,\dots,n\}$; in a special case where $m=1$, we denote $[n] = \{1,\dots,n\}$.
Superscript $^\T$ denotes vector/matrix transpose; column vectors $\mathbf{1}_d = (1,\dots,1)^\T \in \RR^d$, $\mathbf{0}_d = (0,\dots,0)^\T \in \RR^d$, $e_{i,d} \in \RR^{d}$ is the standard basis vector with its $i$-th element being one and matrix $I_d \in \RR^{d \times d}$ denotes the $d$-by-$d$ identity matrix; $\operatorname{tr}(e^A)$ stands for the trace of the matrix exponential of matrix $A$. For vectors $a, b \in \RR^d$, the comparison $a \leq b$ is element-wise. In addition, we use $\nabla$ to denote the derivative operator; we use $\langle \cdot, \cdot \rangle$ to denote the standard inner product in Euclidean space, $\norm{\cdot}_p$ to denote the vector $\ell_p$ norm and $\norm{\cdot}_F$ to denote the matrix $F$-norm.

\section{Discrete-Time Hawkes Network}\label{sec:proposed_method}

\subsection{Set-up and background}
Consider mixed-type observations over a time horizon $T \geq 1$: we observe $d_1$ sequences of binary time series $\{y_1^{(i)},\dots,y_T^{(i)}\}$, $i \in [d_1]$, which represent $d_1$ type of events' occurrences, $d_2$ sequences of continuous-values time series $\{x_1^{(i)},\dots,x_T^{(i)}\}, i \in [d_2]$, and $d_3$ static variables $z_1,\dots,z_{d_3}$. 
In the following, we will refer to the binary variable as node variables, and our primary goal is to recover the graph structure over those $d_1$ nodes.

Linear multivariate Hawkes process (MHP) models the mutual inter-dependence among variables by considering a conditional intensity of event occurrence, which is jointly determined by a deterministic background and a self-exciting (or inhibiting) term depending on its history observations. 
Given that the intensity has a natural interpretation as the instantaneous probability and is inspired by linear MHP with the exponential decaying kernel, we model the probability of occurrence for the $i$-th node variable, $i \in [d_1]$, at time step $t \in [2:T]$ as follows:
\begin{align}\label{eq:notearDAGawkes_model}
    \mathbb{P} ( y_t^{(i)} =1 | \mathcal{H}_{t-1} ) =  \nu_i 
    + \sum_{j = 1}^{d_3} \gamma_{ij} z_j  + \sum_{k = 1}^{t-1} \bigg( \sum_{j = 1}^{d_2} \beta_{ij} x_{t-k}^{(j)} e^{-R k} +  \sum_{j = 1}^{d_1} \alpha_{ij} y_{t-k}^{(j)} e^{-R k}\bigg),
\end{align}
where $\mathcal{H}_t$ denotes the history observation up to time $t$. To ensure the right-hand side (RHS) of the above equation is a valid probability, we add the following constraint:
\begin{equation}\label{eq:notearDAGawkes_positivity}
   0 \leq \nu_i 
    + \sum_{j = 1}^{d_3} \gamma_{ij} z_j  + \sum_{k = 1}^{t-1} \bigg( \sum_{j = 1}^{d_2} \beta_{ij} x_{t-k}^{(j)} e^{-R k} +  \sum_{j = 1}^{d_1} \alpha_{ij} y_{t-k}^{(j)} e^{-R k}\bigg) \leq 1.
\end{equation}

For the $i$-th node variable,
$\gamma_{ij} \in \RR$ represents the influence from $j$-th static variable and contributes to the deterministic background intensity together with $\nu_i \in \RR_+$; 
parameter $\alpha_{ij} \in \RR$ (or $\beta_{ij}  \in \RR$) represents the magnitude of the influence from the $j$-th node variable (or continuous variable) to the $i$-th node variable, which decays exponentially fast with exponent characterized by $R > 0$ --- those parameter interpretations connect \eqref{eq:notearDAGawkes_model} with the conditional intensity function of the MHP, e.g., \cite{wei2022granger}. Moreover, one advantage of the above model is that, as long as \eqref{eq:notearDAGawkes_positivity} is satisfied, we do not restrict $\alpha_{ij}$ or $\beta_{ij}$ to be non-negative, meaning that our model can handle both triggering and inhibiting effects.

The matrix $A = (\alpha_{ij}) \in \RR^{d_1 \times d_1}$ defines a weighted directed graph $\cG(A) = (\cV,\cE)$ on $d_1$ nodes in the following way: $\cV$ is the collection of aforementioned $d_1$ binary node variables; let $\cA \in \{0,1\}^{d_1 \times d_1}$ such that $\cA_{ij} = 1$ if $\alpha_{ij} \not= 0$ and zero otherwise, then $\cA$ defines the adjacency matrix of a directed graph $\cG(A)$, which gives the collection of directed edges $\cE$; the weights of the directed edges in $\cE$ are defined accordingly by matrix $A$. In a slight abuse of notation, we will {call $A$ the (weighted) adjacency matrix of the graph.}

\subsection{Linear model}\label{sec:linear_link_eg} 

One drawback of MHP comes from its scalability; to be precise, considering complete history leads to quadratic complexity with respect to (w.r.t.) the number of events. Since the triggering (or inhibiting) effects from the history observations decay exponentially fast, we typically consider finite memory depth. Similarly, in our discrete-time Hawkes network, we make reasonable simplification by assuming {\it finite memory depth} $\tau \geq 1$ for both continuous and binary observations. 
More specifically, consider given history at time $t \in [1-\tau:0]$. At time $t \in [T]$, we use $w_{t - \tau : t-1}$ to denote the observations from $t - \tau$ to $t-1$:
$$w_{t - \tau : t-1}  = \big(1,z_1,\dots,z_{d_3},x_{t-1}^{(1)},\dots,x_{t - \tau}^{(1)},\dots,x_{t-1}^{(d_2)},\dots,x_{t - \tau}^{(d_2)},y_{t-1}^{(1)},\dots,y_{t - \tau}^{(1)},\dots,y_{t-1}^{(d_1)},\dots,y_{t - \tau}^{(d_1)}\big)^\T.$$ 
To ease the estimation burden, let $\alpha_{ijk} = \alpha_{ij} \exp\{-R k\}$ and $\beta_{ijk} = \beta_{ij} \exp\{-R k\}$; in fact, this re-parameterization gives our model more flexibility and expressiveness. Now, we can rewrite \eqref{eq:notearDAGawkes_model} as follows: \begin{equation}\label{eq:linear_model}
\begin{split}
        & \mathbb{P}\left(y_t^{(i)}=1 \Big| w_{t - \tau : t-1} \right) = w_{t - \tau : t-1}^\T \theta_i, \\
        & \theta_i \in \Theta = \{\theta: 0 \leq w_{t - \tau : t-1}^\T \theta \leq 1, \ t \in [T]\} \subset \RR^d, 
\end{split}
\end{equation}
where $d = 1+d_3+\tau d_2+\tau d_1$ is the dimensionality, $\Theta$ is the feasible region, and $\theta_i$ is the model parameter:
$$\theta_i = (\nu_i,\gamma_{i1},\dots,\gamma_{id_3},\beta_{i11},\dots,\beta_{i1\tau},\dots,\beta_{id_21},\dots,\beta_{id_2\tau},\alpha_{i11},\dots,\alpha_{i1\tau},\dots,\alpha_{id_11},\dots,\alpha_{id_1\tau})^\T.$$ 
This parameter summarizes the influence from all variables to the $i$-th node. Before we move on, we want to briefly mention that, as a special case of the GLM, \eqref{eq:linear_model} can also be re-parameterized into a causal structural model and its parameters $A_k = (\alpha_{ijk}) \in \mathbb{R}^{d_1 \times d_1}, \ k \in [\tau],$ can be taken as causal graphs under the no unobserved confounder assumption. We will elaborate on these in Section~\ref{subsec:GLM}.

\subsubsection{Estimation} 
We leverage a recently developed technique \cite{juditsky2019signal,juditsky2020convex}, which estimates the model parameters by solving stochastic monotone VI, to develop a statistically principled estimator for discrete-time Hawkes network. To be precise, in our linear model \eqref{eq:linear_model}, for $i \in [d_1]$, we use the weak solution to the following VI as the estimator $\hat \theta_i$:
\begin{equation}
   \text {find } \hat \theta_i \in \Theta:\langle  \bar F_{T}^{(i)}(\theta_i), \theta_i-\hat \theta_i\rangle \geq 0, \ \forall \theta_i \in \Theta,   \label{VI_0}\tag*{{VI}$[\bar F_{T}^{(i)}, \Theta]$}
\end{equation}
where $\bar F_{T}^{(i)}(\theta_i)$ is the empirical vector field defined as follows:
\begin{equation}\label{eq:empirical_vec_field_linear}
    \bar F_{T}^{(i)}(\theta_i) = \frac{1}{T} \sum_{t=1}^T w_{t - \tau : t-1} w_{t - \tau : t-1}^\T \theta_i - \frac{1}{T} \sum_{t=1}^T w_{t - \tau : t-1} y_t^{(i)}=\mathbb{W}_{1:T}\theta_i - \frac{1}{T} \sum_{t=1}^T w_{t - \tau : t-1} y_t^{(i)},
\end{equation}
and
\begin{equation}\label{eq:mat_W}
  \mathbb{W}_{1:T} =\frac{1}{T} \sum_{t=1}^T w_{t - \tau : t-1}w_{t - \tau : t-1}^\T \in \mathbb{R}^{d \times d}.
\end{equation}

\subsubsection{Connection to Least Square (LS) estimator} 

One important observation is that the vector field $\bar F_{T}^{(i)}(\theta_i)$ \eqref{eq:empirical_vec_field_linear} is indeed the gradient field of the Least Square (LS) objective, 
meaning that the weak solution to the corresponding VI solves the following LS problem \cite{juditsky2020convex}:
\begin{equation}\label{eq:objective:convex}
\begin{array}{rl}
\underset{\theta_i}{\mbox{min}} & \frac{1}{2T} \norm{\mathbf{w}_{1:T}^\T \theta_i - Y_{1:T}^{(i)}}_2^2, \\
\mbox{subject to} & \mathbf{0}_T \leq \mathbf{w}_{1:T}^\T \theta_i \leq \mathbf{1}_T,
\end{array}
\end{equation}
where
\begin{equation}\label{eq:concat_data}
  \mathbf{w}_{1:T} = (w_{1-\tau : 0},\dots,w_{T - \tau : T-1}) \in \mathbb{R}^{d \times T}, \quad  Y_{1:T}^{(i)} = (y_1^{(i)},\dots,y_T^{(i)})^\T \in \mathbb{R}^T.
\end{equation} 

One approach to solve \eqref{eq:objective:convex} is to leverage the well-developed convex optimization tools, such as \texttt{CVX} \cite{cvx} and \texttt{Mosek} \cite{mosek}. An alternative approach is through projected gradient descent (PGD), where the empirical vector field \eqref{eq:empirical_vec_field_linear} is treated as the gradient field. To be precise, we introduce dual variables $\eta_1 = (\eta_{1,1},\dots,\eta_{1,T})^\T \in \RR_+^T$, $\eta_2 = (\eta_{2,1},\dots,\eta_{2,T})^\T \in \RR_+^T$ and the Lagrangian is given by:
$$
L(\theta_i, \eta_1, \eta_2) = \frac{1}{2T} \norm{\mathbf{w}_{1:T}^\T \theta_i - Y_{1:T}^{(i)}}_2^2 + \eta_1^\T (\mathbf{w}_{1:T}^\T \theta_i - \mathbf{1}_T) - \eta_2^\T \mathbf{w}_{1:T}^\T \theta_i.
$$
The Lagrangian dual function is $\min_{\theta_i} L(\theta_i, \eta_1, \eta_2)$. As we can see, the Lagrangian above is convex w.r.t. $\theta_i$. By setting $\nabla_{\theta_i} L(\theta_i, \eta_1, \eta_2) = 0$, we have
$$
\hat \theta_i (\eta_1, \eta_2) = \frac{1}{T} \mathbb{W}_{1:T}^{-1} \left(\mathbf{w}_{1:T} Y_{1:T}^{(i)}/T - \eta_1 + \eta_2 \right).
$$
As pointed out in \cite{juditsky2020convex}, $\mathbb{W}_{1:T} \in \mathbb{R}^{d \times d}$ \eqref{eq:mat_W} will be full rank (and thus invertible) with high probability when $T$ is sufficiently large.
By plugging $\hat \theta_i (\eta_1, \eta_2)$ into the Lagrangian, we give the dual problem as follows:
\begin{equation*}
\begin{array}{rl}
\underset{\eta_1, \eta_2}{\mbox{max}} & L \big(\hat \theta_i (\eta_1, \eta_2), \eta_1, \eta_2 \big), \\
\mbox{subject to} & \eta_1, \eta_2 \geq \mathbf{0}_T.
\end{array}
\end{equation*}
This problem can be solved by PGD as its projection step simply changes all negative entries to zeros.

\subsection{Generalized linear model}\label{subsec:GLM}

As mentioned earlier, the linear assumption is restrictive, and therefore, we consider the following GLM \cite{efron2022exponential} to enhance its expressiveness:
\begin{equation}\label{eq:model}
    \mathbb{P}\left(y_t^{(i)}=1 \Big| w_{t - \tau : t-1} \right) = g\left(w_{t - \tau : t-1}^\T \theta_i\right), \quad \theta_i \in \Theta,
\end{equation}
where $g: \mathbb{R} \rightarrow [0,1]$ is a monotone {\it link function}. For example, it can be non-linear, such as sigmoid link function $g(x) = 1/(1+e^{-x})$ on a domain $x \in \RR$ and exponential link function $g(x) = 1 - e^{-x}$ on a domain $x \in \RR_+$; also, it can be linear $g(x) = x$ on a domain $x \in [0,1]$, which reduces our GLM \eqref{eq:model} to the linear model \eqref{eq:linear_model}. The feasible region $\Theta$ will vary based on the choice of link functions, and we will see two examples later in Section~\ref{sec:decoupled_VI_estimate}.

\subsubsection{Structural causal model}\label{sec:causal}
One key feature that distinguishes our discrete-time Hawkes network from the existing black-box method is the causal graph under Pearl's framework \cite{pearl2009causality} encoded in the GLM parameters. To be precise, one can uncover the connection between the GLM \eqref{eq:model} and the stochastic SCM via the Gumbel-Max technique \cite{maddison2017concrete,jang2017categorical}: 
Let us denote 
$$p_{t}^{(i)}(1) = \mathbb{P}\left(y_t^{(i)}=1 \Big| w_{t - \tau : t-1} \right) = g\left(w_{t - \tau : t-1}^\T \theta_i\right), \quad p_{t}^{(i)}(0) = 1 - p_{t}^{(i)}(1). $$
Then, our GLM \eqref{eq:model} can be reformulated into an SCM as follows:
\begin{equation}\label{eq:gumbel_max_SCM}
    y_t^{(i)} = \arg \max_{y \in \{0,1\}}  \left(\log(p_{t}^{(i)}(y)) + \epsilon_t^{(i)} \right),
\end{equation}
where $\epsilon_t^{(i)}$ is a Gumbel r.v., i.e., $\epsilon_t^{(i)} \sim$ Gumbel$(0,1)$. The Gumbel-Max technique tells us that that the SCM \eqref{eq:gumbel_max_SCM} is equivalent to our GLM \eqref{eq:model} in that we still have $\PP(y_t^{(i)} = 1| w_{t - \tau : t-1}) = g\left(w_{t - \tau : t-1}^\T \theta_i\right)$. Therefore, under standard conditions that there is {\it no unobserved confounding}, one can see that the adjacency matrices $A_k = (\alpha_{ijk}) \in \mathbb{R}^{d_1 \times d_1}, \ k \in [\tau],$ represent the causal graph structure over $d_1$ nodes.

\begin{remark}[Connection to Granger causality]\label{rmk:GC}
One may find a very close connection between our causal graph with Granger causality in the non-linear autoregressive model \cite{tank2018neural}; See Appendix~\ref{appendix:causality} for further details on Granger causality. The key difference is whether or not there is unmeasured confounding: Those two causality notions will overlap in a world where there are no potential causes. However, this is not a very likely setting and a fundamentally untestable one \cite{lechner2010relation}. To understand this, the argument that ``Christmas trees sales Granger-cause Christmas'' will not hold once one knows that Christmas took place on December 25th for centuries, which can be modeled as a confounding variable that causes both Christmas tree sales and Christmas itself.
\end{remark}


\subsubsection{Estimation with Variational Inequality}\label{sec:decoupled_VI_estimate} 
Similar to \ref{VI_0} for the linear model, we use the weak solution to the following VI as the estimator for our GLM \eqref{eq:model}:
\begin{equation}
   \text {find } \hat \theta_i \in \Theta:\langle  F_{ T}^{(i)}(\theta_i), \theta_i-\hat \theta_i\rangle \geq 0, \ \forall \theta_i \in \Theta,   \label{VI_1}\tag*{{VI}$[ F_{T}^{(i)}, \Theta]$}
\end{equation}
Parameter $\theta_i$ is constrained in a convex set $\Theta \subset \mathbb{R}^d$, which may vary with different non-linear links; we will see two examples later. The main difference from \ref{VI_0} is the empirical vector field, which is defined as follows:
\begin{equation}\label{eq:empirical_vec_field}
    F_{T}^{(i)}(\theta_i) = \frac{1}{T} \sum_{t=1}^T w_{t - \tau : t-1} \left( g\left(w_{t - \tau : t-1}^\T \theta_i\right) -  y_t^{(i)} \right).
\end{equation}
As we can see, the definition above covers that of $\bar F_{T}^{(i)}$ \eqref{eq:empirical_vec_field_linear} for linear link case; thus, we will use $F_{T}^{(i)}$ to denote the empirical vector field for all monotone links in the following. Furthermore, the statistical inference for each node can be {\it decoupled}, and thus we can perform parallel estimation and simplify the analysis.

The intuition behind this VI-based method is straightforward. Let us consider the global counterpart of the above vector field, whose root is the unknown ground truth $\theta^\star_{i}$, i.e.,
\begin{align*}
    F^{(i)}(\theta_i) = \mathbb{E}_{(w,y^{(i)})} \left[w \left( g\left(w^\T \theta_i\right) -  y^{(i)} \right)\right] 
    = \mathbb{E}_{(w,y^{(i)})} \left[w \left( g\left(w^\T \theta_i\right) -   g\left(w^\T \theta^\star_{i}\right) \right)\right].
\end{align*}
Although we cannot access this global counterpart, by solving the empirical one \ref{VI_1} we could approximate the ground truth very well. We will show how well this approximation can be in Section~\ref{sec:theory}. 

\begin{remark}[Comparison with the original work]
    As a generalization of the VI-based estimator for binary time series \cite{juditsky2020convex}, our method can handle mix-type data (i.e., binary and continuous-valued time series and static variables). Furthermore, we show how to leverage regularization in the VI-based estimation framework as well as extend the performance guarantee to general non-linear monotone link functions, on which we will elaborate in Sections~\ref{sec:DAG_est} and \ref{sec:theory}, respectively.
\end{remark}

\subsubsection{Examples for non-linear link function} 
Now, we will give two examples of general non-linear monotone links and briefly discuss how to numerically obtain our proposed estimator.
Note that the equivalence between our proposed estimator and LS estimator only holds for linear link function since the gradient field of LS objective with general link function will be:
$$ \frac{1}{T} \sum_{t=1}^T \nabla g\left(w_{t - \tau : t-1}^\T \theta_i\right) \left( g\left(w_{t - \tau : t-1}^\T \theta_i\right) -  y_t^{(i)} \right) = \frac{1}{T} \sum_{t=1}^T w_{t - \tau : t-1} g'\left(w_{t - \tau : t-1}^\T \theta_i\right) \left( g\left(w_{t - \tau : t-1}^\T \theta_i\right) -  y_t^{(i)} \right),$$
where $g'$ is the derivative of $g$.
However, in the sigmoid link function case, our proposed estimator reduces to the Maximum Likelihood (ML) estimator for the logistic regression.
To be precise, we can show that the empirical vector filed \eqref{eq:empirical_vec_field} is the gradient field of the objective function of the following ML problem:
\begin{equation*}\label{eq:objective:MLE}
\underset{\theta_i}{\mbox{max}} \quad \frac{1}{T} \sum_{t=1}^T y_t^{(i)} \log  g\left(w_{t - \tau : t-1}^\T \theta_i\right) + (1-y_t^{(i)}) \log \left(1-  g\left(w_{t - \tau : t-1}^\T \theta_i\right)\right).
\end{equation*}
Again, this equivalence between our proposed estimator and ML estimator comes from the fact that $g'(x) = g(x) (1-g(x))$ for the sigmoid link function and does not hold for other non-linear link functions. 
One advantage of the sigmoid link function is that we do not need to put additional constraints on the parameter $\theta_i$ to ensure $g\left(w_{t - \tau : t-1}^\T \theta_i\right)$ is a reasonable probability, i.e., the feasible region is $\Theta = \RR^d$. To numerically obtain such our proposed estimator, we can use vanilla gradient descent (GD), where the gradient is the empirical vector field \eqref{eq:empirical_vec_field}. 

Another non-linear example is the exponential link $g(x) = 1 - e^{-x}, \ x \in \RR_+$. Similar to the linear link case, to ensure valid probability, the feasible region is $\Theta = \{\theta : \ w_{t - \tau : t-1}^\T \theta \geq 0, \ t \in [T]\}$. To numerically solve for our proposed estimator, we can again perform PGD on the Lagrangian dual problem. Alternatively, in many real applications where we have prior knowledge that we do not consider inhibiting effect, i.e., the feasible region is $\theta_i \in \RR_+^d \subset \Theta$, we can perform PGD on the primal problem. 

For general non-linear links, PGD is the most sensible approach to obtain our proposed estimator. However, due the serial correlation in the data, we cannot conduct theoretical convergence analysis as \cite{juditsky2019signal} did. Later in Section~\ref{sec:exp}, we will use numerical simulation to demonstrate the good performance of PGD for all three aforementioned link functions.

\section{Data-Adaptive {Convex} Structural Learning}\label{sec:DAG_est}

In causal structural learning \cite{pearl2009causality}, it is often of great interest to recover a DAG from observational data. In our analysis, we want a DAG-like structure that additionally keeps the lagged self-exciting components, i.e., length-1 cycles. This is because a stronger self-exciting effect informs the adjudicator that the corresponding node/event can last for a longer time once triggered. Therefore, our goal is to remove the less explainable {{\it directed cycles with lengths greater than or equal to two} (referred to as {\it cycles} for brevity)} while keeping lagged self-exciting components to improve the result interpretability.

\subsection{Estimation with data-adaptive linear constraints}\label{sec:DAG_linear_formulation}


\subsubsection{Existing characterizations of acyclicity}\label{sec:DAG_background}

The estimation of a DAG structure is challenging due to the combinatorial nature of the acyclicity constraint. One seminal work \cite{zheng2018dags} characterized the acyclicity constraint via the following continuous and differentiable constraint: We consider memory depth $\tau = 1$ and denote $\alpha_{ij} = \alpha_{ij1}, \ A = (\alpha_{ij})$ for brevity (general $\tau \geq 1$ case will be presented later in this subsection); for non-negative weighted adjacency matrix $A \in A \in \RR_+^{d_1 \times d_1}$, its induced graph is a DAG if and only if
\begin{equation}\label{eq:notearDAG}
    h(A) = \operatorname{tr} (e^A) - d_1 = \sum_{L=1}^\infty \frac{\operatorname{tr}(A^L)}{L!} = 0.
\end{equation}
The above DAG characterization can be understood as follows: For $A \in \RR_+^{d_1 \times d_1}$, $\operatorname{tr}(A^L) \geq 0$, and it will be zero if and only if there does not exist any length-$L$ directed cycle in the induced graph; if $h(A) = 0$, then $\operatorname{tr}(A^L) = 0$ for all $L \geq 1$, implying the induced graph is a DAG.

Intuitively, cycles with length $L \geq d_1$ do not contribute to the DAG characterization, and thus one can truncate the infinite series \eqref{eq:notearDAG} \cite{yu2019dag}. Indeed, one can always apply topological ordering to get a lower triangle adjacency matrix $\tilde{A}$ for a DAG, which is nilpotent such that $\tilde{A}^{d_1} = 0$; such a topological re-order of nodes corresponds to applying permutation matrix $P$ to the original adjacency matrix $A$, i.e., $\tilde{A} = P A P^\T$, and one still has $A^{d_1} = 0$ (since a permutation matrix satisfies $P^{-1} = P^\T$); see Proposition 1 in \cite{zhang2022truncated} for an equivalent characterization of DAG as $A^{d_1} = 0$. Quite contrary to the work by \cite{bello2022dagma} which put emphasis on long cycles, \cite{zhang2022truncated} proposed to truncate the series \eqref{eq:notearDAG} to
\begin{equation}\label{eq:truncDAG}
    h_{\rm trunc}(A) = \sum_{L=1}^k \operatorname{tr}(A^L) = 0,
\end{equation}
where $k < d_1$ since they observed that ``higher-order terms that are close to zero''.

\subsubsection{Motivation}\label{sec:DAG_motivation}

Following \cite{zhang2022truncated}, we propose to apply ``soft'' linear constraint to encourage acyclicity while maintaining the convexity. Specifically, for $L \in [2:k]$, we relax the strict characterization $\operatorname{tr}(A^L) = 0$ by constraining the weighted sum for all possible length-$L$ cycles: for $i_{L} \rightarrow i_{L-1} \rightarrow \cdots \rightarrow i_1 \rightarrow i_{L}$:
\begin{equation}
    \alpha_{i_1 i_2} + \alpha_{i_2 i_3} + \cdots + \alpha_{i_{L-1} i_L} + \alpha_{i_{L} i_1} \leq {{\delta}}.
\end{equation}
Notice that we do not put a constraint on $L=1$ case since the self-exciting effects carry meaning and are desirable in our analysis.

\begin{figure*}[!htp]
\centerline{
\includegraphics[width = .3\textwidth]{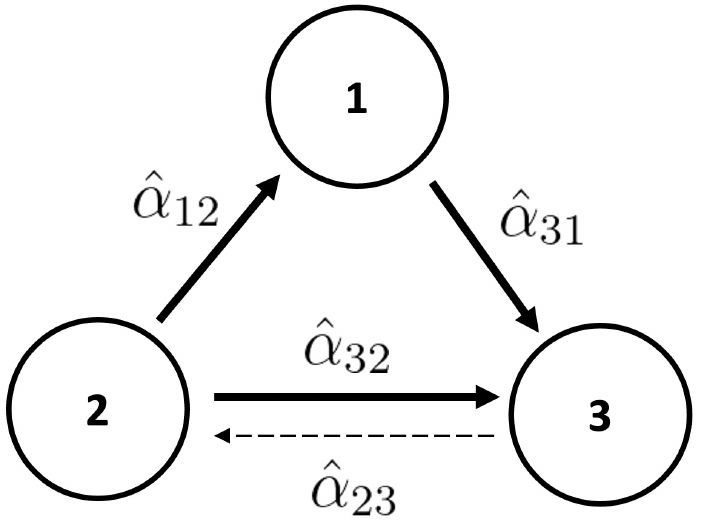} 
}
\caption{{Illustration of the estimated graph without regularization. The existence of an edge represents the corresponding estimated weight is greater than zero and dashed edge $ 3 \rightarrow 2$ indicates that its weight is very small, which could be a result of noisy observations.}}
\label{fig:method_illus}
\end{figure*}

One simple estimation method would be to include the above linear constraints into the feasible region, which will not change the convexity since the intersection of two convex sets is still convex. However, the number of linear constraints will be on the order of $d_1^k$, and the constraint hyperparameter ${{\delta}} \geq 0$ should also vary for different length-$L$ cycles, depending on the ground truth weight parameters in the corresponding cycle.
Fortunately, due to the consistency result (to be presented in the next section), we can address the above issues by obtaining data-adaptive linear constraints. To be precise, as illustrated in Figure~\ref{fig:method_illus}, 
the recovery guarantee implies that the existence of edge $ 3 \rightarrow 2$ might be a result of noise in the observations, and the VI solution tends to output such a false discovery. Therefore, one can simply add the following data-adaptive constraints 
\[\alpha_{23} + \alpha_{32} \leq \hat \alpha_{32}, \quad \alpha_{12} + \alpha_{23} + \alpha_{31} \leq \hat \alpha_{12} + \hat \alpha_{31},\]
into the feasible region when solving the VI. Moreover, in the above illustrative example, oftentimes imposing the first constraint suffices to remove edge $3 \rightarrow 2$, meaning that removing short cycles suffices to remove long cycles; in both our simulation and real experiments, we have demonstrated that it suffices to consider $k=3$ in moderate-sized graph (around $20$ nodes) setting; indeed, we only discover short cycles in our real data example (see Figure~\ref{fig:GC_graph_bp}). We will formally state our method with $k=3$ in the following.


\subsubsection{Proposed constraint}
Consider the causal graphs induced by the estimated adjacency matrices $\hat A_{\ell} = (\hat \alpha_{ij \ell}) \in \RR^{d_1 \times d_1}, \ell \in [\tau]$, using the VI-based estimator \ref{VI_1}. As mentioned earlier, cycles in those graphs are undesirable, and we want to remove them. Let us begin with formally defining cycles: for positive integer $L \geq 2$, if there exist $\ell \in [\tau]$ and mutually different indices $i_1, \dots, i_L \in [d_1]$ such that 
$$\hat \alpha_{i_{1} i_{L} \ell} > 0, \quad \hat \alpha_{i_{k+1} i_{k} \ell} > 0, \quad k \in [L-1],$$
then we say there exists a length-$L$ (directed) cycle in the directed graphs induced by $\hat A_\ell$'s. 

We consider all possible length-2 and length-3 cycles in those graphs, whose indices are as follows:
\begin{align*}
    I_{2,\ell} &= \{(i,j): i \not= j, \ \hat \alpha_{i j \ell} > 0, \ \hat \alpha_{j i \ell} > 0\}, \quad \ell \in [\tau], \\ 
    I_{3,\ell} &= \{(i,j, k): i,j,k \text{ mutually different}, \ \hat \alpha_{i j \ell} > 0, \ \hat \alpha_{j k \ell} > 0, \ \hat \alpha_{k i \ell} > 0\}, \quad \ell \in [\tau].
\end{align*}
{As illustrated in Figure~\ref{fig:method_illus}}, in each cycle, the edge with the least weight could be caused by noisy observation, meaning that we should remove such edges to eliminate the corresponding cycle. To do so, we impose the following {\it data-adaptive linear cycle elimination constraints} to shrink the weights of those ``least important edges'':
\begin{equation}\label{eq:linear_constraint}
    \begin{split}
\alpha_{i j \ell} + \alpha_{j i \ell} &\leq \delta_{2,\ell}(i,j), \quad (i,j) \in I_{2, \ell}, \quad \ell \in [\tau], \\
     \alpha_{i j \ell} + \alpha_{j k \ell} + \alpha_{k i \ell} &\leq  \delta_{3,\ell}(i,j,k), \quad (i,j,k) \in I_{3, \ell}, \quad \ell \in [\tau],
    \end{split}
\end{equation}
where the {\it data-adaptive regularization strength parameters} $\delta_{2,\ell}(i,j), \delta_{3,\ell}(i,j,k)$ are defined as follows:
\begin{equation}\label{eq:linear_constraint_parameter}
    \begin{split}
        \delta_{2,\ell}(i,j) &= \hat \alpha_{i j \ell} + \hat \alpha_{j i \ell} - \min\{\hat \alpha_{i j \ell}, \hat \alpha_{j i \ell}\} = \max\{\hat \alpha_{i j \ell},\hat \alpha_{j i \ell} \}, \quad (i,j) \in I_{2, \ell}, \quad \ell \in [\tau], \\
    \delta_{3,\ell}(i,j,k) &= \hat \alpha_{i j \ell} + \hat \alpha_{j k \ell} + \hat \alpha_{k i \ell} - \min\{\hat \alpha_{i j \ell}, \hat \alpha_{j k \ell}, \hat \alpha_{k i \ell}\}, \quad (i,j,k) \in I_{3, \ell}, \quad \ell \in [\tau].
    \end{split}
\end{equation}

\subsubsection{Constrained joint VI-based estimation}
Different from the aforementioned decoupled learning approach in Section~\ref{sec:decoupled_VI_estimate}, here we need to estimate parameters $\theta_1,\dots,\theta_{d_1}$ jointly to remove cycles and encourage our desired DAG structure. We concatenate the parameter vectors into a matrix, {i.e., \(\theta = (\theta_1,\dots,\theta_{d_1}) \in \RR^{d \times d_1},\)} and the feasible region of the concatenated parameter is then defined as:
\begin{equation}\label{eq:joint_feasible_region}
    \tilde \Theta = \{\theta = (\theta_1,\dots,\theta_{d_1}): \theta_i \in \Theta, \ i \in [d_1]\}.
\end{equation}
The joint estimator coupled with the data-adaptive linear cycle elimination constraint is defined as the weak solution to the following VI:
\begin{equation}
   \text {find } \hat \theta \in \Theta^{\rm DAL}: \langle \operatorname{vec}(F_{T}(\theta)), \operatorname{vec}(\theta-\hat \theta) \rangle \geq 0, \quad \forall \theta \in \Theta^{\rm DAL}, \label{VI_2}\tag*{{VI}$[ F_{T}, \Theta^{\rm DAL}]$}
\end{equation}
where $\operatorname{vec}(A)$ is the vector of columns of $A$ stacked one under the other, the empirical ``vector'' field is
\begin{equation}\label{eq:empirical_VI_whole}
    F_{T}(\theta) = (F_{T}^{(1)}(\theta_1),\dots,F_{T}^{(d_1)}(\theta_{d_1})) \in \RR^{d \times d_1},
\end{equation}
and vector field $F_{T}^{(i)}(\theta_i) \in \RR^{d}$ is defined in \eqref{eq:empirical_vec_field}.
Moreover, the convex set $\Theta^{\rm DAL}$ incorporates the above data-adaptive linear constraints \eqref{eq:linear_constraint} and is defined as follows:
\begin{align*}
    \Theta^{\rm DAL} = \Big\{\theta : \theta \in \tilde \Theta, \ & e_{f_{j,\ell},d}^\T \theta e_{i,d_1} + e_{f_{i,\ell},d}^\T \theta e_{j,d_1} \leq \delta_{2,\ell}(i,j), \ (i,j) \in I_{2, \ell}, \ \ell \in [\tau], \\
     & e_{f_{j,\ell},d}^\T \theta e_{i,d_1} + e_{f_{k,\ell},d}^\T \theta e_{j,d_1} + e_{f_{i,\ell},d}^\T \theta e_{k,d_1} \leq  \delta_{3,\ell}(i,j,k), \ (i,j,k) \in I_{3, \ell}, \ \ell \in [\tau]\Big\},
\end{align*}
where regularization strength parameters $\delta_{2,\ell}(i,j), \delta_{3,\ell}(i,j,k)$ are defined in \eqref{eq:linear_constraint_parameter} and $f_{j,\ell} = 1+d_3+\tau d_2 + (j-1)\tau + \ell$ such that $e_{f_{j,\ell},d}^\T \theta e_{i,d_1} = \alpha_{i j \ell}$.

\subsubsection{A special case: linear link function}\label{subsec:linear_DAG_penalized}
Now, we elaborate on our proposed regularization on a special linear link case. The vector field $F_T(\theta)$ \eqref{eq:empirical_VI_whole} can be expressed as follows:
$$F_{T}(\theta) = \frac{1}{T} \mathbf{w}_{1:T} \mathbf{w}_{1:T}^\T \theta - \frac{1}{T} \mathbf{w}_{1:T} Y = \mathbb{W}_{1:T}\theta - \frac{1}{T} \mathbf{w}_{1:T} Y,$$
where \(Y = (Y_{1:T}^{(1)},\dots,Y_{1:T}^{(d_1)}) \in \RR^{T \times d_1}\) and $Y_{1:T}^{(i)}, \ \mathbf{w}_{1:T} \in \RR^{d \times T}$ are defined in \eqref{eq:concat_data}. Similar to the linear model example in Section~\ref{sec:linear_link_eg}, the above vector field is the gradient field of the least square objective, and our proposed estimator \ref{VI_2} boils down to the LS estimator, which solves the following constrained optimization problem:
\begin{equation}\label{eq:LS_DAG}
\begin{split}
    \underset{\theta}{\mbox{min}} \quad & \frac{1}{2T} \sum_{i=1}^{d_1} \norm{\mathbf{w}_{1:T}^\T \theta_i - Y_{i,1:T}}_2^2 = \frac{1}{2T} \norm{\mathbf{w}_{1:T}^\T \theta - Y}_F^2, \\
\mbox{subject to} \quad & \mathbf{0}_T \leq \mathbf{w}_{1:T}^\T \theta_i \leq \mathbf{1}_T, \ i \in [d_1], \\
& e_{f_{j,\ell},d}^\T \theta e_{i,d_1} + e_{f_{i,\ell},d}^\T \theta e_{j,d_1} \leq \delta_{2,\ell}(i,j), \ (i,j) \in I_{2, \ell}, \ \ell \in [\tau], \\
     &e_{f_{j,\ell},d}^\T \theta e_{i,d_1} + e_{f_{k,\ell},d}^\T \theta e_{j,d_1} + e_{f_{i,\ell},d}^\T \theta e_{k,d_1} \leq  \delta_{3,\ell}(i,j,k), \ (i,j,k) \in I_{3, \ell}, \ \ell \in [\tau].
\end{split}
\end{equation}
Similarly, \eqref{eq:LS_DAG} is convex and can be efficiently solved by a well-developed toolkit such as \texttt{Mosek}. 

Most applications, including our motivating example, only consider triggering effect, meaning that one can replace $\mathbf{w}_{1:T}^\T \theta_i \geq \mathbf{0}_T$ with $\theta_i \geq \mathbf{0}_d$ as a relaxation. In addition, since the prediction of the $i$-th event's occurrence at time $t$ is by comparing the estimated probability $w_{t - \tau:t-1}^\T \theta_i$ with a threshold selected using the validation dataset, we can further relax the constraint $\mathbf{w}_{1:T}^\T \theta_i \leq \mathbf{1}_T$ and treat $w_{t - \tau:t-1}^\T \theta_i$ as a ``score'' instead of a probability. Thus, we can adopt the following penalized form:
\begin{equation}\label{eq:LS_DAG_penalized_form}
\begin{split}
\underset{\theta}{\mbox{min}} \quad & \frac{1}{2T} \norm{\mathbf{w}_{1:T}^\T \theta - Y}_F^2 \ +   \sum_{\ell=1}^\tau  \sum_{(i,j) \in I_{2, \ell}} 
\frac{\lambda}{\delta_{2,\ell}(i,j)} (e_{f_{j,\ell},d}^\T \theta e_{i,d_1} + e_{f_{i,\ell},d}^\T \theta e_{j,d_1}) \\
      & \quad\quad\quad\quad\quad\quad\quad\  + \sum_{\ell=1}^\tau  \sum_{(i,j,k) \in I_{3, \ell}} 
\frac{\lambda}{\delta_{3,\ell}(i,j,k)} (e_{f_{j,\ell},d}^\T \theta e_{i,d_1} + e_{f_{k,\ell},d}^\T \theta e_{j,d_1} + e_{f_{i,\ell},d}^\T \theta e_{k,d_1}), \\
\mbox{subject to} \quad & \theta_i \geq \mathbf{0}_T, \ i \in [d_1],
\end{split}
\end{equation}
where $\lambda$ is a hyperparameter that controls the strength of regularization. The data-adaptive regularization strength parameters $\delta_{2,\ell}(i,j), \delta_{3,\ell}(i,j,k)$ appear in the denominator since smaller $\delta_{2,\ell}(i,j), \delta_{3,\ell}(i,j,k)$ imply stronger penalty, which closely resembles adaptive Lasso \cite{zou2006adaptive}. Most importantly, \eqref{eq:LS_DAG_penalized_form} can be solved efficiently using PGD, where at each iteration, the update rule is as follows:
\begin{equation}\label{eq:penalized_PGD_update}
\begin{split}
    \hat \theta \leftarrow \hat \theta - \eta \Bigg(F_{T}(\hat \theta) \ +  &  \sum_{\ell=1}^\tau  \sum_{(i,j) \in I_{2, \ell}} 
\frac{\lambda}{\delta_{2,\ell}(i,j)} (e_{f_{j,\ell},d} e_{i,d_1}^\T + e_{f_{i,\ell},d} e_{j,d_1}^\T) \\
      + & \sum_{\ell=1}^\tau  \sum_{(i,j,k) \in I_{3, \ell}} 
\frac{\lambda}{\delta_{3,\ell}(i,j,k)} (e_{f_{j,\ell},d} e_{i,d_1}^\T + e_{f_{k,\ell},d} e_{j,d_1}^\T + e_{f_{i,\ell},d} e_{k,d_1}^\T)\Bigg),
\end{split}
\end{equation}
where $\eta$ is the step size/learning rate hyperparameter and empirical field $F_{T}(\cdot)$ is given in \eqref{eq:empirical_VI_whole}. After the above update in each iteration, the projection onto the feasible region $\RR_+^{d \times d_1}$ can be simply done by replacing all negative entries in $\hat \theta$ with zeros.

\subsection{Penalized joint VI-based estimation}
As previously discussed in Section~\ref{sec:decoupled_VI_estimate}, the \ref{VI_1} can be solved by PGD as the feasible region $\Theta$ is a convex set. 
However, \ref{VI_2} additionally incorporates the data-adaptive linear constraints into its feasible region $\Theta^{\rm DAL}$ to encourage a DAG structure with desired lagged self-exciting components, making the projection step harder to implement. Alternatively, it will be much easier if we can transfer the constraints into the penalty. Inspired by the penalized form for the linear link special case \eqref{eq:LS_DAG_penalized_form} (which is very similar to adaptive Lasso \cite{zou2006adaptive}), we propose a {\it data-adaptive linear penalized VI-based estimator}, which is the weak solution to the following VI:
\begin{equation}
   \text {find } \hat \theta \in \tilde \Theta: \langle \operatorname{vec}(F_{T}^{\rm DAL}(\theta)), \operatorname{vec}(\theta-\hat \theta) \rangle \geq 0, \quad \forall \theta \in \tilde \Theta, \label{VI_3}\tag*{{VI}$[ F_{T}^{\rm DAL}, \tilde \Theta]$}
\end{equation}
where the feasible region $\tilde \Theta$ is defined in \eqref{eq:joint_feasible_region} and the {\it data-adaptive linear penalized vector filed} $F_{T}^{\rm DAL}(\theta)$ is defined as follows:
\begin{equation}\label{eq:empirical_VI_whole_AL}
\begin{split}
    F_{T}^{\rm DAL}(\theta) = F_{T}(\theta) \ + &  \sum_{\ell=1}^\tau  \sum_{(i,j) \in I_{2, \ell}} \frac{\lambda}{\delta_{2,\ell}(i,j)} (e_{f_{j,\ell},d} e_{i,d_1}^\T + e_{f_{i,\ell},d} e_{j,d_1}^\T) \\
    + & \sum_{\ell=1}^\tau  \sum_{(i,j,k) \in I_{3, \ell}} 
\frac{\lambda}{\delta_{3,\ell}(i,j,k)} (e_{f_{j,\ell},d} e_{i,d_1}^\T + e_{f_{k,\ell},d} e_{j,d_1}^\T + e_{f_{i,\ell},d} e_{k,d_1}^\T).
\end{split}
\end{equation}
Here, $\lambda$ is a tunable penalty strength hyperparameter, $F_{T}(\theta) = (F_{T}^{(1)}(\theta_1),\dots,F_{T}^{(d_1)}(\theta_{d_1})) \in \RR^{d \times d_1}$ is the concatenated field \eqref{eq:empirical_VI_whole}
and vector field $F_{T}^{(i)}(\theta_i) \in \RR^{d}$ is defined in \eqref{eq:empirical_vec_field}. Compared with \ref{VI_2}, it is much easier to solve \ref{VI_3} using PGD. For example, in the exponential link function case, if we restrict our consideration to triggering effect only, we can use \eqref{eq:penalized_PGD_update} as the update rule in PGD and zero out all negative entries after each update as the projection step in each iteration.

\begin{remark}
One advantage of our data-adaptive linear regularization is its flexibility, and it is the user's choice to decide which potential cycle should be included in the constraint. {Please refer to our work \cite{wei2023causal} for more numerical examples of recovering strict DAGs (i.e., without any lagged self-exciting components) by additionally including length-1 cycles in our data-adaptive linear constraints.} 
\end{remark}

\begin{remark}
    The above idea to transfer constraint into a penalty by adding the penalty's derivative to the empirical vector field opens up possibilities to consider various types of penalties to encourage desired structures when using our proposed VI-based estimator, e.g., the continuous DAG characterization \cite{zheng2018dags} and the adaptive Lasso \cite{zou2006adaptive}; one can see Section~\ref{sec:DAG_baseline_formulation} below for more details on our proposed VI-based estimator coupled with DAG regularization \eqref{eq:empirical_VI_whole_DAG} and $\ell_1$ regularization \eqref{eq:empirical_VI_whole_l1}.
\end{remark}

\section{Non-asymptotic Performance Guarantee}\label{sec:theory}
In this section, we will show our proposed estimator has nice statistical properties, i.e., it is unique and consistent; the proof is deferred to Appendix~\ref{appendix:proof} due to space consideration. In addition, we will also derive a linear program (LP) based confidence interval (CI) of parameters $\theta_i$'s, which we defer to Appendix~\ref{appendix:CI}.
One pitfall of our theoretical analysis is the lack of guarantee for the proposed data-adaptive linear method and we leave this topic for future discussion. We begin with two necessary model assumptions:

\begin{assumption}\label{assumption:vector_field}
The link function $g(\cdot)$ is continuous and monotone, and the vector field $G(\theta) = \mathbb{E}_w[wg(w^\T\theta)]$ is well defined (and therefore monotone along with $g$). Moreover, $g$ is differentiable and has uniformly bounded first order derivative $m_g \leq |g'|\leq M_g$ for $0<m_g\leq M_g$.
\end{assumption}

\begin{assumption}\label{assumption:observation}
The observations (static, binary, and continuous) are bounded almost surely: there exists $M_w>0$ such that at any time step $t$, we have $\norm{w_{t - \tau : t-1}}_\infty \leq M_w$ with probability one.
\end{assumption}

\begin{theorem}[Upper bound on $\norm{\hat \theta_i - \theta^\star_{i}}_2$]\label{thm:upper_err_bound}
Under Assumptions~\ref{assumption:vector_field} and \ref{assumption:observation}, for $i \in [d_1]$ and any $\varepsilon \in (0,1)$, with probability at least $1-\varepsilon$,
the $\ell_2$ distance between ground truth $\theta^\star_{i}$ and the weak solution $\hat \theta_i$ to \ref{VI_1} can be upper bounded as follows:
$$\norm{\hat \theta_i -  \theta^\star_{i}}_2 \leq \frac{M_w}{m_g \lambda_1} \sqrt{\frac{d\log (2d/\varepsilon)}{T }},$$
where $\lambda_1$ is the smallest eigenvalue of $\mathbb{W}_{1:T} = \sum_{t=1}^T w_{t - \tau : t-1}w_{t - \tau : t-1}^\T/T$ \eqref{eq:mat_W}.
\end{theorem}

The above theorem is an extension to the general link function case with mixed-type data of Theorem 1 \cite{juditsky2020convex}.
As pointed out in \cite{juditsky2020convex}, $\mathbb{W}_{1:T} \in \mathbb{R}^{d \times d}$ will be full rank for sufficiently large $T$, i.e., $\lambda_1$ will be a positive constant with high probability.

\begin{remark}[Identifiablility]
    The uniqueness, or rather, the identifiability, comes from the nice property of the underlying vector field. To be precise, in the proof of the above theorem (see Appendix~\ref{appendix:proof}), we have shown the vector field $F_{T}^{(i)}(\theta_i)$ is monotone modulus $m_g \lambda_1$ under Assumption~\ref{assumption:vector_field}. Then, the following lemma tells us that our proposed estimator is unique:
\begin{lemma}[Lemma 3.1 \cite{juditsky2019signal}]
Let $\Theta$ be a convex compact set and $H$ be a monotone vector field on $\Theta$ with monotonicity modulus $\kappa>0$, i.e.,
$$\forall \ z, z' \in \Theta, [H(z) - H(z')]^\T(z-z') \geq \kappa \norm{z-z'}_2^2.$$
Then, the weak solution $\Bar{z}$ to VI$[H,\Theta]$ exists and is unique. It satisfies:
$$H(z)^\T(z-\Bar{z}) \geq \kappa \norm{z-\Bar{z}}_2^2.$$
\end{lemma}
\end{remark}

Next, we will use both simulations and real examples to show the good performance of our method for causal structural learning.

\section{Numerical Simulation}\label{sec:DAG_baseline_formulation}

In this section, we conduct numerical simulations to show the good performance of the VI-based estimator \ref{VI_1}. We will show the competitive performance of our VI-based method compared with benchmark methods such as the Neural Network based method \cite{khanna2019economy}, even under the model mis-specification setting. Importantly, we also show that our proposed data-adaptive linear regularization outperforms other DAG-inducing regularization approaches in structural learning. 
{Due to space consideration, the complete experimental configurations and the comparison between the VI-based estimator and  under the model misspecification setting is deferred to Appendices~\ref{appendix:num_simu_set} and \ref{sec:exp}.}

\subsection{Evaluation metrics}\label{sec:simulation_metrics} 
{Our simulations consider a simple $\tau = 1$ case.} We are interested in the estimation of model parameters: (i) background intensity $\nu = (\nu_1,\dots,\nu_{d_1})^\T$ and (ii) self- and mutual-exciting matrix $A_1 = (\alpha_{ij1})$; for brevity, we drop the last subscript ``$1$'' and denote the adjacency matrix by $A = (\alpha_{ij})$. We consider (i) the $\ell_2$ norm of the background intensity estimation error $\norm{\hat \nu - \nu}_2$ ($\nu$ err.) and (ii) matrix $F$-norm of the self- and mutual-exciting matrix estimation error $\norm{\hat A - A}_F$ ($A$ err.).
Additionally, we report the Structural Hamming Distance (SHD) between $\hat A$ and $A$, which reflects how close the recovered graph is to the ground truth. {This is the primary quantitative metric in the following experiment.} {SHD is the number of edge flips, insertions, and deletions in order to transform between two graphs. In particular, when edge $i \rightarrow j$ is in the true graph, i.e., $\alpha_{ji} > 0$, whereas edge $i \leftarrow j$ is in the estimated graph, i.e., $\hat \alpha_{ij} > 0$, the SHD is increased by $1$ via edge flip instead of $2$ by edge insertion and deletion.} In addition, since we are interested in DAG structure with self-exciting components, we also consider a measure of ``DAG-ness'' on the recovered adjacency matrix (after zeroing out the diagonal entries of $\hat A$), {denoted by $h(A_0)$ \eqref{eq:notearDAG}.}
We need to mention that small $h(A_0)$ with large SHD means we recover a DAG which is not close to the ground truth and this does not imply good structure recovery.

\subsection{Benchmark regularization approaches}
Let us first formally introduce several benchmark regularization approaches. {Recall that we use $A = (\alpha_{ij})$ to denote $A_1 = (\alpha_{ij1})$ in $\tau = 1$ case for brevity:}

\begin{itemize}
\item {\it Continuous DAG regularization and a proposed variant.}
{Recall the continuous and differentiable (but not convex) characterization by \cite{zheng2018dags} in \eqref{eq:notearDAG}, which can measure the DAG-ness of $A$}. Most importantly, this DAG characterization has closed from derivative, i.e., \(\nabla h(A)=\left(e^{A}\right)^{\T}.\)
Inspired by \cite{ng2020role}, we use this characterization as a penalty directly. We take advantage of its differentiability and add its derivative to the concatenated field $F_T(\theta)$ \eqref{eq:empirical_VI_whole}, which will be treated as the gradient field in PGD. More specifically, let $J = (\mathbf{0}_{d_1}, I_{d_1}) \in \RR^{d_1 \times d}$ and we will have $J \theta = A^\T$. Then, the vector field coupled with DAG regularization $F_{T}^{\rm DAG}(\cdot)$ is defined as follows:
\begin{equation}\label{eq:empirical_VI_whole_DAG}
    F_{T}^{\rm DAG}(\theta) = F_{T}(\theta) \ +  \lambda J^\T \nabla h(J \theta) = F_{T}( \theta) \ +  \lambda J^\T e^{ A},
\end{equation}
where tunable hyperparameter $\lambda$ controls the penalty strength. The PGD update rule is given by:
\begin{equation*}
\begin{split}
    \hat \theta \leftarrow \hat \theta - \eta F_{T}^{\rm DAG}(\hat \theta),
\end{split}
\end{equation*}
where $\eta$ is the learning rate hyperparameter and is also tunable. 

One drawback of the aforementioned DAG regularization is that it removes not only cycles but also lagged self-exciting components; this is evidenced in Figure~\ref{fig3:penalty_compare}. 
To keep those informative lagged self-exciting components, we simply zero out the diagonal elements in DAG regularization derivative $\nabla h(J \theta)$ in \eqref{eq:empirical_VI_whole_DAG}. Thus, the PGD update will not shrink the diagonal elements.
\item
{\it $\ell_1$ regularization and adaptive Lasso.}
We adopt the $\ell_1$ penalty as another benchmark, which encourages a sparse graph structure and, in turn, eliminates cycles. The $\ell_1$ penalized vector filed is defined as follows:
\begin{equation}\label{eq:empirical_VI_whole_l1}
    F_{T}^{\ell_1}(\theta) = F_{T}(\theta) +  \lambda J^\T \nabla (|J \theta|_1),
\end{equation}
where $|\cdot|_1$ is the summation of the absolute values of all entries.
Similarly, the VI-based estimator can be efficiently solved by PGD using the following update rule: \begin{equation*}
\begin{split}
    \hat \theta \leftarrow \hat \theta - \eta F_{T}^{\ell_1}(\hat \theta).
\end{split}
\end{equation*}

As a variant of $\ell_1$ regularization, adaptive $\ell_1$ regularization (or adaptive Lasso \cite{zou2006adaptive}) replaces $\lambda |\alpha_{ij}|$ with $\frac{\lambda}{\hat \alpha_{ij}}|\alpha_{ij}|$ in \eqref{eq:empirical_VI_whole_l1}; for $\hat \alpha_{ij} = 0$ case, we adopt a simple remedy by adding penalty term $10^3 \lambda |\alpha_{ij}|$ to enforce $\alpha_{ij}$ to be zero.
\end{itemize}

\subsection{Results}
\begin{figure*}[!htp]
\centering
{\includegraphics[width = \textwidth]{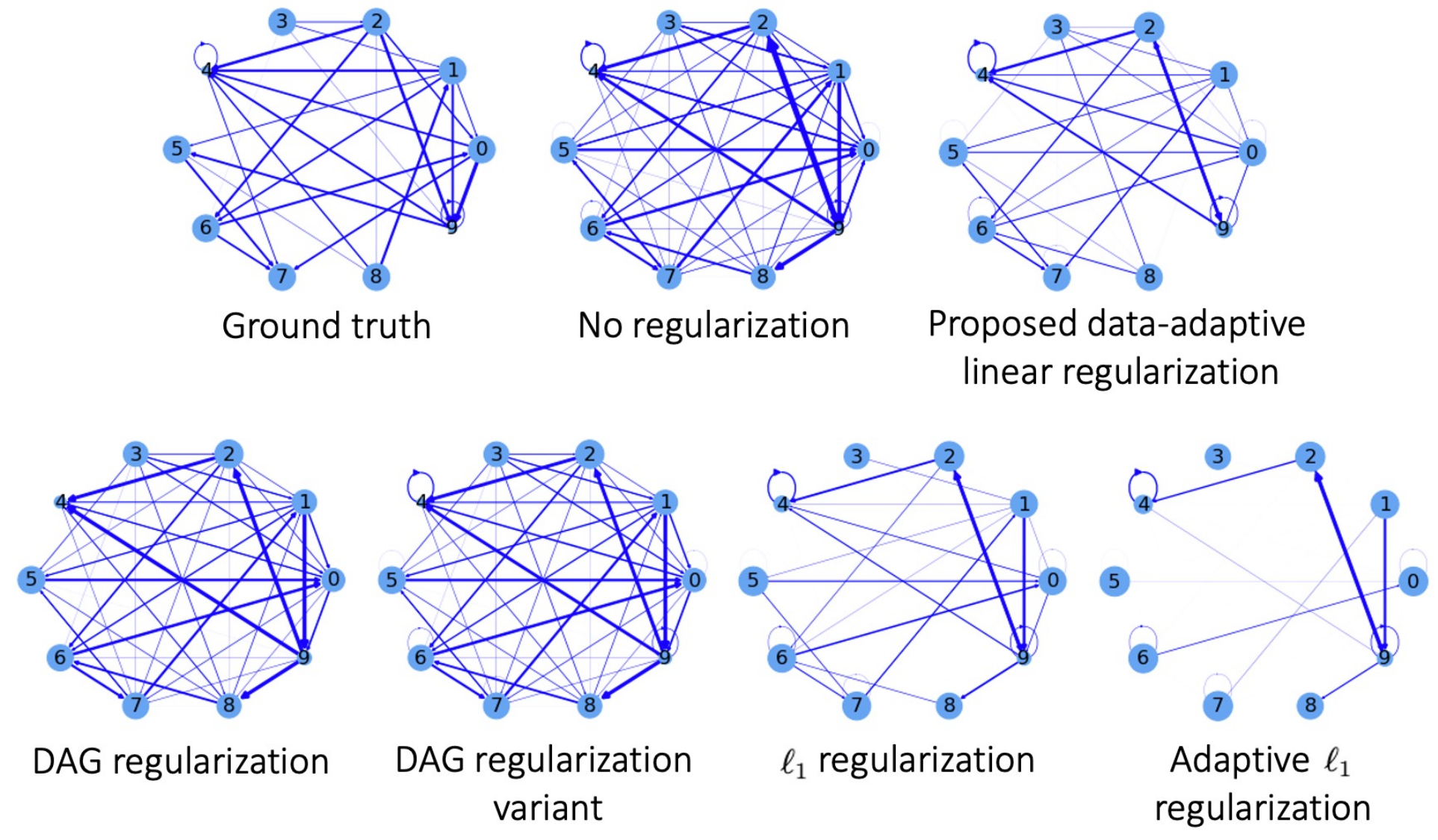}}
\caption{Simulated example: demonstration of the effectiveness of our proposed data-adaptive linear regularization. We visualize the recovered graph structures for a $d_1 = 10$ and $T = 500$ illustrative example using our proposed VI-based estimator coupled with various types of regularization (specified on top of each panel). We can observe that our proposed VI-based estimator coupled with our proposed data-adaptive linear constraint can return the closest graph structure to the ground truth; see quantitative evaluation metrics, such as SHD, in Table~\ref{table:exp3_penalty_compare}. }\label{fig3:penalty_compare}
\end{figure*}

We first demonstrate the competitive performance of our proposed data-adaptive linear method on a $d_1 = 10$ illustrative example, where we adopt an exponential link function. We visualize the recovered graphs using our VI-based estimator with exponential link coupled with various types of regularization in Figure~\ref{fig3:penalty_compare}, and we report all aforementioned quantitative metrics in Table~\ref{table:exp3_penalty_compare}; {additionally, we report the relative errors in the illustrative example in Table~\ref{table:exp3_penalty_compare_relative}.} We observe that our proposed data-adaptive linear regularization can achieve the best weight recovery accuracy (in terms of $\nu.$ err. and $A$ err.) and structure recovery accuracy (in terms of SHD) compared with all benchmark methods. 

\begin{table}[htp]
\caption{Simulated example: quantitative metrics of the example in Figure~\ref{fig3:penalty_compare}. We can observe that our proposed VI-based estimator, coupled with our proposed data-adaptive linear constraint, can achieve better estimation accuracy while encouraging a desired DAG structure. Besides, our proposed method also gives the best structure recovery, i.e., the smallest SHD; although the adaptive $\ell_1$ approach achieves the best DAG-ness, it achieves so by removing many important edges and cannot output a correct graph structure (as evidenced in Figure~\ref{fig3:penalty_compare}).  }\label{table:exp3_penalty_compare}
\begin{center}
\begin{small}

\resizebox{.8\textwidth}{!}{%
\begin{tabular}{lcccccc}
\toprule[1pt]\midrule[0.3pt]
Regularization & None  & Proposed & DAG & DAG-Variant & $\ell_1$ & Ada. $\ell_1$ \\
\cmidrule(l){2-7}
$A$ err. & .3874 & \textbf{.2094} & .3541 & .2949 & {\it .2501} & .3022 \\
$\nu$ err. & .1175 & \textbf{.0775} & .0895 & {\it .0841}  & .0884 & .1251 \\
$h(A_0)$ & .1223 & .0308 & .0337 & {\it .0242}  & .0274 & \textbf{.0232}  \\
SHD & 41 & \textbf{25} & 32 & 34  & 41 & {\it 29} \\
\midrule[0.3pt]\bottomrule[1pt]
\end{tabular}
}

\end{small}
\end{center}
\end{table}

\begin{figure*}[!htp]
\centering
{\includegraphics[width = .8\textwidth]{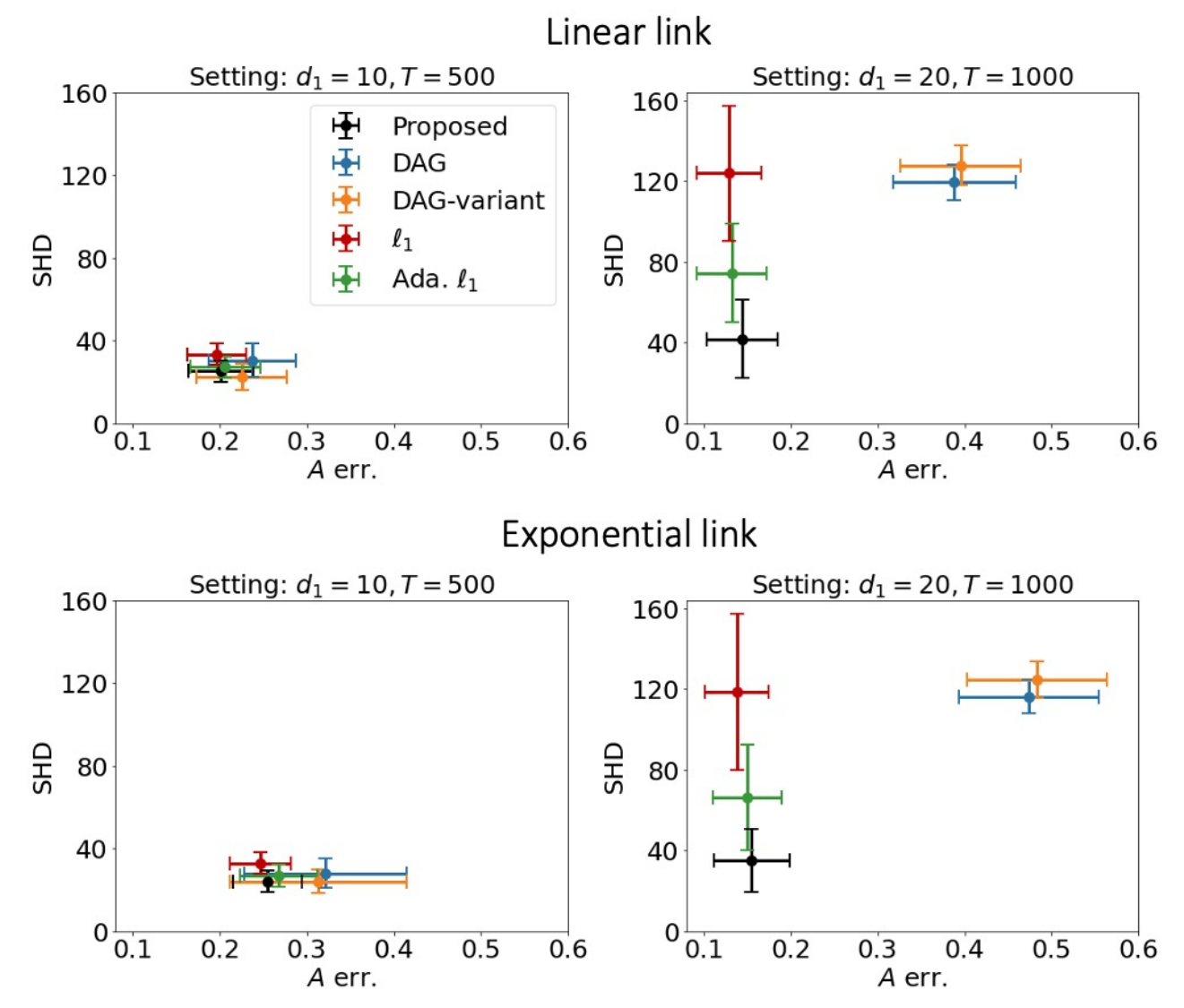}}
\caption{Simulation: mean (dot) and standard deviation (error bar) of matrix $F$-norm of the self- and mutual-exciting matrix estimation error ($A$ err.) and Structural Hamming Distance (SHD) over $100$ independent trials for various types of regularization. For each regularization, the closer it is to the origin, the better it is. We can observe that our proposed data-adaptive linear regularization performs the best (especially in higher dimensional cases) in terms of structure recovery while achieving almost the same weight recovery accuracy with the best result.}\label{fig3:penalty_compare_repeat}
\end{figure*}

To further validate the good performance of our proposed data-adaptive linear method, we run $100$ independent trials for $d_1 = 10, T = 500$ and $d_1 = 20, T = 1000$ cases as well as linear link and exponential link functions cases. We plot the mean and standard deviation of $A$ err. and SHD in Figure~\ref{fig3:penalty_compare_repeat}; for completeness, we also report the raw values of the mean and standard deviation of all four aforementioned metrics in Table~\ref{table:exp2_penalty_comparison}.
{The complete details, including random DAG generation, PGD to solve for the estimators, and additional results (Tables~\ref{table:exp3_penalty_compare_relative} and \ref{table:exp2_penalty_comparison}), are deferred to Appendix~\ref{appendix:num_simu_set_exp3}}.

Figure~\ref{fig3:penalty_compare_repeat} shows that, in low dimensional (i.e., $d_1 = 10$) case, $\ell_1$ regularization does well in weight recovery but fails in structure recovery, whereas our proposed variant of DAG regularization prioritizes the structure recovery but performs poorly in weight recovery.
As a comparison, our proposed data-adaptive linear regularization achieves comparable weight and structure recovery accuracy to $\ell_1$ regularization and the proposed variant of DAG regularization, respectively, suggesting that it can balance the weight recovery accuracy and the structure recovery accuracy in low dimensional case.
In the higher dimensional (i.e., $d_1 = 20$) case, our proposed approach achieves the best structure recovery accuracy while maintaining nearly the same weight recovery accuracy with the best result (achieved by $\ell_1$ regularization-based method).
It is interesting to observe that adaptive $\ell_1$ regularization's performance lies between $\ell_1$ regularization and our proposed regularization.
In addition, our proposed regularization has a dominating performance over DAG regularization-based approaches in terms of both structure recovery accuracy and weight recovery accuracy. 
These observations are also validated by Table~\ref{table:exp2_penalty_comparison} in the appendix.

\section{Real Data Example}\label{sec:real_data} 
In this section, we demonstrate the usefulness of our proposed method in a real study. {We perform a train-validation-test split to select the models and their hyperparameters based on the predictive performance on the held-out test dataset. We show that the proposed discrete-time Hawkes network with a linear link function achieves the best performance. To enhance interpretability, we perform Bootstrap uncertainty quantification (UQ) to remove false discoveries in the causal graphs. Importantly, our proposed DAG-encouraging regularization can further boost both the predictive performance and the causal graph interpretability.}

\subsection{Settings}\label{sec:exp_settings}

\subsubsection{Dataset and Sepsis Associated Derangements}\label{sec:real_details}
{This real study targets a short time window right after the SOFA score turns 2 for ICU patients; see patient demographics in Table~\ref{table:demo}, Section~\ref{sec:motivating_eg}.} To {reduce the complexity of the computations due to high-dimensional raw features (i.e., vital signs and Lab results)}, expert (i.e., clinician) opinion is utilized to identify common and clinically relevant SADs that could be detected using structured EMR data. 
{In particular, those Labs and vital signs are all converted into binary SADs, representing nodes in the graph; As all those raw features are used in SADs' construction, they are not input to the model to avoid undesired correlation amongst nodes.} {Please find further details in Table~\ref{table:SADcutoff} in Appendix~\ref{appendix:lab_vital_SAD}.}


\subsubsection{Evaluation metrics}
The primary quantitative evaluation metric is Cross Entropy (CE) loss, as
our model outputs predicted probabilities for binary SADs sequentially. As SADs do not occur very often for each patient, we also use Focal loss as an alternative metric to account for such class imbalance issues. 
Furthermore, we focus on the interpretability of the resulting causal DAG by (i) counting the number of undesirable length-$L$ cycles, $L \in \{2,3,4,5\}$, and (ii) studying whether or not the inferred interactions align with well-known physiologic relationships. {Please find further details in Appendix~\ref{appendix:real_metric}.}

\subsubsection{Training details}\label{sec:realdata_split}
We perform train-validation-test data split to fine-tune the hyperparameters: we use the 2018 data as the training dataset, and we randomly split the 2019 dataset by half (for both sepsis and non-sepsis patient cohorts) into validation and testing datasets; we will select the hyperparameters based on the CE loss on the validation dataset and demonstrate its performance on the test dataset.
We fit the candidate models using the 2018 real data and select the hyperparameters based on the total CE loss on the 2019 validation dataset. {Please find further details in Appendix~\ref{appendix:real_train}.}

\subsection{Model comparison}\label{sec:realexp_comparison} 
\subsubsection{Candidate models}
We compare the VI-based estimation \ref{VI_1} with the black-box methods --- we choose XGBoost over other models, e.g., Neural Networks since it outperformed other candidate models in the 2019 Physionet Challenge on sepsis prediction \cite{physionet2019}. For the VI-based estimation, we consider linear, sigmoid, and exponential link functions. Furthermore, we incorporate expert opinion/prior medical knowledge that no inhibiting effect exists: on one hand, it remained unclear how to interpret inhibiting effects among SADs; on the other hand, restricting our consideration to triggering effects reduces the feasible region to $\theta \in \RR_+^{d \times d_1}$, which enables us to leverage PGD to numerically obtain the estimators.

\subsubsection{Results}
\begin{figure*}[!htp]
\centerline{
\includegraphics[width = \textwidth]{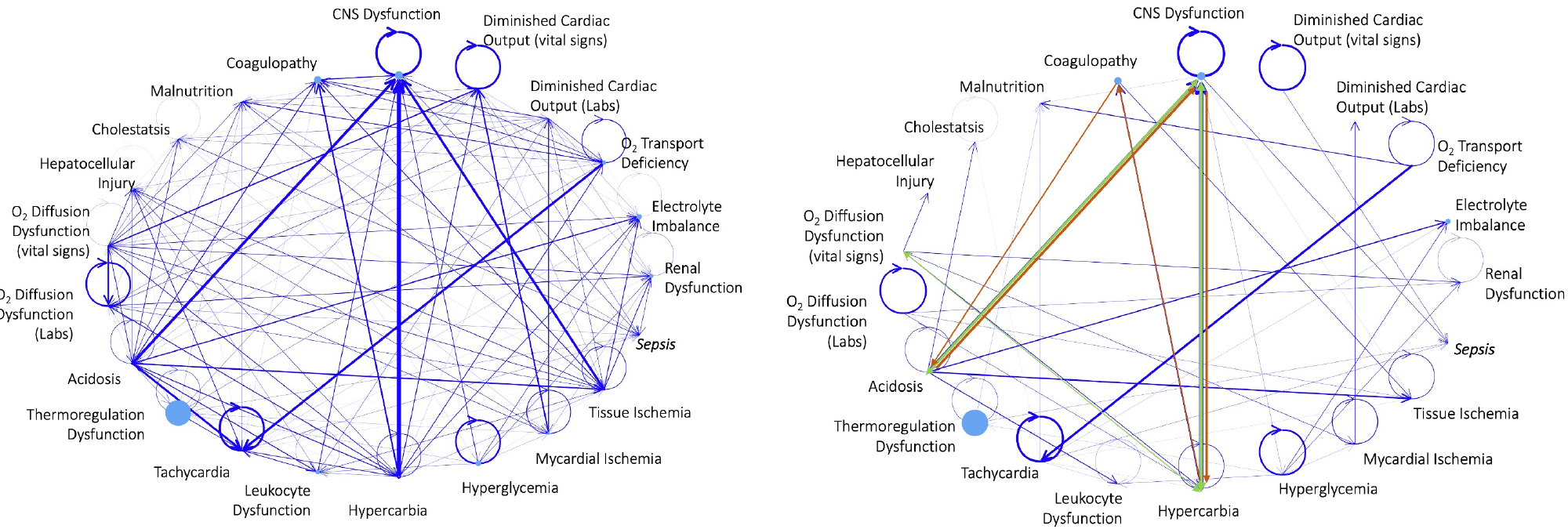}
}
\caption{{Causal graphs recovered by our proposed discrete-time Hawkes network using a linear link without (left) and with (right) Bootstrap uncertainty quantification. We remove the directed edges whose 95$\%$ Bootstrap confidence intervals contain zero. We can observe that, by comparing both graphs, the graph with BP has much fewer edges (which are caused by noisy observation) and is thus more interpretable. However, there still exist three length-2 cycles (green) and one length-4 cycle (orange) in this graph.}
}
\label{fig:GC_graph_bp} 
\end{figure*}

We visualize the recovered causal graphs of the SADs in Figure~\ref{fig:GC_graph_bp} and report the out-of-sample testing metrics in Table~\ref{table:CE_1}. We can observe that our proposed model with linear link function achieves the best performance for predicting most SADs; in particular, it has the smallest out-of-sample CE loss for sepsis prediction. Although the exponential link function outperforms the linear link for some SADs, the improvements are negligible; in addition, it performs rather poorly for CNS Dysfunction and Tachycardia predictions. Therefore, we will continue our real data analysis using the VI-based estimation coupled with the linear link function.
For completeness, further comparisons with XGBoost, VI-based estimation coupled with non-linear links (such as Figure~\ref{fig:GC_graphs_various_link}), and VI-based estimation without prior medical knowledge (i.e., with potential inhibiting effects) are deferred to Appendices~\ref{appendix:AUROC}, \ref{appendix:nonlinearlink}, and \ref{appendix:result_inhibiting}, respectively.

\begin{table}[htp]
\caption{Comparison between the VI-based estimation coupled with various link functions and black-box XGBoost: we report the average and standard deviation of Cross Entropy loss over all patients in the 2019 test dataset for all methods. The best results (before we round the number) are highlighted. We can observe the VI-based estimation coupled with linear link function outperforms other candidate methods when predicting most SADs. Although the exponential link function performs the best for many SADs, its improvements compared with linear links are marginal, let alone it is not robust in the sense that it performs poorly for CNSDys and Tachy predictions.
}\label{table:CE_1}
\begin{center}
\begin{small}
\resizebox{.8\textwidth}{!}{%
\begin{tabular}{lcccc}
\toprule[1pt]\midrule[0.3pt]
 &  Linear link  &  Exponential link  &  Sigmoid link  &  XGBoost \\
 \cmidrule(l){2-5}
RenDys & \textbf{.1246}$_{(.2591)}$ & .1246 $_{(.2586)}$ & .6198 $_{(1.5365)}$ & .3799 $_{(.2207)}$ \\
LyteImbal & .1781 $_{(.2623)}$ & \textbf{.1780}$_{(.2605)}$ & 1.0075 $_{(1.9436)}$ & .4511 $_{(.1644)}$ \\
O2TxpDef & \textbf{.1921}$_{(.2224)}$ & .1922 $_{(.2295)}$ & 1.1341 $_{(1.8396)}$ & .4515 $_{(.1964)}$ \\
DCO (L) & .0129 $_{(.0902)}$ & \textbf{.0128}$_{(.0896)}$ & .0399 $_{(.3269)}$ & .0847 $_{(.0939)}$ \\
DCO (V) & \textbf{.1461}$_{(.2909)}$ & .1491 $_{(.2934)}$ & .9865 $_{(2.7843)}$ & .3522 $_{(.3263)}$ \\
CNSDys & .2604 $_{(.4068)}$ & 1.0357 $_{(3.7069)}$ & .6446 $_{(1.1268)}$ & \textbf{.2491}$_{(.3492)}$ \\
Coag & .2441 $_{(.2034)}$ & \textbf{.2441}$_{(.2002)}$ & 1.5349 $_{(1.8136)}$ & .5254 $_{(.1670)}$ \\
MalNut & .1253 $_{(.2048)}$ & \textbf{.1252}$_{(.2036)}$ & .6470 $_{(1.3905)}$ & .3746 $_{(.2091)}$ \\
Chole & \textbf{.0495}$_{(.1712)}$ & .0495 $_{(.1710)}$ & .1985 $_{(.8692)}$ & .2420 $_{(.2142)}$ \\
HepatoDys & .0839 $_{(.2061)}$ & \textbf{.0838}$_{(.2051)}$ & .3943 $_{(1.1954)}$ & .3207 $_{(.1898)}$ \\
O2DiffDys (L) & \textbf{.0081}$_{(.1482)}$ & .0095 $_{(.1833)}$ & .0141 $_{(.2543)}$ & .0181 $_{(.0455)}$ \\
O2DiffDys (V) & \textbf{.1494}$_{(.3069)}$ & .1524 $_{(.3105)}$ & 1.018 $_{(2.7288)}$ & .3503 $_{(.3170)}$ \\
Acidosis & .0609 $_{(.1850)}$ & \textbf{.0608}$_{(.1841)}$ & .3347 $_{(1.3496)}$ & .2191 $_{(.1457)}$ \\
ThermoDys & \textbf{.0053}$_{(.0470)}$ & .0306 $_{(.4780)}$ & .0306 $_{(.4780)}$ & .4426 $_{(.1440)}$ \\
Tachy & \textbf{.3911}$_{(.4710)}$ & 2.4593 $_{(4.4066)}$ & .5624 $_{(.6535)}$ & .3967 $_{(.3082)}$ \\
LeukDys & .1243 $_{(.2368)}$ & \textbf{.1243}$_{(.2360)}$ & .6385 $_{(1.515)}$ & .3887 $_{(.2047)}$ \\
HypCarb & \textbf{.0511}$_{(.1773)}$ & .0513 $_{(.1782)}$ & .2442 $_{(1.0791)}$ & .2220 $_{(.1911)}$ \\
HypGly & \textbf{.2849}$_{(.2766)}$ & .2879 $_{(.2773)}$ & 2.274 $_{(3.5716)}$ & .5141 $_{(.2314)}$ \\
MyoIsch & \textbf{.0587}$_{(.2313)}$ & .0591 $_{(.2331)}$ & .2746 $_{(1.265)}$ & .2268 $_{(.2191)}$ \\
TissueIsch & .0739 $_{(.1814)}$ & \textbf{.0739}$_{(.1799)}$ & .4139 $_{(1.4378)}$ & .2664 $_{(.1593)}$ \\
SEP3 & \textbf{.1318}$_{(.1803)}$ & .1319 $_{(.1794)}$ & .6915 $_{(1.2001)}$ & .4155 $_{(.2680)}$ \\
\midrule[0.3pt]\bottomrule[1pt]
\end{tabular}
}
\end{small}
\end{center}
\end{table}

\subsection{Causal DAG discovery}
Table~\ref{table:CE_1} shows the competitive performance of the VI-based estimation coupled with the linear link function, and thus we continue our analysis with such mode choice. The objective now is to improve the result interpretability by considering causal structural learning. 

\subsubsection{Bootstrap uncertainty quantification} We report the number of length-$L$ cycles for $L \in \{2,3,4,5\}$ in Table~\ref{table:cycles}. As we can see from that table, the recovered graph without Bootstrap UQ and DAG-inducing regularization contains many cycles, making the results less explainable.

\begin{table}[htp]
\caption{Comparison among various types of regularization: we report the number of cycles for various methods. We can see Bootstrap can remove most of the cycles by removing the ``less important edges'' in the graph; see the comparison between both graphs in Figure~\ref{fig:GC_graph_bp} for a graphical illustration. Moreover, built on top of the Bootstrap UQ, proper DAG-inducing regularization can completely remove cycles and encourage a desired ``DAG with self-exciting components'' structure.
}\label{table:cycles}
\begin{center}
\begin{small}
\resizebox{\textwidth}{!}{%
\begin{tabular}{lcccccccc}
\toprule[1pt]\midrule[0.3pt]
  &  Linear link  &  Exponential link  &  Sigmoid link  &  Linear link   &  Linear link   &  Linear link   &  Linear link  &  Linear link \\
  &    &    &    &   (BP)  &   (BP + proposed)  &   (BP + $\ell_1$)  &   (BP + DAG)  &   (BP + DAG-variant) \\
   \cmidrule(l){2-9}
Num. of Len-2 Cycles &  39 &  45 &  0 &  3 &  0 &  0 &  0 &  3 \\
Num. of Len-3 Cycles &  136 &  192 &  0 &  0 &  0 &  0 &  0 &  0 \\
Num. of Len-4 Cycles &  680 &  1023 &  0 &  1 &  0 &  0 &  0 &  1 \\
Num. of Len-5 Cycles &  3310 &  5419 &  0 &  0 &  0 &  0 &  0 &  0 \\
\midrule[0.3pt]\bottomrule[1pt]
\end{tabular}
}
\end{small}
\end{center}
\end{table}

The left panel in Figure~\ref{fig:GC_graph_bp} shows many edges with very small weights, meaning that the existence of such an edge might be a result of noisy observations. For example, although the edge from Diminished Cardiac Output (vital signs) to sepsis events agrees with the well-known causal relationships in sepsis-related illness, its weight is too small to convince the clinician that such a triggering effect is statistically significant. Thus, before applying regularization, we first perform Bootstrap UQ and the existence of an edge is determined by its Bootstrap confidence interval: we assign zero weight to that edge if its $95\%$ CI contains zero; otherwise, we use the median of the Bootstrap results as the weight. Here, we obtain the CI based on $1500$ Bootstrap trails; complete details are deferred to Appendix~\ref{appendix:uncertainty}.
The resulting graph is reported in the right panel in Figure~\ref{fig:GC_graph_bp} and the CE loss is reported in Table~\ref{table:CE_2}. 

The results in Table~\ref{table:cycles} and Figure~\ref{fig:GC_graph_bp} show that BP can remove a substantive amount of cycles. Importantly, it is good to observe that the well-known triggering effect from Diminished Cardiac Output (vital signs) to sepsis events is statistically significant; see Figure~\ref{fig:GC_graph_bp}.
However, as evidenced by Tables~\ref{table:CE_1} and \ref{table:CE_2}, performing Bootstrap UQ leads to much worse CE loss for almost all SADs 
To improve its prediction performance to make the interpretable graphs more convincing, and to remove the remaining cycles highlighted in Figure~\ref{fig:GC_graph_bp}, we consider causal structural learning via our penalized VI-based estimation such as \ref{VI_3}.

\begin{table}[htp]
\caption{Comparison among various types of regularization: we report the average and standard deviation of Cross Entropy loss over all patients in the 2019 test dataset. The best results (before we round the number) are highlighted. We can observe our proposed data-adaptive linear regularization can achieve the best performance for most SADs compared with other DAG-inducing regularizations; moreover, by comparing this table with Table~\ref{table:CE_1}, we can see it achieves almost the same performance as the best achievable performance. 
}\label{table:CE_2}
\begin{center}
\begin{small}
\resizebox{\textwidth}{!}{%
\begin{tabular}{lcccccc}
\toprule[1pt]\midrule[0.3pt]
  &  Linear link (BP)  &  Linear link (BP + proposed)  &  Linear link (BP + $\ell_1$)  &  Linear link (BP + DAG)  &  Linear link (BP + DAG-variant) \\
  \cmidrule(l){2-6}
RenDys & .2342 $_{(.6967)}$ & .1253 $_{(.2655)}$ & \textbf{.1240}$_{(.2586)}$ & .1261 $_{(.2686)}$ & .2342 $_{(.6967)}$ \\
LyteImbal & .1801 $_{(.2793)}$ & .1801 $_{(.2741)}$ & \textbf{.1792}$_{(.2709)}$ & .1794 $_{(.2790)}$ & .1801 $_{(.2793)}$ \\
O2TxpDef & .5044 $_{(.8425)}$ & \textbf{.1920}$_{(.2310)}$ & .1921 $_{(.2320)}$ & .1970 $_{(.2511)}$ & .5044 $_{(.8425)}$ \\
DCO (L) & .0308 $_{(.2553)}$ & .0129 $_{(.0967)}$ & \textbf{.0128}$_{(.0898)}$ & .0320 $_{(.2615)}$ & .0308 $_{(.2553)}$ \\
DCO (V) & .3142 $_{(.7536)}$ & \textbf{.1451}$_{(.2892)}$ & .1453 $_{(.2852)}$ & .1784 $_{(.4096)}$ & .3142 $_{(.7536)}$ \\
CNSDys & .2583 $_{(.4001)}$ & .2589 $_{(.3810)}$ & .2617 $_{(.3882)}$ & .4609 $_{(.2358)}$ & \textbf{.2583}$_{(.4001)}$ \\
Coag & .2461 $_{(.2236)}$ & .2454 $_{(.2177)}$ & .2451 $_{(.2138)}$ & \textbf{.2450}$_{(.2141)}$ & .2461 $_{(.2236)}$ \\
MalNut & \textbf{.1270}$_{(.2209)}$ & .1290 $_{(.2343)}$ & .1290 $_{(.2304)}$ & .1337 $_{(.2466)}$ & .1270 $_{(.2209)}$ \\
Chole & .0689 $_{(.2679)}$ & .0689 $_{(.2679)}$ & \textbf{.0496}$_{(.1890)}$ & .0976 $_{(.3813)}$ & .0689 $_{(.2679)}$ \\
HepatoDys & .3155 $_{(.9563)}$ & .0888 $_{(.2393)}$ & \textbf{.0860}$_{(.2232)}$ & .0887 $_{(.2379)}$ & .3155 $_{(.9563)}$ \\
O2DiffDys (L) & .0075 $_{(.1495)}$ & \textbf{.0051}$_{(.0883)}$ & .0052 $_{(.0876)}$ & .0113 $_{(.2034)}$ & .0075 $_{(.1495)}$ \\
O2DiffDys (V) & .3370 $_{(.9067)}$ & .1507 $_{(.3177)}$ & \textbf{.1504}$_{(.3146)}$ & .1849 $_{(.4181)}$ & .3370 $_{(.9067)}$ \\
Acidosis & .0696 $_{(.2577)}$ & .1123 $_{(.4437)}$ & \textbf{.0611}$_{(.1864)}$ & .0779 $_{(.2809)}$ & .0696 $_{(.2577)}$ \\
ThermoDys & .0066 $_{(.0476)}$ & .0057 $_{(.0478)}$ & \textbf{.0055}$_{(.0480)}$ & .0098 $_{(.1183)}$ & .0064 $_{(.0476)}$ \\
Tachy & .4835 $_{(.7772)}$ & .3736 $_{(.3732)}$ & \textbf{.3730}$_{(.3725)}$ & .5379 $_{(.1770)}$ & .4835 $_{(.7772)}$ \\
LeukDys & .1688 $_{(.4368)}$ & .1261 $_{(.2506)}$ & \textbf{.1255}$_{(.2452)}$ & .1271 $_{(.2538)}$ & .1688 $_{(.4368)}$ \\
HypCarb & .0512 $_{(.1814)}$ & .0513 $_{(.1791)}$ & .0514 $_{(.1801)}$ & .0593 $_{(.2290)}$ & \textbf{.0512}$_{(.1815)}$ \\
HypGly & .4121 $_{(.6075)}$ & \textbf{.2853}$_{(.2793)}$ & .2853 $_{(.2785)}$ & .3189 $_{(.3336)}$ & .4121 $_{(.6075)}$ \\
MyoIsch & .1040 $_{(.5856)}$ & .1295 $_{(.6292)}$ & \textbf{.0585}$_{(.2312)}$ & .0687 $_{(.2952)}$ & .1040 $_{(.5856)}$ \\
TissueIsch & .0741 $_{(.1936)}$ & .0743 $_{(.1891)}$ & .0745 $_{(.1857)}$ & .1067 $_{(.3488)}$ & \textbf{.0741}$_{(.1936)}$ \\
SEP3 & .1323 $_{(.1848)}$ & .1323 $_{(.1848)}$ & .1349 $_{(.1794)}$ & \textbf{.1320}$_{(.1810)}$ & .1323 $_{(.1848)}$ \\
\midrule[0.3pt]\bottomrule[1pt]
\end{tabular}
}
\end{small}
\end{center}
\end{table}

\subsubsection{Causal DAG recovery via regularization}
We adopt the regularization approaches described in Section~\ref{sec:DAG_est} and Section~\ref{sec:DAG_baseline_formulation}; again, for each regularization, we perform Bootstrap UQ with $1500$ trials and $95\%$ confidence level; we select the regularization strength hyperparameter $\lambda$ using grid search based on the validation total CE loss.  We report the CE loss on the test dataset for each regularization (with corresponding selected $\lambda$'s) in Table~\ref{table:CE_2}; 
the resulting graphs are visualized at the beginning of this paper in Figure~\ref{fig:GC_graph_bp_reg}. 
From Table~\ref{table:CE_2}, we can observe that our proposed data-adaptive linear regularization can not only remove cycles while keeping the lagged self-exciting components but also reduce the out-of-sample prediction CE loss. This suggests that Bootstrap coupled with our proposed DAG-inducing regularization outputs a highly interpretable causal DAG (Figure~\ref{fig:GC_graph_bp_reg} and Table~\ref{table:cycles}) while achieving almost identical out-of-sample prediction performance (Tables~\ref{table:CE_1} and \ref{table:CE_2}). {Additionally, we report an additional metric --- Focal loss --- in Table~\ref{table:FL}, Appendix~\ref{appendix:FL}, and hyperparameter $\lambda$ selection table in Table~\ref{table:hyperparameter}, Appendix~\ref{appendix:DAG_effect_lambda}, re-affirming the aforementioned findings.}

\subsubsection{Interpretation}\label{sec:real_exp_interpretation}
Figure~\ref{fig:GC_graph_bp_reg} elucidates which relationships are most important in the graph, which is an essential aspect of interpretability. For example, the triggering effect from Diminished Cardiac Output (vital signs) to sepsis events remains significant after the Bootstrap UQ and cycle elimination; in fact, nearly all triggering effects of sepsis events remain significant. {We provide the top causes of sepsis discovered in Figure~\ref{fig:GC_graph_bp_reg}, showing their similarity to the results of XGBoost \cite{yang2020explainable} on clinically published data \cite{physionetChallenge}. Due to space consideration, one can find those results in Appendix~\ref{appendix:sepsis_causes}.}

The primary outcome of interest for this work was sepsis. {Meanwhile,} as demonstrated in Figure~\ref{fig:GC_graph_bp_reg}, the causal relationship between any node pair can be estimated: Indeed, we can identify several strong triggering effects shared by both graphs, which are commonly recognized (though in other types of patients). 
For example, the exciting effect from
Hyperglycemia to Electrolyte Imbalance is commonly seen in type 2 diabetes patients \cite{khan2019pattern}, 
and the observation that
Acidosis precedes Cholestatsis is common for patients with pregnancy \cite{sterrenburg2014acidosis}.
Meanwhile, our model can predict all SAD events in the graph, and this gives clinician users insight into the probability of observing subsequent SADs after sepsis. Even though our prediction of sepsis events is not perfect, the ability to predict other SADs that are on the path to sepsis or identify different potential pathways to an adverse event is also very important for clinicians to respond to those potential adverse events accordingly. Overall, the fact that identified triggering effects agree with the well-known physiologic relationships and the satisfying predictive performance affirm the usefulness of our proposed method.

\section{Conclusion}\label{sec:discussion}

In this work, we present a GLM for causal DAG discovery from mutually exciting time series data.
Most importantly, our proposed data-adaptive linear DAG-inducing regularization helps formulate the model estimation as a convex optimization problem.
Furthermore, we establish a non-asymptotic estimation error upper bound for the GLM, which is verified numerically; we also give a confidence interval by solving linear programs. 
Both our numerical simulation and real data example show the good performance of our proposed method, making its future adoption in conducting continuous surveillance under medical settings and other similar problems much more likely.  Meanwhile, there are a few interesting topics that the current work does not cover. For example, the convexity inherent in our proposed data-adaptive linear causal discovery method opens up the possibility of establishing performance guarantees, which we leave for future discussion.

\section*{Acknowledgment}
The work of Song Wei and Yao Xie is partially supported by an NSF CAREER CCF-1650913, and NSF DMS-2134037, CMMI-2015787, CMMI-2112533, DMS-1938106, DMS-1830210, and an Emory Hospital grant.


\bibliographystyle{IEEEtran}
\bibliography{ref}

\newpage

\appendices

\addcontentsline{toc}{section}{Appendix} 
\part{\centering \LARGE Appendix of Causal Graph Discovery \\ from Self and Mutually Exciting Time Series} 

\topskip0pt
\vspace*{\fill}

\parttoc 

\vspace*{\fill}

\newpage

\section{Extended Literature Survey}\label{appendix:extended_literature}

\begin{figure*}[!htp]
\centerline{
\includegraphics[width = .6\textwidth]{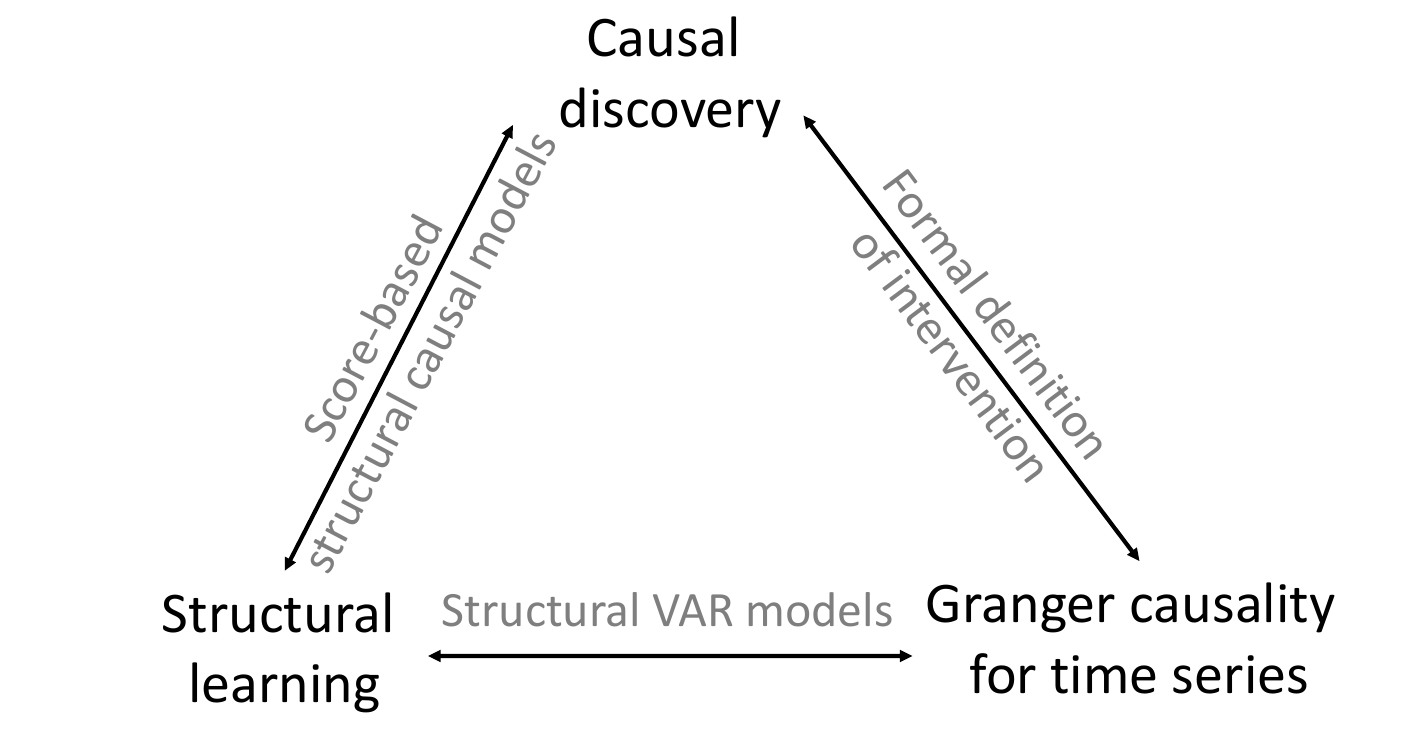}
}
\caption{A diagram illustrating the connections among causal discovery, structural learning and Granger causality for time series.
}
\label{fig:literature_diagram} 
\end{figure*}

In this section, in addition to the literature survey on structural learning and Granger causality in Section~\ref{subsec:literature}, we review related works in causal discovery, causality for time series and Granger causality for point process data. To help readers understand how those areas connect with each other, we draw a diagram in Figure~\ref{fig:literature_diagram}.

\subsection{Causal inference}
Causal discovery from observational data mainly consists of constraint-based and score-based methods, aiming at finding causal relationships represented on DAGs \cite{pearl2009causality}. The state-of-the-art (conditional independence) constraint-based methods include PC, and Fast Causal Inference (FCI) \cite{spirtes2000constructing}. Both algorithms can output the underlying true DAG structure in the large sample limit. However, PC cannot deal with unobserved confounding, whereas FCI is capable of dealing with confounders. 
Since both algorithms rely on conditional independence tests to eliminate edges from the complete graph, they are not scalable when the number of nodes becomes large; existing methods to handle this include imposing sparse structure \cite{kalisch2007estimating}, leveraging the parallel computing technique \cite{le2016fast}, and score-based methods.

In score-based causal discovery methods, the first step is to define a score function measuring how well the model fits the observations, e.g., Bayesian Information Criterion (BIC), (log-)likelihood, and so on. For example, Greedy Equivalence Search \cite{chickering2002optimal} starts with an empty graph and recursively adds/eliminates edges based on the BIC score.
In particular, a line of research focuses on modeling the causal relationship using structural causal models (or structural equation models), and the joint probability distribution over a DAG structure helps define a log-likelihood score function. For example,
\cite{shimizu2006linear} studied a Linear, Non-Gaussian, Acyclic Model and proposed independent component analysis to infer the deterministic SCM with additive non-Gaussian noise; most importantly, \cite{shimizu2006linear} discovered that linear-non-Gaussian assumption helps identify the full causal model. 
Another notable topic in this direction is the variational-inference-based approach for Bayesian casual discovery, where the difficulty comes from efficiently searching a combinatorially large DAG family to maximize the score. Examples include
\cite{cundy2021bcd}, who factorized the adjacency matrix into a permutation matrix and a strictly lower-triangular matrix to ensure DAG-ness, 
\cite{lorch2021dibs, annadani2021variational}, who leveraged a new algebraic characterization of DAG \cite{zheng2018dags} in the prior distribution.
In addition to the aforementioned causal discovery from observational data, there are also works focusing on recovering the causal graph from interventional data, where the log-likelihood score function is the joint interventional density. Notable contributions include \cite{brouillard2020differentiable}, who proposed a continuous optimization-based approach by leveraging the aforementioned continuous and differentiable DAG characterization \cite{zheng2018dags}. 
For a systematic survey of recent developments on causal inference, we refer readers to \cite{glymour2019review}.

\subsection{Causality for time series data}
When we consider observations exhibiting serial correlation, i.e., time series, Granger causality \cite{granger1969investigating,granger1980testing,granger1988some}, which assesses whether the past of one time series is predictive of another, is a popular notion of causality due to its broad application, ranging from
economics \cite{chiou2008economic} and finance \cite{hong2009granger}, meteorology \cite{mosedale2006granger}, neuroscience \cite{seth2015granger}, health-care \cite{wei2022granger} and so on.
In the seminal work, \cite{eichler2010granger} went beyond Granger causality by formally defining intervention and studying the causality in terms of the effect of the intervention on the VAR model. In addition to using the VAR model as the SCM, \cite{peters2022causal} proposed an ordinary differential equation modification of SCM; \cite{peters2022causal} also formally defined interventions and went beyond Granger causality to study the causality as the notion of interventional effect. 
However, they assumed known graph structure and did not answer the causal graph discovery question as SVAR models did. Thus, one may still need the aforementioned SVAR models to uncover the causal graph structure from observational time series data.
Another line of research studied direction information, a generalized notion of Granger causality, as the notion of causality for time series data \cite{quinn2011estimating,quinn2011equivalence,quinn2015directed}.
Moreover, \cite{quinn2011equivalence} pointed out that both Granger causal graph and direction information graph are equivalent to minimal generative model graphs and, therefore, can be used for causal inference in the same manner as Bayesian networks are used for correlative statistical inference. 
One limitation of the Granger causality framework comes from the no unobserved confounding assumption; existing efforts to tackle this issue include an FCI algorithm for time series to handle confounders \cite{entner2010causal}. 

\subsection{Granger causality for point process}
Recently, \cite{kim2011granger} initiated the study of Granger causality for multivariate Hawkes Processes \cite{hawkes1971point,hawkes1971spectra,hawkes1974cluster}, i.e., Hawkes network. Notable contributions include alternating direction method of multipliers coupled with low-rank structure in mutual excitation matrix \cite{zhou2013learning},  Fourier-transform based time series approach \cite{etesami2016learning},
expectation–maximization (EM) algorithm with various constraints \cite{xu2016learning,chen2022learning,ide2021cardinality}, 
Neural Network-based approach \cite{zhang2020cause}, a regularized scalable gradient-based algorithm \cite{wei2022granger}, and so on.
Our model is different from the aforementioned method in the following aspects: (i)
Theoretically speaking, the guarantee is largely missing for Neural Network and EM-based methods --- EM only maximizes a lower bound on log-likelihood and could converge to a local solution, let alone the Neural Network method. Moreover, EM is not scalable due to its quadratic complexity in the number of events. 
(ii) Practically speaking, our model handles multiple short time series sequences (most patients have short ICU stays, and the measurements are recorded hourly), whereas the aforementioned methods dealing with one long point process sequence cannot be easily reformulated to handle our target dataset. 
Nevertheless, we numerically compare with our previously proposed scalable gradient-based approach \cite{wei2022granger} that shares a similar problem formulation with our method.
Both our numerical simulation and real data example show our method has a better performance over the MHP method \cite{wei2022granger}, potentially due to the lack of granularity of time series data (compared with point process data). 


\section{Additional Technical Details}

\subsection{Granger causality}\label{appendix:causality}
The Granger non-causality can be obtained as follows:
\begin{proposition}
Time series $\{y_t^{(j)}, t \in [T]\}$ is Granger non-causal for time series $\{y_t^{(i)}, t \in [T]\}$ if and only if (iff) $\alpha_{ijk} = 0$ for all $k \in [\tau]$ in our GLM \eqref{eq:model}.
\end{proposition}

We need to mention that inferring Granger causality needs ``all the information in the universe'' and hence we can only infer Granger non-causality and {\it prima facie causality} given partially observed data \cite{granger1969investigating,granger1980testing,granger1988some}. In the graph induced by $A_k = (\alpha_{ijk}), k \in [\tau]$, the absence of an edge means Granger non-causality, whereas the presence of an edge in $A$ only implies prima facie causality. In this work, we will assume there is {\it no unobserved confounding} and we say time series $\{y_t^{(j)}, t \in [T]\}$ Granger causes time series $\{y_t^{(i)}, t \in [T]\}$ if there exists $k \in [\tau]$ such that $\alpha_{ijk} \not= 0$.

\begin{remark}[Interpretation as DIG]
This excitation/inhibition matrix $A$ can be understood as a directed information graph (DIG) \cite{etesami2016learning}, which is a generalized causal notion of Granger causality. To be precise, in DIG, we determine the causality by comparing two conditional probabilities in the KL-divergence sense: one is the conditional probability of $N^i_{t+dt}$ given full history, and the other one is the conditional probability of $N^i_{t+dt}$ given full history except that of type-$j$ event. Last but not least, both Granger causal graph and DIG are equivalent to minimal generative model graphs \cite{quinn2011equivalence} and therefore can be used for causal inference in the same manner Bayesian networks are used for correlative statistical inference.
\end{remark}

\begin{remark}[Effect of intervention]
The study of causality as a notion of interventional effect requires formally defining intervention, which is straightforward in the SCM. To be precise, an intervention will modify the joint distribution by changing the adjacency matrix and thus the causal graph structure, which is represented by a changed parameter $\theta$. This enables us to have a ``score'' that corresponds to the interventional joint density, which we will see later is actually captured by a vector field. Therefore, when we can access the interventional data, we can study the interventional effect as \cite{brouillard2020differentiable} did. Here, we will leave it for future discussion since we restrict our consideration to observational data in this work.
\end{remark}


\subsection{Proof of the non-asymptotic performance guarantee}\label{appendix:proof}
We define an auxiliary vector field 
$$\Tilde{F}_{T}^{(i)}(\theta_i) = \frac{1}{T} \sum_{t=1}^T w_{t - \tau : t-1}( g(w_{t - \tau : t-1}^\T \theta_i) -  g(w_{t - \tau : t-1}^\T \theta^\star_{i}) ),$$
where $\theta^\star_{i}$ is the unknown ground truth.
This vector field changes the constant term in ${F}_{T}^{(i)}(\theta_i)$ to ensure its unique root/weak solution to corresponding VI is $\theta^\star_{i}$. Recall that our proposed estimator $\hat \theta_i$ is the root of ${F}_{T}^{(i)}(\theta_i)$.
In the following, we will bound the difference between $\hat \theta_i$ and $\theta^\star_{i}$ via the difference between the empirical vector field ${F}_{T}^{(i)}(\theta_i)$ and the auxiliary vector field $\Tilde{F}_{T}^{(i)}(\theta_i)$, i.e.,
$$\Delta^{(i)} = {F}_{T}^{(i)}(\theta_i) - \Tilde{F}_{T}^{(i)}(\theta_i) = {F}_{T}^{(i)}(\theta^\star_{i}).$$

\begin{proposition}\label{lma:bound_delta}
Under Assumptions~\ref{assumption:vector_field} and \ref{assumption:observation}, for any $\varepsilon \in (0,1)$, with probability at least $1-\varepsilon$, for any $ i \in [d_1],$ $\Delta^{(i)}$ can be bounded as follows:
\begin{equation}\label{eq:bound}
    \norm{\Delta^{(i)}}_\infty \leq M_w \sqrt{{\log(2d/\varepsilon)}/{T}}, \text{ which implies } \norm{\Delta^{(i)}}_2 \leq M_w \sqrt{d{\log(2d/\varepsilon)}/{T}}.
\end{equation}
\end{proposition}

\begin{proof}
Denote random vector 
$\xi_t= w_{t - \tau : t-1} \left( g\left(w_{t - \tau : t-1}^\T \theta^\star_{i}\right) -  y_t^{(i)} \right).$
We can re-write $\Delta^{(i)} = \sum_{t=1}^T \xi_t/T.$ Define $\sigma$-field $\mathcal{F}_t = \sigma (W_t)$, and $\mathcal{F}_0 \subset \mathcal{F}_1 \subset \cdots \mathcal{F}_T$ form a filtration. We can show $$\mathbb{E}[\xi_t|\mathcal{F}_{t-1}] = 0, \quad {\rm Var}(\xi_t|\mathcal{F}_{t-1}) =  g\left(w_{t - \tau : t-1}^\T \theta_i\right) \left(1-g\left(w_{t - \tau : t-1}^\T \theta_i\right)\right) \leq 1/4.$$ 
This means $\xi_t, t \in [T],$ is a Martingale Difference Sequence. Moreover, its infinity norm is bounded by $M_w$ almost surely (Assumption~\ref{assumption:observation}). Therefore, by Azuma's inequality, we have
$$\mathbb{P}\left( |\Delta^{(i)}_k| > u \right) \leq 2 \exp\left\{ - \frac{T u^2}{2M_w^2}\right\}, \ k \in [d], \ \forall \  u>0,$$
where $\Delta^{(i)}_k$ is the $k$-th entry of vector $\Delta^{(i)}$. By union bound, we have 
$$\mathbb{P}\left( |\Delta^{(i)}_k| > u, \ k = 1,\dots,d \right) \leq 2d \exp\left\{ - \frac{T u^2}{2M_w^2}\right\}, \ \forall \  u>0.$$
By setting the RHS of above inequality to $\varepsilon$ and solving for $u$, we prove infinity norm bound in \eqref{eq:bound}. Besides,
since $\norm{\Delta}_2 \leq \sqrt{d}  \norm{\Delta}_\infty$ holds for any vector $\Delta \in \mathbb{R}^d$, we can easily prove the $\ell_2$ norm bound using the infinity norm bound in \eqref{eq:bound}.
\end{proof}

The proof of Proposition~\ref{lma:bound_delta} leverages the concentration property of martingales. Similar results could also be found in \cite{juditsky2020convex,wei2021inferring}. 
By leveraging this proposition, we can prove the non-asymptotic estimation error bound in Theorem~\ref{thm:upper_err_bound} as follows:

\begin{proof}[Proof of Theorem~\ref{thm:upper_err_bound}]
Under Assumption~\ref{assumption:vector_field}, the vector field $F_{T}^{(i)}(\theta_i)$ is monotone modulus $m_g \lambda_1$ since
\begin{align*}
    \left(F_{T}^{(i)}(\theta) - F_{T}^{(i)}(\theta')\right)^\T (\theta - \theta') &= \frac{1}{T} \sum_{t=1}^T  \left(w_{t - \tau : t-1}^\T\theta - w_{t - \tau : t-1}^\T\theta'\right) \left(g\left(w_{t - \tau : t-1}^\T\theta\right) - g\left(w_{t - \tau : t-1}^\T\theta'\right)\right)\\
    &\geq m_g \frac{1}{T} \sum_{t=1}^T \norm{ w_{t - \tau : t-1}^\T (\theta - \theta')}_2^2 = m_g (\theta - \theta')^\T \mathbb{W}_{1:T} (\theta - \theta')\\
    & \geq m_g \lambda_1 \norm{\theta - \theta'}_2^2.
\end{align*}
In particular, we have:
$$\left({F}_{T}^{(i)}(\hat \theta_i)- {F}_{T}^{(i)}(\theta^\star_{i})\right)^\T (\hat \theta_i - \theta^\star_{i}) \geq m_g\lambda_1 \norm{\hat \theta_i -  \theta^\star_{i}}_2^2.$$
Notice that our weak solution $\hat \theta_i$ is also a strong solution to the VI since the empirical vector field is continuous, which gives us
$$\left({F}_{T}^{(i)}(\hat \theta_i)\right)^\T (\hat \theta_i - \theta^\star_{i}) \leq 0.$$
By Cauchy Schwartz inequality, we also have $$- \left({F}_{T}^{(i)}(\theta^\star_{i})\right)^\T (\hat \theta_i - \theta^\star_{i}) = - \Delta_i^\T (\hat \theta_i - \theta^\star_{i}) \leq \norm{\Delta_i}_2 \norm{\hat \theta_i -  \theta^\star_{i}}_2.$$ 
Together with the $\ell_2$ norm bound in \eqref{eq:bound}, Proposition~\ref{lma:bound_delta}, we complete the proof.
\end{proof}


\subsection{Linear program-based confidence interval}\label{appendix:CI}

As pointed out in section II.E \cite{juditsky2020convex}, for general non-linear link function $g$, it would be hard to separate $\theta_i$ from $\sum_{t=1}^T w_{t - \tau : t-1} g\left(w_{t - \tau : t-1}^\T \theta_i\right)/T$. First, we derive a CI for linear link function case via a more precise data-driven bound for $F_{T}^{(i)}(\theta_i)\in \mathbb{R}^d$ as \cite{juditsky2020convex} did in Lemma 2 (see its proof in Appendix~\ref{appendix:CI}):

\begin{proposition}[Confidence interval for linear transform of $\theta_i$ for linear link function case]\label{prop:CI}
Under Assumptions~\ref{assumption:vector_field} and \ref{assumption:observation}, for $i \in [d_1]$, and every $s > 1$, the following holds with probability at least $1-2d\{s[\log((s-1)T)+2] + 2\} e^{1-s}$:
$$\theta_\ell[W_T,s;i] \leq a^\T \theta_i \leq \theta_u[W_T,s;i], \quad \forall a \in \mathbb{R}^d,$$where $\theta_\ell[W_T,s;i]$ and $\theta_u[W_T,s;i]$ are defined in \eqref{eq:CI_lower}, \eqref{eq:CI_upper}.

\end{proposition}

This CI is obtained by solving LPs \eqref{eq:CI_lower} and \eqref{eq:CI_upper}. For general non-linear link function $g$, since it is typically constrained in a compact subset to satisfy Assumption~\ref{assumption:vector_field}, we can obtain linear bounds on the non-linear link function and then repeat the above techniques to obtain similar CI.

\begin{proof}
We first define some notations:
\begin{align}
    \theta_\ell[W_T,s;i] &=\min\left\{\begin{array}{ll}
a^\T \theta_i: & \theta_i \in \Theta, \\
& \psi_\ell \left(a[W_T;i]  ,T;s\right) \leq \frac{(\mathbb{W}_{1:T}\theta_i)_k}{M_w} \leq \psi_u \left(a[W_T;i]  ,T;s\right), k \in [d_1].
\end{array}\right\} \label{eq:CI_lower}\\
\theta_u[W_T,s;i] &=\max\left\{\begin{array}{ll}
a^\T \theta_i: & \theta_i \in \Theta, \\
& \psi_\ell \left(a[W_T;i]  ,T;s\right) \leq \frac{(\mathbb{W}_{1:T}\theta_i)_k}{M_w} \leq \psi_u \left(a[W_T;i]  ,T;s\right), k  \in [d_1].
\end{array}\right\} \label{eq:CI_upper}
\end{align}

Here, \begin{align*}
{\psi_\ell}(\nu, d ; y)&=\left\{\begin{array}{l}
(d+2 y)^{-1}\left[d \nu+\frac{2 u}{3}-\sqrt{2 d \nu y+\frac{y^{2}}{3}-\frac{2 u}{d}\left(\frac{y}{3}-\nu d\right)^{2}}\right] \text { if } \nu>\frac{y}{3 d} \\
0 \quad \text { otherwise }
\end{array}\right.\\
{\psi_u}(\nu, d ; y)&=\left\{\begin{array}{l}
(d+2 y)^{-1}\left[d \nu+\frac{4 y}{3}+\sqrt{2 d \nu y+\frac{5 y^{2}}{3}-\frac{2 y}{d}\left(\frac{y}{3}+\nu d\right)^{2}}\right] \text { if } \nu<1-\frac{y}{3 d} \\
1 \text { otherwise }
\end{array}\right.
\end{align*}

For $i \in [d_1]$, 
we will derive a more precise data-driven bound for $F_{T}^{(i)}(\theta_i)\in \mathbb{R}^d$ as \cite{juditsky2020convex} did in Lemma 2:

\begin{lemma}\label{lma:bound_CI}
Under Assumptions~\ref{assumption:vector_field} and \ref{assumption:observation}, for $i \in [d_1]$, and every $s > 1$, the following holds with probability at least $1-2d\{s[\log((s-1)T)+2] + 2\} e^{1-s}$:
$$ M_w\psi_\ell \left(a[W_T;i]  ,T;s\right)\mathbf{1}_d - a[W_T;i] \leq  F_{T}^{(i)}(\theta_i) \leq   M_w\psi_u \left(a[W_T;i]  ,T;s\right)\mathbf{1}_d - a[W_T;i],$$
where $\mathbf{1}_d \in \mathbb{R}^d$ is the vector of ones, the inequality between vectors is element-wise, and $$a[W_T;i]=\sum_{t=1}^T w_{t - \tau : t-1} y_t^{(i)}/T.$$
\end{lemma}

Then, it is easy to complete the proof, since by the above Lemma, $\theta_i$ is a feasible solution to the linear programs \eqref{eq:CI_lower} and \eqref{eq:CI_upper} with probability at least $1-2d\{s[\log((s-1)T)+2] + 2\} e^{1-s}$.

\end{proof}

\begin{proof}[Proof of Lemma~\ref{lma:bound_CI}]
Similar to the proof of Lemma 2 \cite{juditsky2020convex}, we will make use of Lemma 4 therein to prove our Lemma~\ref{lma:bound_CI} here. 

We choose $\gamma_t$ to be the $k$-th entry, $k \in [d]$, of $w_{t - \tau : t-1}$ normalized by $M_w$ such that it stays within $[0,1]$, i.e. $\gamma_i = (w_{t - \tau : t-1})_k/M_w$. Then, $\mu_t$ would be $g\left(w_{t - \tau : t-1}^\T \theta_i\right)$ and $\nu_t$ is the $k$-th entry of $a[W_T;i]$ normalized by $M_w$, i.e. $\nu_t=a[W_T;i]_k/M_w = \gamma_t y_t^{(i)}$. 
By Lemma~4 \cite{juditsky2020convex}, we complete the proof.
\end{proof}

\vspace{0.1in}

\noindent
{\it Extension to general non-linear link function.}
As for the general monotone non-linear link function, Lemma 4 still holds, but the vector field is no longer a linear transform of the coefficient vector. Nevertheless, $\theta_i$ would still be a feasible solution to the following optimization problem with high probability:

\begin{align}
    \theta_\ell[W_T,s;i] &=\min\left\{\begin{array}{ll}
a^\T \theta_i: & \theta_i \in \Theta, \\
& \psi_\ell \left(a[W_T;i]  ,T;s\right) \leq \frac{1}{TM_w}\sum_{t=1}^T (w_{t - \tau : t-1})_k g(w_{t - \tau : t-1}^\T \theta_i) \\
& \quad \quad  \quad \quad  \quad \quad  \quad \quad  \quad \quad  \quad \quad \leq \psi_u \left(a[W_T;i]  ,T;s\right), k \in [d_1].
\end{array}\right\} \label{eq:CI_lower_non_linear}\\
\theta_u[W_T,s;i] &=\max\left\{\begin{array}{ll}
a^\T \theta_i: & \theta_i \in \Theta, \\
& \psi_\ell \left(a[W_T;i]  ,T;s\right) \leq \frac{1}{TM_w}\sum_{t=1}^T (w_{t - \tau : t-1})_k g(w_{t - \tau : t-1}^\T \theta_i) \\
& \quad \quad  \quad \quad  \quad \quad  \quad \quad  \quad \quad  \quad \quad \leq \psi_u \left(a[W_T;i]  ,T;s\right), k  \in [d_1].
\end{array}\right\} \label{eq:CI_upper_non_linear}
\end{align}
However, even by assuming $g$ to be convex or concave, we cannot say the above optimization is convex. So direct generalization may be problematic. Nevertheless, we could use linear bounds on the non-linear link function such that we can still get an LP. Again, we take logistic regression as an example, where the link function is sigmoid function $g(t) =  \frac{1}{1+e^{-t}}$:
Let the upper and lower linear bound be $f_1(x) = a_1 x + b_1$ and $f_2(x) = a_2 x + b_2$, respectively. we have $-a_1M+b_1 = 1/(1+1+e^M)$ and $-a_1 (\log (1+\sqrt{1-4a_1}) - \log (1-\sqrt{1-4a_1}))+b_1 = (1+\sqrt{1-4a_1})/2$. Although we cannot give an analytic solution, we can numerically solve this equation and obtain $a_1, b_1$ and $a_2,b_2$.

\section{Additional Details for Numerical Simulation}\label{appendix:simulation}

\subsection{Additional model training details}\label{appendix:num_simu_set}


\subsubsection{Settings} 
We consider point process data on time horizon $T = 2000$ generated from a linear MHP with exponential decay kernel. We adopt a uniform decay rate of $0.9$. The background intensity and the self- and mutual-exciting magnitude are visualized as ground truth in Figure~\ref{fig1:setting1}. The time series data (which we use to fit the models) are constructed using the number of events observed at discrete time grids $\{1,\dots,T\}$. It is important to recognize this setting mimics the real EMR dataset --- although the vital signs and Lab results all have their exact occurrence times, they are binned together in an hourly manner. Through this experiment, we aim to show our model can output meaningful results even under the model misspecification. In addition, this experiment is also used to show why the point process model developed in our prior work \cite{wei2022granger} failed to handle such real-world time series data in terms of prediction. 

\subsubsection{Training details}
We use PGD, and the optimization package \texttt{Mosek} \cite{mosek} to train the model. As for benchmark procedures, we use the algorithms and the open source implementations in \cite{wei2022granger,khanna2019economy} for MHP and NN methods, respectively. In addition, we fit the ground truth model (i.e., linear MHP with exponential decay) on the original point process data. 

Here are detailed configurations of the training procedure: For the PGD approach to solve for our proposed estimator, we use $5 \times 10^{-3}$ as the initial learning rate and decrease it by half every $2000$ iteration (in total, there are $6000$ iterations). For simplicity, we assume we know that there is no inhibiting effect and perform PGD on the primal problem to solve for our proposed estimator, i.e. when learning for $i$-th node, $i = [d_1]$, we adopt feasible region $\RR_+^d$ and project negative entries back to zeros at each iteration in PGD; see detailed discussions in Sections~\ref{sec:linear_link_eg}, \ref{sec:decoupled_VI_estimate} and \ref{subsec:linear_DAG_penalized}.
Since there is no hyperparameter for the \texttt{Mosek} method, we do not give further instructions and refer readers to the documentation \cite{mosek} on how to fit the model.

\begin{figure*}[htbp]
\centering
  {%
    \subfigure[{Setting: point process data observed on discrete time grids.}
]{\label{fig:MHP_conv_ppdata}%
      \includegraphics[width = 0.45\textwidth]{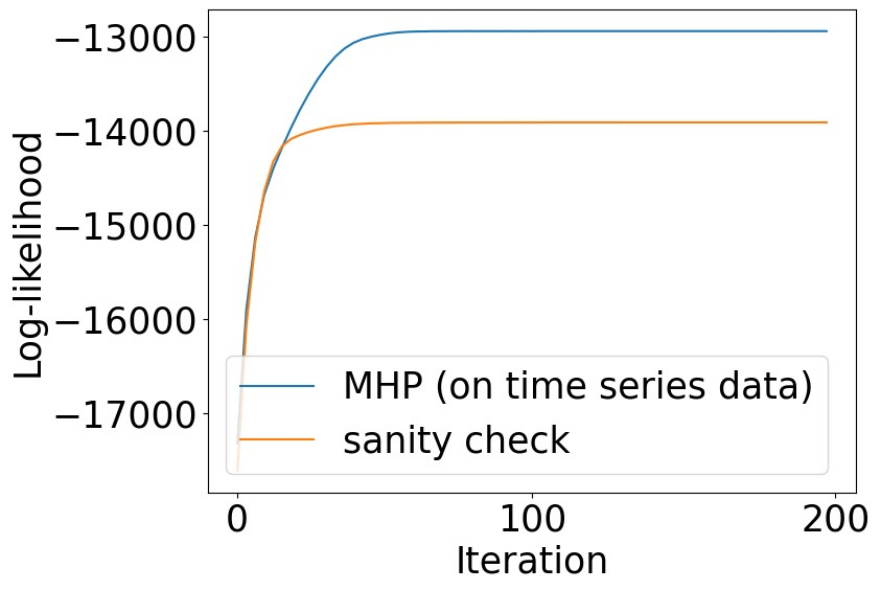} }%
      $ \ $
    \subfigure[Setting: time series data.]{\label{fig:MHP_conv_tsdata}%
      {
\includegraphics[width = 0.45\textwidth]{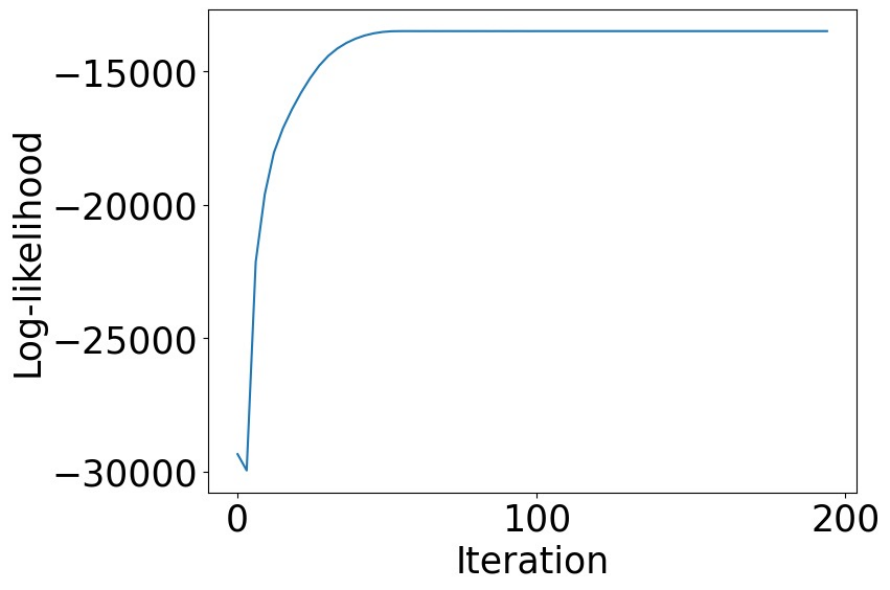} 
 }}
  }
  \caption{Simulation: evidence of convergence for MHP benchmark methods. The left panel corresponds to Figure~\ref{fig1:setting1} and Table~\ref{table:exp1_benchmark_comparison_setting1} and the right panel corresponds to Figure~\ref{fig1:setting2} and Table~\ref{table:exp1_benchmark_comparison_setting2}}\label{fig:MHP_conv}
\end{figure*}

For the MHP benchmark method, we only adopt the first phase (which is PGD) proposed in \cite{wei2022granger} to learn the parameters. We use $2\times 10^{-2}$ as the initial learning rate and decrease it by half every $100$ iterations (in total, there are $200$ iterations). We assume the decay parameter (i.e., the exponent in the exponential decay kernel) is known. In Figure~\ref{fig:MHP_conv} we plot the log-likelihood trajectories to verify the convergence.

For the NN-based method, we choose $10^{-3}$ as the initial learning rate in total $2000$ iterations. The hyperparemeters are selected using grid search $\mu_1 \in \{0.001,0.01,1\}, \ \mu_2 \in \{0.001,0.01,1,2\}, \ \mu_3 \in \{0.001,0.01,1,2\}$ and $F = 60$.

\subsection{Effectiveness of VI-based estimation}\label{sec:exp}

In the first part, we aim to show the good performance of \ref{VI_1} by comparing it with various benchmark estimators in terms of weight recovery metrics.
The benchmark methods considered here are linear MHP \cite{wei2022granger} and Neural Network (NN) based method \cite{khanna2019economy}.

\begin{figure*}[htbp]
\centering
  {\includegraphics[width = 0.9\textwidth]{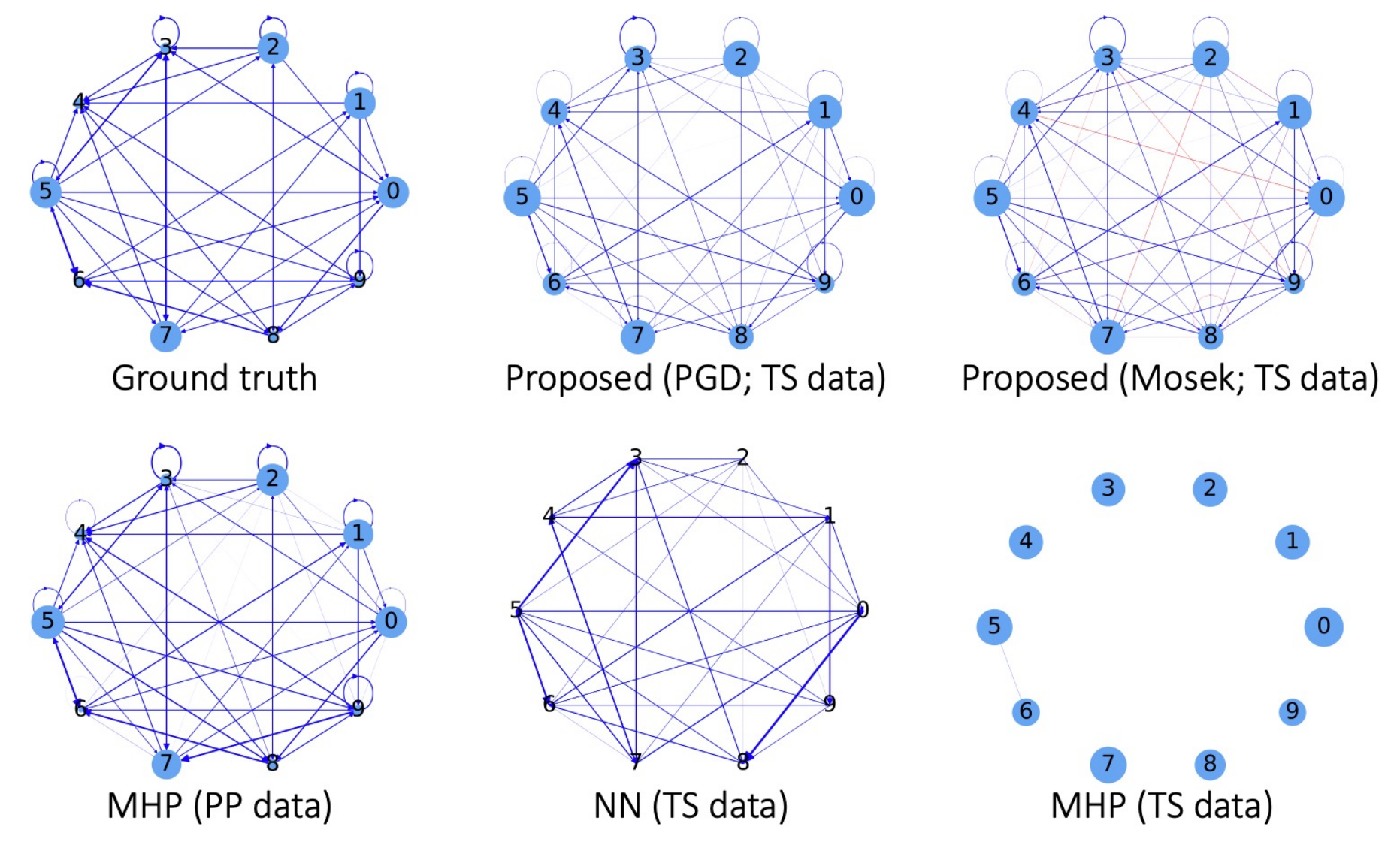}
  }
  \caption{Simulated example: demonstration of the effectiveness of \ref{VI_1}. We visualize the recovered (directed) self- and mutual-exciting graph and the background intensity: the size of the node is proportional to the background intensity, and the width of the edge is proportional to the exciting (blue) or inhibiting (red) effect magnitude. We specify the data on which we fit the model: point process (PP) or time series (TS) data.
  The NN-based method does not consider the background intensities and the self-exciting patterns, and MHP fails to recover the interactions among the nodes. In contrast, our proposed model is robust and can faithfully capture the graph structure even under model misspecification.}\label{fig1:setting1}
\end{figure*}

We visualize the recovered graphs in Figure~\ref{fig1:setting1} {and report quantitative metrics $\nu$ err. and $A$ err. in Table~\ref{table:exp1_benchmark_comparison_setting1}. Those results} demonstrate that our proposed model is capable of estimating the parameters accurately even under model misspecification and achieves the best performance among all benchmarks. {In particular, Table~\ref{table:exp1_benchmark_comparison_setting1} shows that} our proposed method has the smallest error among all methods fitted on the time series data. 
While the NN-based approach can capture most of the interactions among events, it does not capture the self-exciting patterns as well as the background intensities at all. 
There is an interesting finding for the MHP method: Although it can handle time series data, the loss of time granularity when converting raw point process data to time series data makes it unable to capture the true dynamics among the events --- it over-estimates the background intensities and fails to capture the interactions among those nodes, even under the correct model specification.
In contrast, our proposed discrete-time Hawkes network is much more robust and can recover the true dynamics accurately even under such a lack of granularity, which suggests its usefulness in the real-world EMR dataset.

\begin{table}[htbp]
\caption{Simulated example: quantitative metrics of the example in Figure~\ref{fig1:setting1}. We can observe our proposed \eqref{VI_1} achieves the best results among all models fitted on the time series data, suggesting the good and robust performance of \eqref{VI_1} under model misspecification.}\label{table:exp1_benchmark_comparison_setting1}
\begin{center}
\begin{small}
\resizebox{0.85\textwidth}{!}{%
\begin{tabular}{lccccc}
\toprule[1pt]\midrule[0.3pt]
Method & Proposed (PGD) & Proposed (Mosek) & NN & \multicolumn{2}{c}{{MHP}} \\
Data & time series & time series & time series & time series & point process \\
\cmidrule(l){2-6} 
  $\nu$ err. &  {\it .264} & .286 & $-$ & .425 & .0452 \\
$A$ err. &  {\it .308} & .323 & .345 & .629 & .130
 \\
\midrule[0.3pt]\bottomrule[1pt]
\end{tabular}
}
\end{small}
\end{center}
\end{table}



\subsubsection{Effectiveness of VI-based estimator: a sanity check}\label{appendix:num_simu_set_exp1_2}

\begin{figure*}[htbp]
\centering
  {\includegraphics[width = 0.9\textwidth]{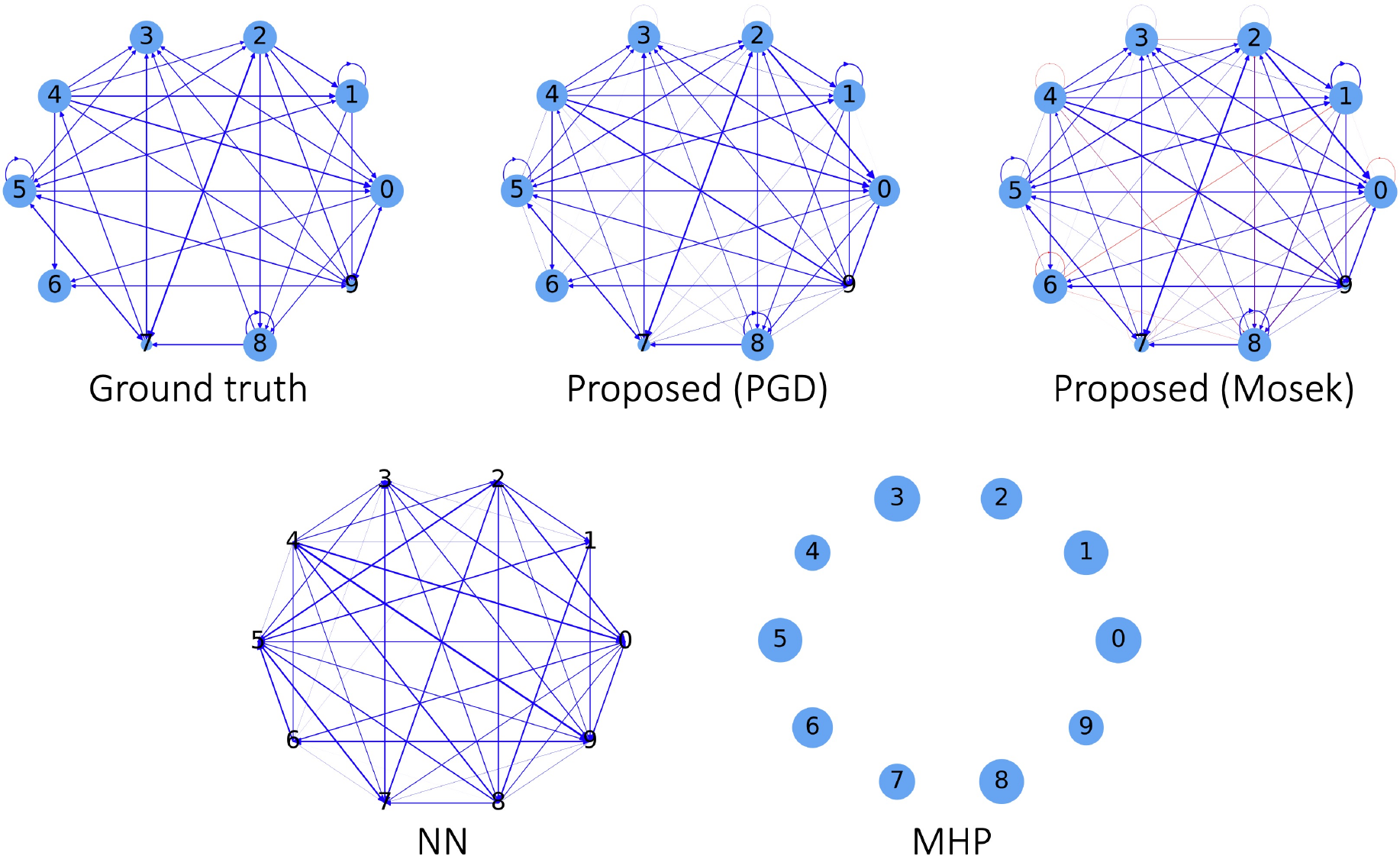}
  }
  \caption{Simulated example: a sanity check to demonstrate of the effectiveness of \ref{VI_1}. We visualize the recovered graph structures using our proposed VI-based estimator with linear link function as well as benchmark methods. We can observe that \ref{VI_1} achieves the best recovery under the correct model specification; in contrast, NN cannot capture self-excitation, and MHP completely misses the interactions.}\label{fig1:setting2}
\end{figure*}

We consider time series data on time horizon $T = 2000$ generated following our proposed linear model \eqref{eq:linear_model}. The background intensity and the self- and mutual-exciting magnitude are visualized as ground truth in Figure~\ref{fig1:setting2}. This setting serves as a sanity check, aiming to further demonstrate our proposed method works well under the correct model specification.

We visualize the recovered graphs in Figure~\ref{fig1:setting2} and report quantitative metrics in Table~\ref{table:exp1_benchmark_comparison_setting2}, where we can find \eqref{VI_1} achieves the best result among all benchmark methods.

\begin{table}[htp]
\caption{Simulated example: quantitative metrics of the example in Figure~\ref{fig1:setting2}. We can observe \ref{VI_1} achieves the best results under the correct model specification.}\label{table:exp1_benchmark_comparison_setting2}
\begin{center}
\begin{small}

\resizebox{0.6\textwidth}{!}{%
\begin{tabular}{lccccccccc}
\toprule[1pt]\midrule[0.3pt] 
Method & Proposed (PGD) & Proposed (Mosek) & NN & MHP \\
\cmidrule(l){2-5} 
  $\nu$ err. & {\it .054} & \textbf{.033} & $-$ & .514   \\
$A$ err. & \textbf{.106} & {\it .112} & .366
 & .594 \\
\midrule[0.3pt]\bottomrule[1pt]
\end{tabular}
}

\end{small}
\end{center}

\end{table}

\subsubsection{Effectiveness of PGD}\label{appendix:num_simu_set_exp2}
As mentioned in Section~\ref{sec:decoupled_VI_estimate}, PGD can solve for \ref{VI_1} with all monotone links. Despite the lack of theoretical guarantee, we use numerical evidence to show the good performance of PGD for all three link functions considered above: linear, sigmoid, and exponential links. 

\begin{figure*}[!htp]
\centering
{\includegraphics[width = \textwidth]{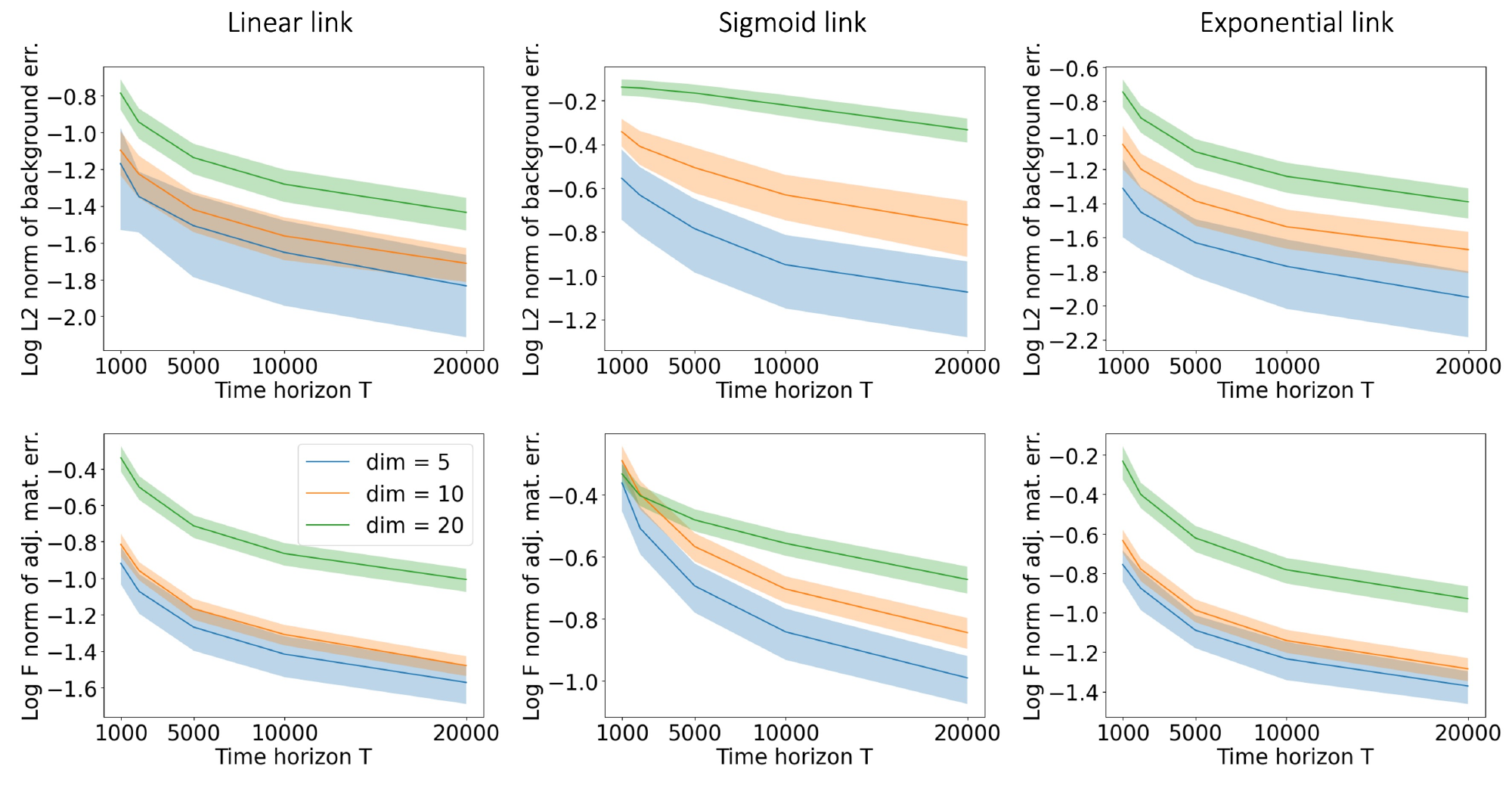}}
\caption{Simulation: mean (solid line) and standard deviation (shaded area) of estimation errors over $100$ independent trials. We plot the $\ell_2$ norm of the background intensity estimation error (the first row) and the matrix $F$-norm of the self- and mutual-exciting matrix estimation error (the second row). We can observe our proposed estimator obtained by PGD converges to the ground truth with increasing $T$ for all types of links and graph size (dim) considered here.}\label{fig2:PGD}
\end{figure*}

Here, we use PGD to obtain our proposed estimator; the training procedure is the same as Experiment 1 and can be found in Appendix~\ref{appendix:num_simu_set}. For each link function, we run $100$ independent experiments for $d_1 \in \{5, 10, 20\}$ over time horizons $T \in \{1000,2000,5000,10000,20000\}$ with randomly generated ground truth parameters. We plot the mean and standard deviation of the $\ell_2$ norm of the background intensity estimation error and the matrix $F$-norm of the self- and mutual-exciting matrix estimation error over those $100$ independent trials in Figure~\ref{fig2:PGD}. From this figure, we can observe that our proposed estimator obtained by PGD can converge to the ground truth with increasing $T$, which could serve as numerical evidence of the consistency of our proposed estimator. Moreover, the finite sample performances for the moderate-size graphs considered here are satisfying. This evidence shows the empirical success of PGD in obtaining our proposed estimator.

\subsection{Effectiveness of our data-adaptive linear regularization}\label{appendix:num_simu_set_exp3}
In this part of numerical simulation, the graph structure $A$ is again randomly generated. To ensure the DAG-ness, we use $A$ as initialization and apply vanilla GD to minimize the DAG characterization $h(A)$ and get $\tilde A$ such that $h({\tilde A}) = 0$.
Finally, we set the diagonal elements diag$(\tilde A) = $ diag$(A)$ to maintain those self-exiting components. 
Recall that we consider both linear and exponential link functions, where we can adopt the feasible region $\tilde \Theta = \RR_+^{d \times d_1}$; see the discussion in Section~\ref{subsec:linear_DAG_penalized} on how to relax the constraints and adopt this feasible region in linear link case. Therefore, after the GD update in each iteration, the projection back to $\hat \theta \in \RR_+^{d \times d_1}$ can be simply achieved by replacing all negative entries in $\hat \theta$ with zeros. Further details, such as the choice of learning rate $\eta$ and total iteration number, can be found in Appendix~\ref{appendix:num_simu_set} above.

The regularization strength hyperparameter $\lambda$ is selected from $\{1 \times 10^{-5}, 2 \times 10^{-5}, 3 \times 10^{-5}, 4 \times 10^{-5}, 5 \times 10^{-5}, 6 \times 10^{-5}, 7 \times 10^{-5}, 8 \times 10^{-5},
9 \times 10^{-5}, 1 \times 10^{-4}, 2 \times 10^{-4}, 3 \times 10^{-4}, 4 \times 10^{-4}, 5 \times 10^{-4}, 6 \times 10^{-4}, 7 \times 10^{-4},
8 \times 10^{-4}, 9 \times 10^{-4}, 1 \times 10^{-4}, 2 \times 10^{-4}, 3 \times 10^{-4}, 4 \times 10^{-4}, 5 \times 10^{-4}, 6 \times 10^{-4},
7 \times 10^{-4}, 8 \times 10^{-4}, 9 \times 10^{-4}, 1 \times 10^{-2}, 2 \times 10^{-2}, 3 \times 10^{-2}, 4 \times 10^{-2}, 5 \times 10^{-2},
6 \times 10^{-2}, 7 \times 10^{-2}, 8 \times 10^{-2}, 9 \times 10^{-2}, 1 \times 10^{-1} \}$. We choose the one with the smallest $A$ err. for each type of regularization.

\begin{figure*}[!htp]
\centering
{\includegraphics[width = .85\textwidth]{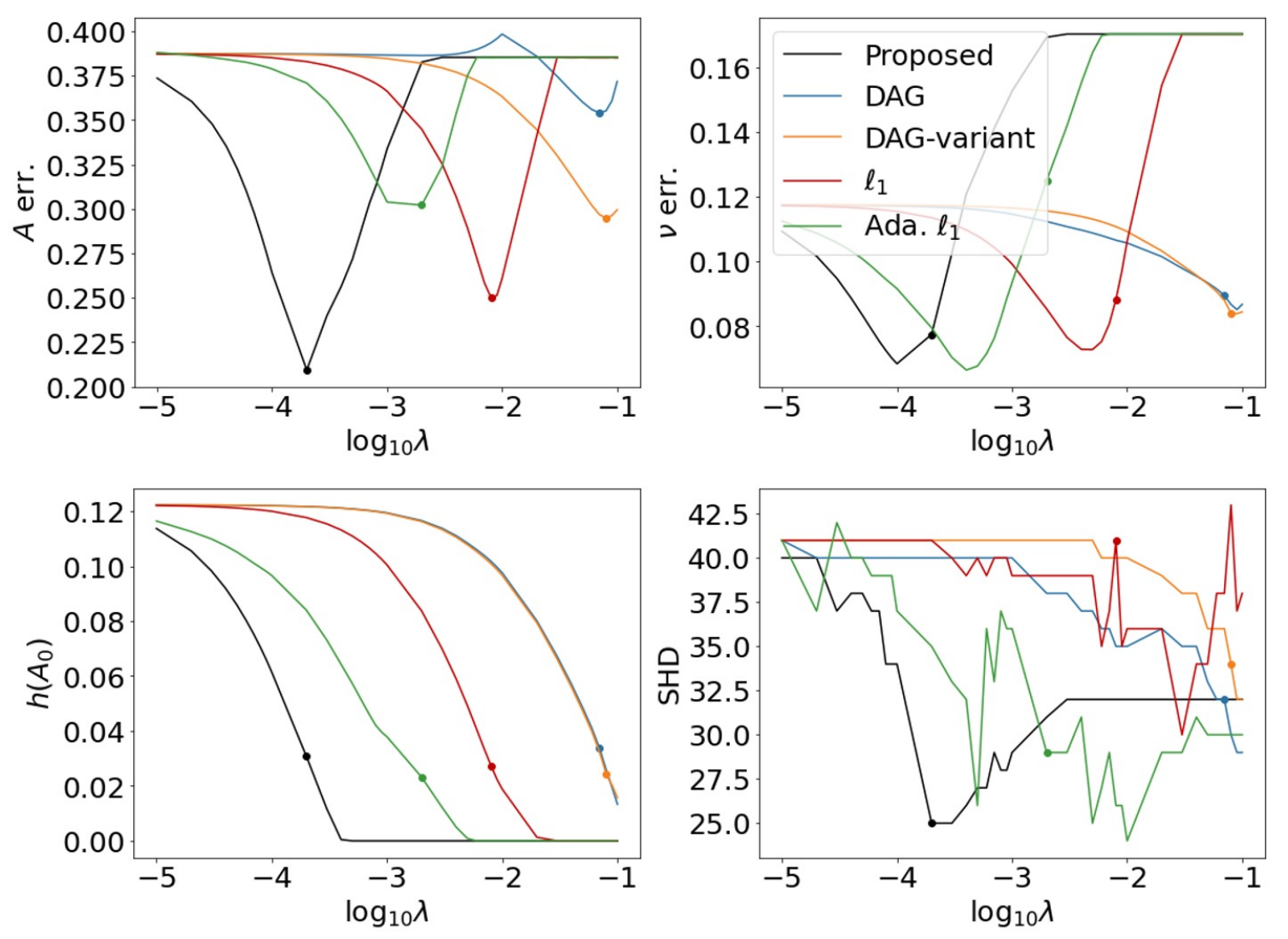}}
\caption{Simulation: the performance metrics with different hyperparameter $\lambda$ for the illustrating example in Figure~\ref{fig3:penalty_compare}. In our numerical simulation, $\lambda$ is selected to minimize $A$ err., which is marked with a dot on all panels for each type of regularization.}\label{fig3:penalty_compare_hyperparameter}
\end{figure*}

In the $d_1 = 10$ illustrative example, we plot the trajectories of all four metrics with varying hyperparameters in Figure~\ref{fig3:penalty_compare_hyperparameter}. We can observe from this figure that (i) this hyperparameter selection method does give the best structure estimation, i.e., the smallest SHD, for our proposed data-adaptive linear approach; (ii) our proposed data-adaptive linear approach gives the most accurate structure estimation for a wide range of hyperparameters if we consider same regularization strength. 


For this illustrative example, in addition to absolute errors reported in Table~\ref{table:exp3_penalty_compare}, we also report relative errors: For non-zero entries in the adjacency matrix $A$, the relative error is the absolute error divided by the true value; This is denoted as ``relative $A$ err. (non-zero entries)''. For zero entries in the adjacency matrix $A$, the absolute error is considered since the relative error is not well defined; This is denoted as ``absolute $A$ err. (zero entries)''. Slightly different for baseline intensity $\nu$, absolute errors are considered for small entries (nodes \#4 and \#9 have baseline intensity 0.02) and relative errors for the rest nodes (baseline intensity 0.2). The mean and standard deviation of the aforementioned metrics are reported in Table~\ref{table:exp3_penalty_compare_relative}.
    
\begin{table}[htp]
\caption{Relative errors of the results in Table~\ref{table:exp3_penalty_compare}.}
\label{table:exp3_penalty_compare_relative}
\vspace{0.05in}
\centering
\begin{small}
\resizebox{\textwidth}{!}{%
\begin{tabular}{lcccccc}
\toprule
& None & Proposed & DAG & DAG-Variant & $\ell_1$ & Ada. $\ell_1$ \\
\midrule
Relative $A$ err. (non-zero entries) & $.4958_{(.2990)}$ & $.5666_{(.3715)}$ & $.5683_{(.2999)}$ & $.5309_{(.2879)}$ & $.7085_{(.3006)}$ & $.8579_{(.2765)}$ \\
Absolute $A$ err. (zero entries) &  $.0277_{(.0490)}$ & $.0094_{(.0230)}$ & $.0181_{(.0397)}$ & $.0186_{(.0391)}$ & $.0090_{(.0272)}$ & $.0058_{(.0255)}$ \\
Relative $\nu$ err. (large entries) & $.1792_{(.1046)}$ & $.0608_{(.0706)}$ & $.1207_{(.0749)}$ & $.1255_{(.0681)}$ & $.0985_{(.0614)}$ & $.1486_{(.0600)}$ \\
Absolute $\nu$ err. (near-zero entries) & $.0040_{(.0003)}$ & $.0379_{(.0134)}$ & $.0268_{(.0078)}$ & $.0134_{(.0097)}$ & $.0408_{(.0094)}$ & $.0594_{(.0137)}$ \\
\bottomrule
\end{tabular}
}
\end{small}
\end{table}

Table~\ref{table:exp3_penalty_compare_relative} shows that $\ell_1$ regularization removes most of the edges to achieve the DAG-ness (which agrees with what we observed in Figure~\ref{fig3:penalty_compare}), and therefore it has much smaller absolute $A$ err. for zero entries and very large relative $A$ err. for non-zero entries. Our proposed regularization can help encourage DAG-ness by removing most of the false discoveries, achieving much smaller absolute $A$ err. for zero entries whereas maintaining the relative $A$ err. for non-zero entries at a reasonable level. That is why one can observe improved overall $A$ err. in Table~\ref{table:exp3_penalty_compare}.


Lastly, we report all four metrics in our experiments with $100$ trials in Table~\ref{table:exp2_penalty_comparison} for completeness. {It is not surprising to observe that the DAG regularization-based approaches do shrink $h(A_0)$ but fail to return an accurate graph structure (in terms of SHD); it is also not unexpected that adaptive $\ell_1$ approach achieves the smallest $h(A_0)$, since it is designed to output a highly-spare graph which in turn eliminates all cycles (the bottom right penal in Figure~\ref{fig3:penalty_compare}).
Overall, our proposed regularization achieves the best structural learning result in terms of SHD, and a very close weight recovery result to $\ell_1$ approach, suggesting its superior empirical performance over the aforementioned benchmarks.}

\begin{table}[htp]
\caption{Simulation: raw values of all metrics in Figure~\ref{fig3:penalty_compare_repeat}. We report the mean (and standard deviation) of all performance metrics over $100$ trials for different types of regularization. We report the $\ell_2$ norm of the background intensity estimation error ($\nu$ err.), the matrix $F$-norm of the self- and mutual-exciting matrix estimation error ($A$ err.), the DAG-ness measured by $h(A_0)$ and the Structural Hamming Distance (SHD). We can observe that our proposed method does the best in terms of structure recovery achieving almost the same performance with the best achievable weight recovery (by $\ell_1$ regularization-based approach).
}\label{table:exp2_penalty_comparison}

\begin{center}
\begin{small}
\resizebox{.9\textwidth}{!}{%
\begin{tabular}{lcccccc}
\multicolumn{7}{c}{\normalsize{Linear link function case.}} \\ 
\multicolumn{7}{c}{\small{Dimension $d_1 = 10$, Time Horizon $T = 500$.}} \\ 
\toprule[1pt]\midrule[.3pt] 
Regularization & None  & Proposed & DAG & DAG-Variant & $\ell_1$ & Ada. $\ell_1$ \\
\cmidrule(l){2-7} 
$A$ err. & $0.2510_{(0.0502)}$ & $0.2015_{(0.0374)}$ & $0.2372_{(0.0498)}$ & $0.2253_{(0.0518)}$ & $0.1968_{(0.0335)}$ & $0.2063_{(0.0403)}$ \\
$\nu$ err. & $0.1017_{(0.0300)}$ & $0.0758_{(0.0208)}$ & $0.0898_{(0.0286)}$ & $0.0808_{(0.0264)}$ & $0.0750_{(0.0188)}$ & $0.0774_{(0.0211)}$ \\
$h(A_0)$ & $0.0586_{(0.0190)}$ & $0.0183_{(0.0133)}$ & $0.0336_{(0.0253)}$ & $0.0051_{(0.0111)}$ & $0.0190_{(0.0129)}$ & $0.0142_{(0.0138)}$ \\
SHD & $36.45_{(5.32)}$ & $25.17_{(5.28)}$ & $30.52_{(7.99)}$ & $22.6_{(6.39)}$ & $33.41_{(5.02)}$ & $27.29_{(5.17)}$ \\
\midrule[.3pt]\bottomrule[1pt]
\end{tabular}
}

\end{small}
\end{center}


\begin{center}
\begin{small}

\resizebox{.9\textwidth}{!}{%
\begin{tabular}{lcccccc}
\multicolumn{7}{c}{\small{Dimension $d_1 = 20$, Time Horizon $T = 1000$.}} \\ 
\toprule[1pt]\midrule[.3pt] 
Regularization & None  & Proposed & DAG & DAG-Variant & $\ell_1$ & Ada. $\ell_1$ \\
\cmidrule(l){2-7} 
$A$ err. & $0.4403_{(0.0669)}$ & $0.1442_{(0.0408)}$ & $0.3879_{(0.0704)}$ & $0.3955_{(0.0694)}$ & $0.1293_{(0.0376)}$ & $0.1323_{(0.0402)}$ \\
$\nu$ err. & $0.1609_{(0.0365)}$ & $0.0656_{(0.0180)}$ & $0.1233_{(0.0290)}$ & $0.1316_{(0.0303)}$ & $0.0631_{(0.0168)}$ & $0.0567_{(0.0134)}$ \\
$h(A_0)$ & $0.0425_{(0.0106)}$ & $0.0011_{(0.0021)}$ & $0.0006_{(0.0002)}$ & $0.0006_{(0.0002)}$ & $0.0013_{(0.0024)}$ & $0.0002_{(0.0006)}$ \\
SHD & $187.78_{(11.80)}$ & $41.88_{(19.24)}$ & $119.49_{(8.96)}$ & $127.85_{(9.73)}$ & $124.03_{(33.37)}$ & $74.58_{(24.35)}$ \\
\midrule[.3pt]\bottomrule[1pt]
\end{tabular}
}

\end{small}
\end{center}

\vspace{.2in}

\begin{center}
\begin{small}

\resizebox{.9\textwidth}{!}{%
\begin{tabular}{lcccccc}
\multicolumn{7}{c}{\normalsize{Exponential link function case.}} \\ 
\multicolumn{7}{c}{\small{Dimension $d_1 = 10$, Time Horizon $T = 500$.}} \\ 
\toprule[1pt]\midrule[.3pt] 
Regularization & None  & Proposed & DAG & DAG-Variant & $\ell_1$ & Ada. $\ell_1$ \\
\cmidrule(l){2-7} 
$A$ err. & $0.3520_{(0.1038)}$ & $0.2550_{(0.0395)}$ & $0.3213_{(0.0937)}$ & $0.3136_{(0.1015)}$ & $0.2467_{(0.0345)}$ & $0.2681_{(0.0443)}$ \\
$\nu$ err. & $0.1125_{(0.0326)}$ & $0.0854_{(0.0231)}$ & $0.0929_{(0.0300)}$ & $0.0917_{(0.0280)}$ & $0.0835_{(0.0204)}$ & $0.0924_{(0.0241)}$ \\
$h(A_0)$ & $0.0793_{(0.0318)}$ & $0.0151_{(0.0169)}$ & $0.0247_{(0.0299)}$ & $0.0073_{(0.0161)}$ & $0.0164_{(0.0160)}$ & $0.0115_{(0.0151)}$ \\
SHD & $36.83_{(4.66)}$ & $24.18_{(5.13)}$ & $28.03_{(7.00)}$ & $24.16_{(5.56)}$ & $32.9_{(5.13)}$ & $26.78_{(5.49)}$ \\
\midrule[.3pt]\bottomrule[1pt]
\end{tabular}
}

\end{small}
\end{center}


\begin{center}
\begin{small}

\resizebox{.9\textwidth}{!}{%
\begin{tabular}{lcccccc}
\multicolumn{7}{c}{\small{Dimension $d_1 = 20$, Time Horizon $T = 1000$.}} \\ 
\toprule[1pt]\midrule[.3pt] 
Regularization & None  & Proposed & DAG & DAG-Variant & $\ell_1$ & Ada. $\ell_1$ \\
\cmidrule(l){2-7} 
$A$ err. & $0.5359_{(0.0743)}$ & $0.1552_{(0.0439)}$ & $0.4740_{(0.0806)}$ & $0.4832_{(0.0803)}$ & $0.1380_{(0.0365)}$ & $0.1505_{(0.0394)}$ \\
$\nu$ err. & $0.1778_{(0.0376)}$ & $0.0706_{(0.0151)}$ & $0.1355_{(0.0296)}$ & $0.1443_{(0.0302)}$ & $0.0680_{(0.0152)}$ & $0.0655_{(0.0133)}$ \\
$h(A_0)$ & $0.0583_{(0.0148)}$ & $0.0005_{(0.0011)}$ & $0.0007_{(0.0004)}$ & $0.0007_{(0.0004)}$ & $0.0008_{(0.0010)}$ & $0.0001_{(0.0004)}$ \\
SHD & $186.86_{(11.60)}$ & $35.1_{(15.61)}$ & $116.28_{(8.46)}$ & $124.51_{(8.93)}$ & $118.57_{(38.58)}$ & $66.15_{(26.05)}$ \\
\midrule[.3pt]\bottomrule[1pt]
\end{tabular}
}

\end{small}
\end{center}
\end{table}

\newpage

\section{Additional Details for Real Data Example}\label{appendix:realdata}

\subsection{Additional experimental details}\label{appendix:training_details}

\subsubsection{Definition of SADs}\label{appendix:lab_vital_SAD}
We consider in total 6 vital signs and 33 Lab results.
In Table~\ref{table:SADcutoff}, we report the raw patients' features/measurements used for SADs' construction. Since those measurement names explain themselves, and we do not give further descriptions of those measurements. In our previous study \cite{wei2021inferringb,wei2022granger}, the constructed SADs were shown to be appropriate by the fact that the Sepsis-3 cohort demonstrated a closer relationship with the SADs than the Non-Septic cohort and other real-data results therein. As in our previous study, a SAD is considered present if the corresponding measurements are outside of normal limits, which are determined jointly by already established criteria \cite{ABIM_ref_range} and expert opinions shown in Table~\ref{table:SADcutoff}. Here, our goal is to demonstrate that our proposed model can recover an explainable graph structure of those SADs while maintaining prediction accuracy. 

\begin{table}[!htp]
\caption{SAD construction based on thresholding observed vital signs and Lab results via medical knowledge. In our study, we incorporate in total 39 patient features, including 33 Lab results (L) and 6 vital signs (V).}\label{table:SADcutoff}
\begin{center}
\begin{small}
\resizebox{\textwidth}{!}{%
\begin{tabular}{llllc}
\toprule[1pt]\midrule[0.3pt]
Full SAD name & Abbreviation &  Type & Measurement name & Abnormal threshold  \\ 
\midrule[0.3pt]
\textbf{Renal Dysfunction} & RenDys & Lab & creatinine & $>1.3$ \\
& & & blood\_urea\_nitrogen\_(bun) & $>20$ \\
\cmidrule(l){4-5}
\textbf{Electrolyte Imbalance} & LyteImbal & Lab & calcium & $>10.5$ \\
& & & chloride & $<98$ or $>106$ \\
& & & magnesium & $<1.6$ \\
& & & potassium & $>5.0$ \\
& & & phosphorus & $>4.5$ \\
\cmidrule(l){4-5}
\textbf{Oxygen Transport Deficiency} & O2TxpDef & Lab & hemoglobin & $<12$ \\
\cmidrule(l){4-5}
\textbf{Coagulopathy} & Coag & Lab & partial\_prothrombin\_time\_(ptt) & $>35$ \\
& & & fibrinogen & $<233$ \\
& & & platelets & $<150000$ \\
& & & d\_dimer & $>0.5$ \\
& & & thrombin\_time & $>20$ \\
& & & prothrombin\_time\_(pt) & $>13$ \\
& & & inr & $>1.5$ \\
\cmidrule(l){4-5}
\textbf{Malnutrition} & MalNut & Lab & transferrin & $<0.16$ \\
& & & prealbumin & $<16$ \\
& & & albumin & $<3.3$ \\
\cmidrule(l){4-5}
\textbf{Cholestatsis} & Chole & Lab & bilirubin\_direct & $>0.3$ \\
& & & bilirubin\_total & $>1.0$ \\
\cmidrule(l){4-5}
\textbf{Hepatocellular Injury} & HepatoDys & Lab & aspartate\_aminotransferase\_(ast) & $>40$ \\
& & & alanine\_aminotransferase\_(alt) & $>40$ \\
& & & ammonia & $>70$ \\
\cmidrule(l){4-5}
\textbf{Acidosis} & Acidosis & Lab & base\_excess & $<-3$ \\
& & & ph & $<7.32$ \\
\cmidrule(l){4-5}
\textbf{Leukocyte Dysfunction} & LeukDys & Lab & white\_blood\_cell\_count & $<4$ or $>12$ \\
\cmidrule(l){4-5}
\textbf{Hypercarbia} & HypCarb & Lab & end\_tidal\_co2 & $>45$ \\
& & & partial\_pressure\_of\_carbon\_dioxide\_(paco2) & $>45$ \\
\cmidrule(l){4-5}
\textbf{Hyperglycemia} & HypGly & Lab & glucose & $>125$ \\
\cmidrule(l){4-5}
\textbf{Mycardial Ischemia} & MyoIsch & Lab & troponin & $>0.04$ \\
\cmidrule(l){4-5}
\textbf{Oxygen Diffusion Dysfunction (L)} & O2DiffDys (L)  & Lab & saturation\_of\_oxygen\_(sao2) & $<92$ \\
\cmidrule(l){4-5}
\textbf{Diminished Cardiac Output (L)} & DCO (L) & Lab & b-type\_natriuretic\_peptide\_(bnp) & $>100$ \\
\cmidrule(l){4-5}
\textbf{Tissue Ischemia} & TissueIsch & Lab & base\_excess & $<-3$ \\
& & & lactic\_acid & $>2.0$ \\
\cmidrule(l){4-5}
\textbf{Diminished Cardiac Output (V)} & DCO (V) & Vital signs & best\_map & $<65$ \\
\cmidrule(l){4-5}
\textbf{CNS Dysfunction} & CNSDys & Vital signs & gcs\_total\_score & $<14$ \\
\cmidrule(l){4-5}
\textbf{Oxygen Diffusion Dysfunction (V)} & O2DiffDys (V) & Vital signs & spo2 & $<92$ \\
& & & fio2 & $>21$ \\
\cmidrule(l){4-5}
\textbf{Thermoregulation Dysfunction} & ThermoDys & Vital signs & temperature & $<36$ or $>38$ \\
\cmidrule(l){4-5}
\textbf{Tachycardia} & Tachy & Vital signs & pulse & $>90$ \\
\midrule[0.3pt]\bottomrule[1pt]
\end{tabular}
}
\end{small}
\end{center}
\end{table}

\subsubsection{Evaluation metrics}\label{appendix:real_metric}
We first formally define the quantitative metrics:
for binary time series $y_1,\dots,y_T$ and predicted probabilities $p_1,\dots,p_T$, the CE loss and Focal loss are defined as:
\begin{align*}
    \text{CE loss} & = -\frac{1}{T} \sum_{i=1}^T y_i \cdot \log \left(p_i\right)+\left(1-y_i\right) \cdot \log \left(1-p_i\right), \\ 
    \text{Focal loss} & = -\frac{1}{T} \sum_{i=1}^T y_i \cdot \alpha(1-p_i)^\gamma \log (p_i) +\left(1-y_i\right) \cdot (1-\alpha) p_i^\gamma \log (1-p_i).
\end{align*}
Here, we choose $\alpha = 1/4$ and $\gamma = 4$ to account for class imbalance issues in Focal loss. For $p_i = 0, 1$ extreme cases, we replace the predicted probabilities with $\epsilon, 1- \epsilon$ with $\epsilon = 10^{-10}$ to ensure the metrics are well-defined.

\subsubsection{Training details}\label{appendix:real_train}
For our proposed method with various link functions, we leverage the prior knowledge that no inhibiting effect exists among SADs by projecting all negative iterates back to zeros when performing PGD on the primal problem, which is similar to what we have done in our numerical simulation. 
The learning rate and total iteration number hyperparameters for PGD are selected by performing a grid search on $\{.0002,.0004,\dots,.02\} \times \{1000,2000,5000,10000,20000\}$. We choose the learning rate to be $0.0094$ and the iteration number to be $2000$, which maximizes the total CE loss (which is the summation of average CE loss for all SADs) on the 2019 validation dataset. 

As for the benchmark XGBoost method, we use the well-developed package \texttt{xgboost} \cite{Chen:2016:XST:2939672.2939785} to fit the model. To be precise, we tune the hyperparameter using grid search over n\_estimators $\in \{10,20,50,500\}$, 
max\_depth $\in \{{\rm None}, 5, 20\}$,
learning\_rate $\in \{.0005,.001,.01\}$,
subsample $\in \{.3,.5\}$,
colsample\_bytree $\in \{.3,.5\}$,
colsample\_bylevel $\in \{.3,.5\}$ and
colsample\_bynode $\in \{.3,.5\}$.
Although there are other hyperparameters that we do not consider here, the main drawback of XGBoost is its interpretability. Therefore, even though it might perform well on the prediction tasks (which was already validated in the 2019 Physionet Challenge \cite{physionet2019}), it cannot provide the clinician on why an alarm is raised.

Lastly, we consider our VI-based estimation without prior knowledge (i.e., we allow potential inhibiting effects among SADs). For convenience, we use the well-developed package \texttt{Mosek} to learn the model parameters for the linear link case; as for the sigmoid link case, we use vanilla GD with learning rate $= 0.0094$ and iteration number $= 2000$.

\subsubsection{Bootstrap uncertainty quantification}\label{appendix:uncertainty}

The Bootstrap confidence intervals are obtained based on $1500$ Bootstrap samples. We sample $409$ sepsis patients with equal probability with replacement from the original $409$ sepsis patients in the year 2018 data; similarly, we draw $1369$ non-sepsis patients uniformly with replacement from the original $1369$ non-sepsis patients in the year 2018 data. For each Bootstrap sample we obtained, we fit our proposed model. 
When we leverage the prior knowledge that no inhibiting effect exists, i.e., the PGD-based approach, we use the $5\%$-percentile and $100\%$-percentile of the Bootstrap results as the $95\%$ CI; otherwise, we use the $2.5\%$-percentile and $97.5\%$-percentile of the Bootstrap results as the $95\%$ CI. If the Bootstrap CI contains zero, we then claim that such an edge does not exist and eliminate it in the graph. For those edges that exist, we use the median of the Bootstrap results as the weight and plot the graph. A similar method applies to the background intensities.

\subsection{Additional results}

\subsubsection{Comparison with XGBoost}\label{appendix:AUROC}
As shown in Table~\ref{table:CE_1}, XGBoost, as a black-box algorithm, does achieve smaller CE loss for CNS Dysfunction, but such improvement is again marginal compared with our chosen model above. The reason why this powerful XGBoost fails to outperform our method is two-fold: on one hand, it has so many hyperparameters and we only fine-tune it on relatively small grids (due to high computational cost); on the other hand, since SADs occurrences are rare, CE loss may not be a perfect evaluation metric compared with the Focal loss. To address the second issue, we report the Focal loss for all models with the corresponding selected hyperparameters in Table~\ref{table:FL} in Appendix~\ref{appendix:FL}, from which we do observe that XGBoost performs the best for most SADs. Nevertheless, such a black-box method cannot output interpretable (causal) graphs for SADs for scientific discovery.

We fit those aforementioned candidate models (see Section~\ref{sec:realexp_comparison}) to the real data in the year 2018. 
To quantitatively compare those models, we calculate the in-sample Area Under the Receiver Operating Characteristic (AUROC) using the training data (the year 2018) and the out-of-sample AUROC using testing data (the year 2019) for each SAD as the response variable (which is a bit different from our previous train-validation-test split). We report the in-sample and the out-of-sample AUROCs in Table~\ref{table:AUC}. 

From Table~\ref{table:AUC}, we can observe that our proposed model with linear link function achieves the best performance for predicting most SADs. Although the sigmoid link function seems a better choice in some cases, the improvement compared to the linear link function is marginal, except for predicting Oxygen Diffusion Dysfunction (Lab results). XGBoost, as a black-box algorithm, does achieve great performance in the 2018 training data but fails to generalize to 2019 testing data, i.e., its out-of-sample AUROCs are much smaller than the in-sample ones for nearly all cases. We need to remark that selecting hyperparameters from finer grids may lead to better out-of-sample performance for XGBoost, but such a grid search approach is too time-consuming since there are too many hyperparameters to tune, and the results remain inexplicable.

\begin{table}[htp]
\caption{Comparison of AUROCs for our proposed method with linear link function, Sigmoid link function, and the benchmark XGBoost method. For our proposed method with linear link function, we use \texttt{Mosek} and PGD to obtain our proposed estimator. We report the in-sample AUROC using the training data (the year 2018) and the out-of-sample AUROC using testing data (the year 2019). The best out-of-sample AUROCs are highlighted. We can observe \ref{VI_1} coupled with linear link achieves the best out-of-sample prediction accuracy for most SADs.}\label{table:AUC}
\begin{center}
\begin{small}
\resizebox{0.75\textwidth}{!}{%
\begin{tabular}{lcccccccccc}
\toprule[1pt]\midrule[0.3pt]
Resp. SAD & \multicolumn{2}{c}{{Linear (Mosek)}} & \multicolumn{2}{c}{{Linear (PGD)}} & \multicolumn{2}{c}{{Sigmoid}} & \multicolumn{2}{c}{{XGBoost}} & Occurrence \\ 
 Abbrev. & 2018 & 2019 & 2018 & 2019 & 2018 & 2019 & 2018 & 2019 &  ratio (2018)\\
\cmidrule(l){2-3} \cmidrule(l){4-5} \cmidrule(l){6-7} \cmidrule(l){8-9} \cmidrule(l){10-10}
RenDys & .642 & .607 & .596 & .586 & .650 & \textbf{.608} & .612 & .543 & 0.0207\\
LyteImbal & .617 & \textbf{.602} & .594 & .569 & .612 & .592 & .628 & .522 & 0.0366\\
O2TxpDef & .655 & \textbf{.639} & .649 & .634 & .657 & .637 & .660 & .538 & 0.0481\\
DCO (L) & .763 & \textbf{.692} & .743 & .669 & .645 & .500 & .500 & .500 & 0.002\\
DCO (V) & .838 & \textbf{.805} & .823 & .796 & .837 & .801 & .812 & .726 & 0.0502\\
CNSDys & .964 & \textbf{.968} & .961 & .965 & .964 & .967 & .936 & .945 & 0.3702\\
Coag & .596 & .566 & .542 & .519 & .600 & \textbf{.576} & .612 & .558 & 0.0629\\
MalNut & .641 & .612 & .620 & .604 & .632 & \textbf{.615} & .609 & .530 & 0.0237\\
Chole & .638 & .571 & .633 & .570 & .647 & \textbf{.573} & .574 & .502 & 0.0065\\
HepatoDys & .622 & \textbf{.658} & .587 & .635 & .614 & .649 & .516 & .500 & 0.0126\\
O2DiffDys (L) & .870 & .729 & .849 & .795 & .809 & \textbf{.903} & .500 & .500 & 0.0007\\
O2DiffDys (V) & .830 & \textbf{.795} & .829 & \textbf{.795} & .828 & .778 & .870 & .680 & 0.0397\\
Acidosis & .771 & .796 & .773 & \textbf{.800} & .773 & .786 & .648 & .521 & 0.0109\\
ThermoDys & .982 & \textbf{.999} & .987 & \textbf{.999} & .989 & \textbf{.999} & .649 & .452 & 0.9983\\
Tachy & .921 & \textbf{.916} & .920 & \textbf{.916} & .921 & .914 & .843 & .830 & 0.4769\\
LeukDys & .639 & .604 & .613 & .590 & .635 & \textbf{.617} & .623 & .520 & 0.0216\\
HypCarb & .752 & .780 & .744 & \textbf{.788} & .741 & .734 & .634 & .516 & 0.0086\\
HypGly & .735 & \textbf{.722} & .731 & .717 & .739 & \textbf{.722} & .701 & .602 & 0.1028\\
MyoIsch & .787 & .712 & .765 & \textbf{.723} & .725 & .659 & .614 & .529 & 0.0074\\
TissueIsch & .775 & \textbf{.784} & .775 & \textbf{.784} & .759 & .778 & .690 & .542 & 0.0164\\
Sepsis & .653 & \textbf{.641} & .632 & .626 & .658 & .637 & .673 & .566 & -\\
\midrule[0.3pt]\bottomrule[1pt]
\end{tabular}
}
\end{small}
\end{center}
\end{table}

Additionally, we report the 2018 training data (im)balancesness for each SAD in Table~\ref{table:AUC}. One interesting finding is that the out-of-sample AUROC is better (greater than $0.9$) when the occurrence ratio is large enough (greater than 0.3); Those SADs are CNSDys, ThermoDys, Tachy.

\subsubsection{Causal graphs using non-linear link functions}\label{appendix:nonlinearlink}

\begin{figure*}[!htp]
\centerline{
\includegraphics[width = \textwidth]{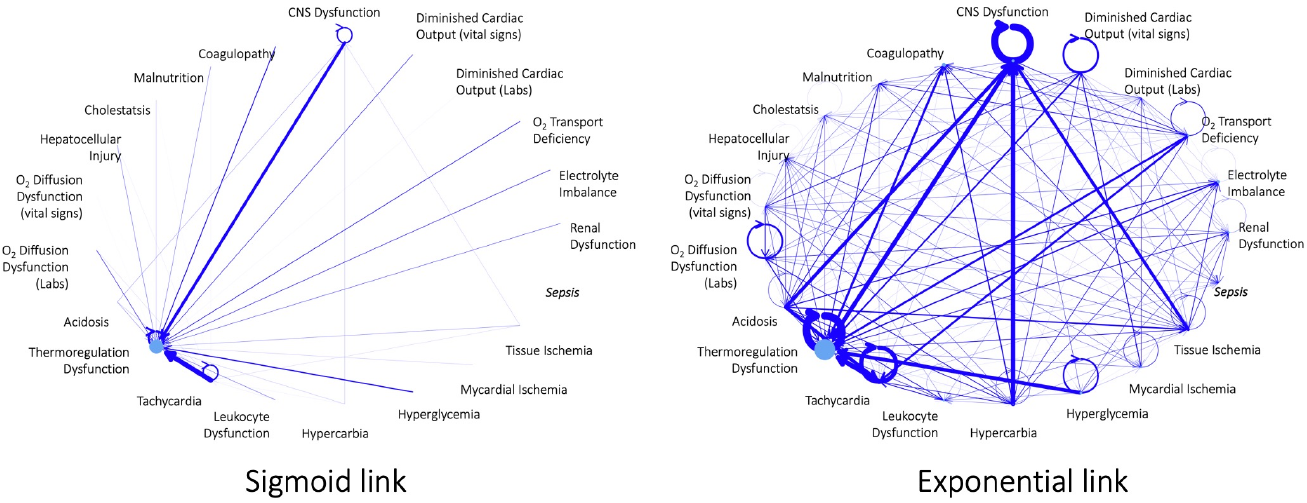}
}
\caption{Causal graphs for SADs obtained via discrete-time Hawkes network using various link functions. The node's size is proportional to the background intensity, and the width of the directed edge is proportional to the exciting effect magnitude. 
The quantitative comparison among three links in Tables~\ref{table:CE_1} and \ref{table:FL} tells us the linear link function performs the best on our real dataset; although sigmoid link function is commonly used for predicting binary variables in practice, we can observe that the resulting graph is not informative enough to uncover the interactions among SADs, let alone it has a rather poor quantitative performance.
}
\label{fig:GC_graphs_various_link} 
\end{figure*}

For completeness purpose, we report the causal graphs recovered using VI-based methoc coupled with non-linear link functions is Figure~\ref{fig:GC_graphs_various_link}.

\subsubsection{Causal graphs with potential inhibiting effects}\label{appendix:result_inhibiting}

As mentioned above, we consider our proposed discrete-time Hawkes network with potential inhibiting effects using linear and sigmoid link functions. For completeness, we also perform Bootstrap UQ for both links.

\begin{figure*}[!htp]
\centerline{
\includegraphics[width = \textwidth]{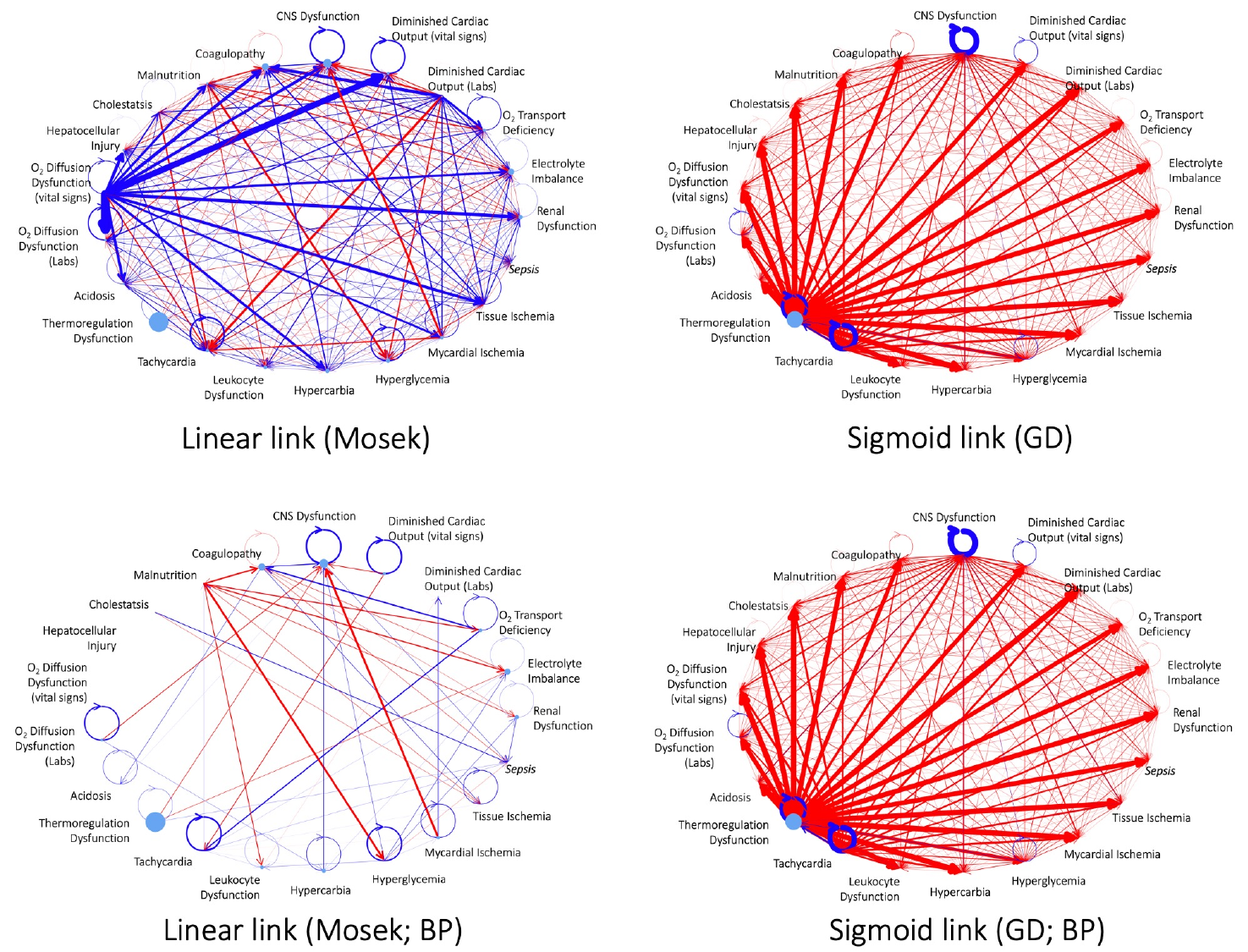} }
\caption{Causal graphs recovered by our proposed discrete-time Hawkes network with inhibiting effects. The size of the node is proportional to the background intensity, and the width of the directed edge is proportional to the exciting (blue) or inhibiting (red) effect magnitude. Mosek and GD represent the method to obtain the estimators, which are different from the PGD approach when we consider prior medical knowledge that no inhibiting effects exist. Compared with Figure~\ref{fig:GC_graphs_various_link}, we can observe that prior medical knowledge that no inhibiting effect exists can help output a more sparse and interpretable graph structure.
}
\label{fig:GC_graph_inhibiting} 
\end{figure*}

We report the CE loss and Focal loss for our discrete-time Hawkes network with potential inhibiting effects in Tables~\ref{table:CE_3_inhibiting} and \ref{table:FL}. Moreover, we plot the resulting causal graphs in Figure~\ref{fig:GC_graph_inhibiting}. From those results, we can see the Bootstrap UQ does not affect the sigmoid link result for most of SADs (except for CNSDys and Tachy).

\begin{table}[htp]
\caption{The average and standard deviation of Cross Entropy loss over all patients in the 2019 test dataset for our discrete-time Hawkes network with potential inhibiting effects. Compared with the first column in Table~\ref{table:CE_1}, we observe that allowing potential inhibiting does reduce the CE loss for most SADs; however, such loss reduction is marginal. Besides, we can see the Bootstrap UQ does not affect the sigmoid link result for most of SADs (except for CNSDys and Tachy). 
}\label{table:CE_3_inhibiting}
\begin{center}
\begin{small}
\resizebox{0.7\textwidth}{!}{%
\begin{tabular}{lcccc}
\toprule[1pt]\midrule[0.3pt]
&  Linear link  &  Linear link (BP)  &  Sigmoid link  &  Sigmoid link (BP) \\
\cmidrule(l){2-5}
RenDys & .1236 $_{(.2544)}$ & .1236 $_{(.2326)}$ & .6198 $_{(1.5365)}$ & .6198 $_{(1.5365)}$ \\
LyteImbal & .1758 $_{(.2573)}$ & .1807 $_{(.2878)}$ & 1.0075 $_{(1.9436)}$ & 1.0075 $_{(1.9436)}$ \\
O2TxpDef & .1921 $_{(.2312)}$ & .1950 $_{(.2518)}$ & 1.1341 $_{(1.8396)}$ & 1.1341 $_{(1.8396)}$ \\
DCO (L) & .0131 $_{(.0886)}$ & .0144 $_{(.1075)}$ & .0399 $_{(.3269)}$ & .0399 $_{(.3269)}$ \\
DCO (V) & .1462 $_{(.2930)}$ & .1455 $_{(.3009)}$ & .9865 $_{(2.7843)}$ & .9865 $_{(2.7843)}$ \\
CNSDys & .2703 $_{(.3080)}$ & .2718 $_{(.3076)}$ & 2.145 $_{(4.2959)}$ & 2.1474 $_{(4.3007)}$ \\
Coag & .2434 $_{(.2032)}$ & .2477 $_{(.1886)}$ & 1.5349 $_{(1.8136)}$ & 1.5349 $_{(1.8136)}$ \\
MalNut & .1264 $_{(.2153)}$ & .2076 $_{(.4672)}$ & .6470 $_{(1.3905)}$ & .6470 $_{(1.3905)}$ \\
Chole & .0489 $_{(.1692)}$ & .1985 $_{(.8692)}$ & .1985 $_{(.8692)}$ & .1985 $_{(.8692)}$ \\
HepatoDys & .0836 $_{(.2057)}$ & .3943 $_{(1.1954)}$ & .3943 $_{(1.1954)}$ & .3943 $_{(1.1954)}$ \\
O2DiffDys (L) & .0052 $_{(.0841)}$ & .0062 $_{(.1119)}$ & .0141 $_{(.2543)}$ & .0141 $_{(.2543)}$ \\
O2DiffDys (V) & .1489 $_{(.3035)}$ & .1526 $_{(.3312)}$ & 1.018 $_{(2.7288)}$ & 1.018 $_{(2.7288)}$ \\
Acidosis & .0609 $_{(.1832)}$ & .0728 $_{(.2891)}$ & .3347 $_{(1.3496)}$ & .3347 $_{(1.3496)}$ \\
ThermoDys & .0066 $_{(.0474)}$ & .0065 $_{(.0483)}$ & .0306 $_{(.4780)}$ & .0306 $_{(.4780)}$ \\
Tachy & .3738 $_{(.3679)}$ & .5262 $_{(.9272)}$ & 2.684 $_{(4.3505)}$ & 2.6769 $_{(4.3359)}$ \\
LeukDys & .1242 $_{(.2352)}$ & .1246 $_{(.2333)}$ & .6385 $_{(1.515)}$ & .6385 $_{(1.515)}$ \\
HypCarb & .0522 $_{(.1780)}$ & .0639 $_{(.1378)}$ & .2442 $_{(1.0791)}$ & .2442 $_{(1.0791)}$ \\
HypGly & .2883 $_{(.3134)}$ & .4607 $_{(.7418)}$ & 2.274 $_{(3.5716)}$ & 2.274 $_{(3.5716)}$ \\
MyoIsch & .0582 $_{(.2281)}$ & .0596 $_{(.1966)}$ & .2746 $_{(1.265)}$ & .2746 $_{(1.265)}$ \\
TissueIsch & .0783 $_{(.2025)}$ & .0809 $_{(.2316)}$ & .4139 $_{(1.4378)}$ & .4139 $_{(1.4378)}$ \\
SEP3 & .1307 $_{(.1802)}$ & .1316 $_{(.1899)}$ & .6915 $_{(1.2001)}$ & .6915 $_{(1.2001)}$ \\
\midrule[0.3pt]\bottomrule[1pt]
\end{tabular}
}
\end{small}
\end{center}
\end{table}

Figure~\ref{fig:GC_graph_inhibiting} illustrates the resulting causal graphs of SAD nodes that follow closely with known and expected causal relationships in sepsis-related illness. 
For example, the findings that Renal Dysfunction promotes sepsis events or that Coagulopathy strongly promotes Tissue Ischemia (after the Bootstrap UQ) are well-appreciated relationships. 
Despite the popularity of sigmoid link function in practice, it outputs a graph with nearly all inhibiting effects among SADs, from which it is difficult to extract any useful information (the right two panels).

However, even for the interpretable graphs (the left two panels), not using prior knowledge results in relationships that suggest certain SADs actually have inhibitory effects. While there are possible explanations for these inhibitory effects, they are not commonly known or expected from a physiologic perspective. Overall, the graphs that incorporate physicians' prior knowledge demonstrate a more meaningful series of relationships consistent with known physiologic responses to sepsis (see Figures~\ref{fig:GC_graph_bp_reg}, \ref{fig:GC_graphs_various_link} and \ref{fig:GC_graph_bp}). 

Moreover, the comparison between the left two panels further demonstrates the usefulness of Bootstrap UQ in helping output a more sparse and interpretable graph structure (bottom left panel). 
From a physiologic perspective, one would not normally expect a SAD to inhibit another SAD. While there could be unappreciated physiologic relationships or biases in the data that explain such an inhibitory relationship, these inhibitory relationships disappear (as seen in the bottom left in Figure~\ref{fig:GC_graph_inhibiting}) after Bootstrap UQ, suggesting that they are not meaningful and should be eliminated. The elimination of inhibitory relationships further highlights the importance of Bootstrap UQ.

\subsubsection{The Focal loss table}\label{appendix:FL}
Here, we report the Focal loss results for all aforementioned methods in Table~\ref{table:FL}. Since Focal loss takes the class imbalance issue into account, it is not surprising that the powerful XGBoost performs all other candidates for almost all SADs. However, it is good to observe that our proposed discrete-time Hawkes network coupled with linear link function performs the best for the rest of the SADs, especially when we allow potential inhibiting effects. Most importantly, it is good to observe that our proposed discrete-time Hawkes network coupled with Bootstrap UQ and our proposed data-adaptive linear regularization can achieve almost the same Focal loss with the best achievable result for most of the SADs, suggesting its practical usefulness in uncovering a reliable and interpretable causal DAG. 

\begin{table}[htp]
\caption{Comparison among all aforementioned methods: we report the average and standard deviation of the Focal loss over all patients in the 2019 test dataset for our proposed discrete-time Hawkes network coupled with various link functions as well as the black box XGBoost method. In this table, $-$ represents ``no applicable''; \faTimes \ means we do not apply Bootstrap UQ, regularization or we do not allow potential inhibiting effects in the corresponding method, and \faCheck \ stands for the opposite meaning.
}\label{table:FL}
\begin{center}
\begin{small}
\resizebox{\textwidth}{!}{%
\begin{tabular}{lccccccccccccc}
\toprule[1pt]\midrule[0.3pt]
 Method &  Linear link  &  Exp. link  &  Logit link  &  XGBoost  &  Linear link   &  Linear link   &  Linear link   &  Linear link   &  Linear link   &  Linear link  &  Linear link   &  Sigmoid link  &  Sigmoid link  \\
 Bootstrap UQ &  \faTimes  &  \faTimes  &  \faTimes  &  $-$  &  \faCheck  &  \faCheck  &  \faCheck  &  \faCheck  &  \faCheck  &  \faTimes  &  \faCheck  &  \faTimes  &  \faCheck \\
 Regularization &  \faTimes  &  \faTimes  &  \faTimes  &  $-$  &  \faTimes  &   Proposed  &  $\ell_1$  &  DAG  &  DAG-variant  &  \faTimes  &  \faTimes  &  \faTimes  & \faTimes  \\
 Inhibiting effects &  \faTimes  &  \faTimes  & \faTimes  &  $-$  &  \faTimes  &  \faTimes  &  \faTimes  &  \faTimes  &  \faTimes &  \faCheck  &  \faCheck  &  \faCheck  &  \faCheck \\
 \cmidrule(l){2-14}
RenDys & .0236 $_{(.0593)}$ & .0235 $_{(.0591)}$ & .1550 $_{(.3841)}$ & \textbf{.0185}$_{(.0279)}$ & .0561 $_{(.1732)}$ & .0244 $_{(.0612)}$ & .0238 $_{(.0591)}$ & .0254 $_{(.0628)}$ & .0561 $_{(.1732)}$ & .0231 $_{(.0579)}$ & .0204 $_{(.0515)}$ & .1550 $_{(.3841)}$ & .1550 $_{(.3841)}$ \\
LyteImbal & .0296 $_{(.0558)}$ & .0293 $_{(.0553)}$ & .2519 $_{(.4859)}$ & \textbf{.0156}$_{(.0169)}$ & .0324 $_{(.0611)}$ & .0311 $_{(.0590)}$ & .0308 $_{(.0577)}$ & .0327 $_{(.0608)}$ & .0324 $_{(.0611)}$ & .0288 $_{(.0540)}$ & .0325 $_{(.0634)}$ & .2519 $_{(.4859)}$ & .2519 $_{(.4859)}$ \\
O2TxpDef & .0272 $_{(.0419)}$ & .0286 $_{(.0441)}$ & .2835 $_{(.4599)}$ & \textbf{.0177}$_{(.0240)}$ & .1186 $_{(.2088)}$ & .0292 $_{(.0446)}$ & .0294 $_{(.0450)}$ & .0332 $_{(.0537)}$ & .1186 $_{(.2088)}$ & .0290 $_{(.0450)}$ & .0333 $_{(.0516)}$ & .2835 $_{(.4599)}$ & .2835 $_{(.4599)}$ \\
DCO (L) & .0027 $_{(.0224)}$ & .0027 $_{(.0222)}$ & .0100 $_{(.0817)}$ & \textbf{.0016}$_{(.0108)}$ & .0076 $_{(.0638)}$ & .0029 $_{(.0241)}$ & .0027 $_{(.0223)}$ & .0080 $_{(.0654)}$ & .0076 $_{(.0638)}$ & .0026 $_{(.0220)}$ & .0033 $_{(.0268)}$ & .0100 $_{(.0817)}$ & .0100 $_{(.0817)}$ \\
DCO (V) & .0185 $_{(.0467)}$ & .0186 $_{(.0465)}$ & .2466 $_{(.6961)}$ & .0345 $_{(.0950)}$ & .0662 $_{(.1750)}$ & .0189 $_{(.0459)}$ & .0183 $_{(.0446)}$ & .0321 $_{(.0905)}$ & .0662 $_{(.1750)}$ & \textbf{.0183}$_{(.0473)}$ & .0204 $_{(.0499)}$ & .2466 $_{(.6961)}$ & .2466 $_{(.6961)}$ \\
CNSDys & .0378 $_{(.1887)}$ & .7078 $_{(2.7696)}$ & .0987 $_{(.2518)}$ & .0390 $_{(.1299)}$ & .0347 $_{(.1676)}$ & .0328 $_{(.1447)}$ & .0348 $_{(.1597)}$ & \textbf{.0116}$_{(.0288)}$ & .0347 $_{(.1676)}$ & .0186 $_{(.0470)}$ & .0185 $_{(.0465)}$ & .5868 $_{(1.2367)}$ & .5873 $_{(1.2377)}$ \\
Coag & .0347 $_{(.0389)}$ & .0339 $_{(.0380)}$ & .3837 $_{(.4534)}$ & \textbf{.0238}$_{(.0217)}$ & .0389 $_{(.0454)}$ & .0369 $_{(.0436)}$ & .0358 $_{(.0423)}$ & .0360 $_{(.0424)}$ & .0389 $_{(.0454)}$ & .0345 $_{(.0397)}$ & .0286 $_{(.0356)}$ & .3837 $_{(.4534)}$ & .3837 $_{(.4534)}$ \\
MalNut & .0225 $_{(.0447)}$ & .0224 $_{(.0443)}$ & .1618 $_{(.3476)}$ & \textbf{.0163}$_{(.0306)}$ & .0243 $_{(.0497)}$ & .0254 $_{(.0541)}$ & .0247 $_{(.0531)}$ & .0288 $_{(.0575)}$ & .0243 $_{(.0497)}$ & .0228 $_{(.0478)}$ & .0488 $_{(.1153)}$ & .1618 $_{(.3476)}$ & .1618 $_{(.3476)}$ \\
Chole & .0102 $_{(.0407)}$ & .0102 $_{(.0406)}$ & .0496 $_{(.2173)}$ & .0114 $_{(.0377)}$ & .0163 $_{(.0660)}$ & .0163 $_{(.0660)}$ & .0106 $_{(.0458)}$ & .0239 $_{(.0949)}$ & .0163 $_{(.0660)}$ & \textbf{.0100}$_{(.0401)}$ & .0496 $_{(.2173)}$ & .0496 $_{(.2173)}$ & .0496 $_{(.2173)}$ \\
HepatoDys & .0164 $_{(.0474)}$ & .0163 $_{(.0471)}$ & .0986 $_{(.2988)}$ & \textbf{.0112}$_{(.0237)}$ & .0789 $_{(.2391)}$ & .0190 $_{(.0576)}$ & .0175 $_{(.0527)}$ & .0189 $_{(.0572)}$ & .0789 $_{(.2391)}$ & .0164 $_{(.0473)}$ & .0986 $_{(.2988)}$ & .0986 $_{(.2988)}$ & .0986 $_{(.2988)}$ \\
O2DiffDys (L) & .0017 $_{(.0369)}$ & .0021 $_{(.0457)}$ & .0035 $_{(.0636)}$ & \textbf{.0006}$_{(.0103)}$ & .0018 $_{(.0373)}$ & .0012 $_{(.0220)}$ & .0012 $_{(.0219)}$ & .0028 $_{(.0509)}$ & .0018 $_{(.0373)}$ & .0011 $_{(.0209)}$ & .0015 $_{(.0280)}$ & .0035 $_{(.0636)}$ & .0035 $_{(.0636)}$ \\
O2DiffDys (V) & .0216 $_{(.0563)}$ & .0220 $_{(.0570)}$ & .2545 $_{(.6822)}$ & .0347 $_{(.0901)}$ & .0729 $_{(.2178)}$ & .0226 $_{(.0600)}$ & .0222 $_{(.0590)}$ & .0350 $_{(.0939)}$ & .0729 $_{(.2178)}$ & \textbf{.0213}$_{(.0553)}$ & .0241 $_{(.0644)}$ & .2545 $_{(.6822)}$ & .2545 $_{(.6822)}$ \\
Acidosis & .0099 $_{(.0371)}$ & .0098 $_{(.0368)}$ & .0837 $_{(.3374)}$ & \textbf{.0042}$_{(.0118)}$ & .0134 $_{(.0586)}$ & .0250 $_{(.1078)}$ & .0102 $_{(.0376)}$ & .0169 $_{(.0681)}$ & .0134 $_{(.0586)}$ & .0099 $_{(.0366)}$ & .0142 $_{(.0672)}$ & .0837 $_{(.3374)}$ & .0837 $_{(.3374)}$ \\
ThermoDys & \textbf{.0010}$_{(.0163)}$ & .0229 $_{(.3585)}$ & .0229 $_{(.3585)}$ & .0037 $_{(.0074)}$ & .0011 $_{(.0174)}$ & .0011 $_{(.0177)}$ & .0011 $_{(.0179)}$ & .0056 $_{(.0875)}$ & .0011 $_{(.0174)}$ & .0011 $_{(.0170)}$ & .0012 $_{(.0185)}$ & .0229 $_{(.3585)}$ & .0229 $_{(.3585)}$ \\
Tachy & .0658 $_{(.2332)}$ & 1.669 $_{(3.2743)}$ & .0588 $_{(.1477)}$ & .0346 $_{(.0640)}$ & .0697 $_{(.1831)}$ & .0465 $_{(.0879)}$ & .0489 $_{(.0960)}$ & \textbf{.0121}$_{(.0143)}$ & .0697 $_{(.1831)}$ & .0480 $_{(.0966)}$ & .0772 $_{(.2205)}$ & .9505 $_{(1.6498)}$ & .9533 $_{(1.6578)}$ \\
LeukDys & .0228 $_{(.0527)}$ & .0226 $_{(.0524)}$ & .1596 $_{(.3787)}$ & \textbf{.0170}$_{(.0275)}$ & .0385 $_{(.1069)}$ & .0245 $_{(.0569)}$ & .0237 $_{(.0552)}$ & .0250 $_{(.0585)}$ & .0385 $_{(.1069)}$ & .0225 $_{(.0522)}$ & .0220 $_{(.0518)}$ & .1596 $_{(.3787)}$ & .1596 $_{(.3787)}$ \\
HypCarb & .0096 $_{(.0390)}$ & .0096 $_{(.0394)}$ & .0610 $_{(.2698)}$ & .0086 $_{(.0285)}$ & .0099 $_{(.0403)}$ & .0098 $_{(.0397)}$ & .0099 $_{(.0400)}$ & .0126 $_{(.0556)}$ & .0099 $_{(.0403)}$ & .0096 $_{(.0392)}$ & \textbf{.0065}$_{(.0267)}$ & .0610 $_{(.2698)}$ & .0610 $_{(.2698)}$ \\
HypGly & .0294 $_{(.0419)}$ & .0286 $_{(.0425)}$ & .5685 $_{(.8929)}$ & \textbf{.0269}$_{(.0360)}$ & .0715 $_{(.1451)}$ & .0299 $_{(.0431)}$ & .0297 $_{(.0428)}$ & .0374 $_{(.0565)}$ & .0715 $_{(.1451)}$ & .0297 $_{(.0556)}$ & .0845 $_{(.1813)}$ & .5685 $_{(.8929)}$ & .5685 $_{(.8929)}$ \\
MyoIsch & .0119 $_{(.0530)}$ & .0121 $_{(.0537)}$ & .0687 $_{(.3163)}$ & .0111 $_{(.0471)}$ & .0245 $_{(.1446)}$ & .0312 $_{(.1555)}$ & .0120 $_{(.0531)}$ & .0157 $_{(.0725)}$ & .0245 $_{(.1446)}$ & .0117 $_{(.0517)}$ & \textbf{.0095}$_{(.0424)}$ & .0687 $_{(.3163)}$ & .0687 $_{(.3163)}$ \\
TissueIsch & .0108 $_{(.0343)}$ & .0106 $_{(.0338)}$ & .1035 $_{(.3595)}$ & \textbf{.0061}$_{(.0169)}$ & .0121 $_{(.0385)}$ & .0118 $_{(.0368)}$ & .0114 $_{(.0357)}$ & .0233 $_{(.0831)}$ & .0121 $_{(.0385)}$ & .0118 $_{(.0411)}$ & .0151 $_{(.0502)}$ & .1035 $_{(.3595)}$ & .1035 $_{(.3595)}$ \\
SEP3 & .0211 $_{(.0395)}$ & .0209 $_{(.0392)}$ & .1729 $_{(.3000)}$ & .0233 $_{(.0390)}$ & .0222 $_{(.0410)}$ & .0222 $_{(.0410)}$ & .0230 $_{(.0399)}$ & .0215 $_{(.0398)}$ & .0222 $_{(.0410)}$ & \textbf{.0207}$_{(.0392)}$ & .0225 $_{(.0424)}$ & .1729 $_{(.3000)}$ & .1729 $_{(.3000)}$ \\
\midrule[0.3pt]\bottomrule[1pt]
\end{tabular}
}
\end{small}
\end{center}
\end{table}


\subsubsection{Effect of DAG-inducing regularization}\label{appendix:DAG_effect_lambda}
Table~\ref{table:hyperparameter} shows that our proposed data-adaptive linear regularization can eliminate cycles the most efficiently, i.e., it can remove all the cycles even when the regularization strength is not very strong.
Although the continuous DAG regularization can also remove cycles, it cannot keep the informative lagged self-exciting components (as evidenced by the top right penal in Figure~\ref{fig:GC_graph_bp_reg}). Fortunately, our proposed variant can fix this issue and return a graph without cycles but with lagged self-exciting components. However, both DAG regularization-based approaches cannot achieve comparable prediction performance (in terms of both CE loss and Focal loss) with our proposed linear regularization.
Lastly, $\ell_1$ regularization does achieve better prediction performance (for some SADs it even achieves the best CE loss), but the resulting graph, i.e., the bottom left panel in Figure~\ref{fig:GC_graph_bp_reg}, shows that $\ell_1$ regularization removes all edges, leading to a very uninformative graph. That is to say, although $\ell_1$ regularization is capable of improving recovery accuracy, it cannot output a meaningful DAG without removing all edges, which agrees with our findings in the numerical simulation.

\begin{table}[htp]
\caption{Comparison of the quantitative evaluation metrics for different structural learning regularization strength hyperparameters $\lambda$'s. We select $\lambda$ (highlighted) based on the total CE loss on the validation dataset. We can observe that our proposed data-adaptive linear regularization can encourage the desired DAG structure most efficiently --- the number of directed cycles becomes zero when the regularization strength is just $10^{-5}$. Moreover, we can observe that proper regularization can help improve the prediction performance; in particular, our proposed regularization can achieve comparable performance to the best achievable one (see Tables~\ref{table:CE_1}, \ref{table:CE_2} and \ref{table:FL} for more evidence); although $\ell_1$ regularization performs the best quantitatively, it cannot output a meaningful and interpretable graph (see Figure~\ref{fig:GC_graph_bp_reg}).}\label{table:hyperparameter}
\begin{center}
\begin{small}
\resizebox{.95\textwidth}{!}{%
\begin{tabular}{lcccccccccccc}
\multicolumn{13}{c}{\normalsize{Proposed data-adaptive linear cycle elimination regularization.}} \\ 
\toprule[1pt]\midrule[0.3pt]
\multicolumn{2}{l}{{Regularization strength $\lambda$}} &  0 &  $10^{-6}$ &  $5 \times 10^{-6}$ &  $10^{-5}$ &  $5 \times 10^{-5}$ &  $10^{-4}$ &  ${10^{-3}}$ &  $\boldsymbol{10^{-2}}$ & $10^{-1}$ &  1 &  10 \\
\cmidrule(l){3-13}
\multirow{2}{*}{Total CE loss} & Validation & 4.2372 & 4.3367 & 4.1904 & 3.302 & 3.1296 & 3.1281 & 2.9746 & 2.9745 & 2.9745 & 2.9745 & 2.9745 \\
& Test & 4.1261 & 4.2158 & 4.0638 & 3.1982 & 3.0534 & 3.0482 & 2.8928 & 2.8927 & 2.8927 & 2.8927 & 2.8927 \\
\multirow{2}{*}{Total Focal loss} & Validation & 0.8281 & 0.8562 & 0.8161 & 0.5649 & 0.5189 & 0.517 & 0.4739 & 0.474 & 0.474 & 0.474 & 0.474 \\
& Test & 0.8113 & 0.8366 & 0.7944 & 0.5493 & 0.5101 & 0.506 & 0.4625 & 0.4625 & 0.4625 & 0.4625 & 0.4625 \\
\multicolumn{2}{l}{{Num. of Length-2 Cycles}} &  3 &  2 &  1 &  0 &  0 &  0 &  0 &  0 &  0 &  0 &  0 \\
\multicolumn{2}{l}{{Num. of Length-3 Cycles}} &  0 &  0 &  0 &  0 &  0 &  0 &  0 &  0 &  0 &  0 &  0 \\
\multicolumn{2}{l}{{Num. of Length-4 Cycles}} &  1 &  0 &  0 &  0 &  0 &  0 &  0 &  0 &  0 &  0 &  0 \\
\multicolumn{2}{l}{{Num. of Length-5 Cycles}} &  0 &  0 &  0 &  0 &  0 &  0 &  0 &  0 &  0 &  0 &  0 \\
\midrule[0.3pt]\bottomrule[1pt]
\end{tabular}
}
\end{small}
\end{center}

\begin{center}
\begin{small}
\resizebox{.95\textwidth}{!}{%
\begin{tabular}{lcccccccccccc}
\multicolumn{13}{c}{\normalsize{Continuous DAG regularization.}} \\ 
\toprule[1pt]\midrule[0.3pt]
\multicolumn{2}{l}{{Regularization strength $\lambda$}} &  0 &  $10^{-6}$ &  $5 \times 10^{-6}$ &  $10^{-5}$ &  $5 \times 10^{-5}$ &  $10^{-4}$ &  $10^{-3}$ &  ${10^{-2}}$ & $\boldsymbol{10^{-1}}$ &  1 &  10 \\
\cmidrule(l){3-13}
\multirow{2}{*}{Total CE loss} & Validation & 4.2372 & 4.2424 & 4.2429 & 4.2433 & 4.2578 & 4.2606 & 4.4538 & 5.6227 & 3.5055 & 3.7798 & 3.9898 \\
& Test & 4.1261 & 4.1276 & 4.1281 & 4.1284 & 4.1386 & 4.1408 & 4.3762 & 5.2663 & 3.3732 & 3.6167 & 3.8347 \\
\multirow{2}{*}{Total Focal loss} & Validation & 0.8281 & 0.8295 & 0.8296 & 0.8298 & 0.8346 & 0.8361 & 0.8943 & 1.2037 & 0.4979 & 0.4892 & 0.5426 \\
& Test & 0.8113 & 0.8119 & 0.812 & 0.8121 & 0.8158 & 0.817 & 0.88 & 1.1076 & 0.4584 & 0.444 & 0.5 \\
\multicolumn{2}{l}{{Num. of Length-2 Cycles}} &  3 &  3 &  3 &  3 &  3 &  3 &  3 &  0 &  0 &  0 &  0 \\
\multicolumn{2}{l}{{Num. of Length-3 Cycles}} &  0 &  0 &  0 &  0 &  1 &  1 &  1 &  0 &  0 &  0 &  0 \\
\multicolumn{2}{l}{{Num. of Length-4 Cycles}} &  1 &  1 &  1 &  1 &  1 &  1 &  0 &  0 &  0 &  0 &  0 \\
\multicolumn{2}{l}{{Num. of Length-5 Cycles}} &  0 &  0 &  0 &  0 &  0 &  0 &  1 &  0 &  0 &  0 &  0 \\
\midrule[0.3pt]\bottomrule[1pt]
\end{tabular}
}
\end{small}
\end{center}

\begin{center}
\begin{small}
\resizebox{.95\textwidth}{!}{%
\begin{tabular}{lcccccccccccc}
\multicolumn{13}{c}{\normalsize{Proposed variant of continuous DAG regularization.}} \\ 
\toprule[1pt]\midrule[0.3pt]
\multicolumn{2}{l}{{Regularization strength $\lambda$}} &  0 &  $10^{-6}$ &  $\boldsymbol{5 \times 10^{-6}}$ &  $10^{-5}$ &  $5 \times 10^{-5}$ &  $10^{-4}$ &  ${10^{-3}}$ &  $10^{-2}$ & $10^{-1}$ &  1 &  10 \\
\cmidrule(l){3-13}
\multirow{2}{*}{Total CE loss} & Validation & 4.2372 & 4.2372 & 4.2371 & 4.2371 & 4.2372 & 4.2374 & 4.4226 & 4.7451 & 4.8069 & 4.7351 & 4.2944 \\
& Test & 4.1261 & 4.1261 & 4.126 & 4.126 & 4.1261 & 4.1262 & 4.3257 & 4.6049 & 4.7039 & 4.6503 & 4.2195 \\
\multirow{2}{*}{Total Focal loss} & Validation & 0.8281 & 0.8281 & 0.8281 & 0.8281 & 0.8281 & 0.8282 & 0.8791 & 0.9631 & 0.977 & 0.9575 & 0.8396 \\
& Test & 0.8113 & 0.8114 & 0.8114 & 0.8114 & 0.8114 & 0.8114 & 0.8655 & 0.9385 & 0.9623 & 0.9474 & 0.832 \\
\multicolumn{2}{l}{{Num. of Length-2 Cycles}} &  3 &  3 &  3 &  3 &  3 &  3 &  0 &  0 &  0 &  0 &  0 \\
\multicolumn{2}{l}{{Num. of Length-3 Cycles}} &  0 &  0 &  0 &  0 &  0 &  0 &  0 &  0 &  0 &  0 &  0 \\
\multicolumn{2}{l}{{Num. of Length-4 Cycles}} &  1 &  1 &  1 &  1 &  1 &  1 &  0 &  0 &  0 &  0 &  0 \\
\multicolumn{2}{l}{{Num. of Length-5 Cycles}} &  0 &  0 &  0 &  0 &  0 &  0 &  0 &  0 &  0 &  0 &  0 \\
\midrule[0.3pt]\bottomrule[1pt]
\end{tabular}
}
\end{small}
\end{center}

\begin{center}
\begin{small}
\resizebox{.95\textwidth}{!}{%
\begin{tabular}{lcccccccccccc}
\multicolumn{13}{c}{\normalsize{$\ell_1$ regulairzation.}} \\ 
\toprule[1pt]\midrule[0.3pt]
\multicolumn{2}{l}{{Regularization strength $\lambda$}} &  0 &  $10^{-6}$ &  $5 \times 10^{-6}$ &  $10^{-5}$ &  $5 \times 10^{-5}$ &  $10^{-4}$ &  ${10^{-3}}$  &  $\boldsymbol{10^{-2}}$ & $10^{-1}$ &  1 &  10 \\
\cmidrule(l){3-13}
\multirow{2}{*}{Total CE loss} & Validation & 4.2372 & 4.2368 & 4.2561 & 4.259 & 4.0539 & 3.9913 & 2.9867 & 2.8397 & 2.8407 & 2.8407 & 2.8407 \\
& Test & 4.1261 & 4.1256 & 4.1257 & 4.1295 & 3.9449 & 3.8988 & 2.8634 & 2.75 & 2.7504 & 2.7505 & 2.7505 \\
\multirow{2}{*}{Total Focal loss} & Validation & 0.8281 & 0.8281 & 0.8329 & 0.8335 & 0.7816 & 0.7626 & 0.4676 & 0.4348 & 0.4368 & 0.4368 & 0.4368 \\
& Test & 0.8113 & 0.8113 & 0.8112 & 0.8121 & 0.7652 & 0.7522 & 0.4469 & 0.4216 & 0.4234 & 0.4234 & 0.4234 \\
\multicolumn{2}{l}{{Num. of Length-2 Cycles}} &  3 &  3 &  2 &  2 &  2 &  3 &  0 &  0 &  0 &  0 &  0 \\
\multicolumn{2}{l}{{Num. of Length-3 Cycles}} &  0 &  0 &  0 &  0 &  0 &  0 &  0 &  0 &  0 &  0 &  0 \\
\multicolumn{2}{l}{{Num. of Length-4 Cycles}} &  1 &  1 &  1 &  1 &  1 &  3 &  0 &  0 &  0 &  0 &  0 \\
\multicolumn{2}{l}{{Num. of Length-5 Cycles}} &  0 &  0 &  0 &  0 &  0 &  0 &  0 &  0 &  0 &  0 &  0 \\
\midrule[0.3pt]\bottomrule[1pt]
\end{tabular}
}
\end{small}
\end{center}
\end{table}


\subsubsection{Additional interpretation: top sepsis-contributing SADs}\label{appendix:sepsis_causes}
The ranking of SADs that cause Sepsis is reported in Figure 1 based on the triggering/exciting effect magnitude in Table~\ref{tab:content-summary}.
There are several key observations one can make from the above table: Firstly, different types of regularization all indicate almost the same set of causes of sepsis, re-affirming the plausibility of our discovered causal graph. Secondly, our identified top sepsis-contributing factors share certain similarities with the results of the winning teams using XGBoost \cite{du2019automated,zabihi2019sepsis,yang2020explainable} in the 2019 Physionet Challenge on Early Sepsis Prediction \cite{physionetChallenge}; In particular, compared to Table 2 in \cite{yang2020explainable}:

\begin{table}[htp]
\caption{Top sepsis-contributing SADs in recovered causal DAGs in Figure~\ref{fig:GC_graph_bp_reg}.}\label{tab:content-summary}
\begin{center}
\begin{small}
\resizebox{.9\textwidth}{!}{%
\begin{tabular}{l|l}
\hline
{Regularization (reg.)} & {Rank of causes of Sepsis} \\
\hline
Data-adaptive linear cycle elimination reg. & CNSDys $>$ DCO (V) $>$ Tachy $>$ ThermoDys \\
$\ell_1$ reg. & (Not avaliable)\\
Variant of Continuous DAG reg. & CNSDys $>$ DCO (V) $>$ RenDys $>$ Tachy $>$ ThermoDys \\
Continuous DAG reg. & CNSDys $>$ DCO (V) $>$ Tachy $>$ ThermoDys \\
\hline
\end{tabular}
}
\end{small}
\end{center}
\end{table}

\begin{itemize}
    \item CNSDys, defined as Glasgow Coma Scale (GCS) score $ < 14$ (which helps quantify Traumatic Brain Injuries (TBI)), is highly relevant to some ``severity indicators'' such as  ``ICU length of stay'' and ``time between hospital and ICU admission'', which are the top 2 contributing factors in \cite{yang2020explainable}. 
    
    Our analysis handled the ``severity'' by including patients with controlled Sequential Organ Failure Assessment (SOFA) scores. Since our dataset contains more information than the public-available version for the open challenge \cite{physionetChallenge}, the real example did not include less explainable variables such as ``ICU length of stay'' and ``time between hospital and ICU admission''. Furthermore, the identified top sepsis-contributing SAD (i.e., CNSDys) may help answer the question of which ``severity indicator'' actually contributes to sepsis onset.

    \item DCO (V) and Tachy are related to arterial pressure and the pulse (heart rate), which are also identified as top sepsis-contributing factors (No. 11, 13-15, 19 in Table 2 \cite{yang2020explainable}).

    Additionally, ThermoDys is related to temperature, which corresponds to the No. 3 and 20 factors in \cite{yang2020explainable}; RenDys, which is defined through creatinine and blood urea nitrogen (BUN), corresponds to factors No. 8 and 10 in \cite{yang2020explainable}.
\end{itemize}

\end{document}